\documentclass[10pt,journal,letterpaper,compsoc]{IEEEtran}
\ifCLASSOPTIONcompsoc

\ifCLASSINFOpdf
   \usepackage[pdftex]{graphicx}
   \graphicspath{{../images}}
\else
\fi

\usepackage[cmex10]{amsmath}
\usepackage{amssymb}
\usepackage{algorithmic}
\usepackage{algorithm}
\usepackage{xcolor}
\usepackage{multirow}
\usepackage{amsfonts, tikz}
\usepackage{pgflibraryshapes}
\usepackage{booktabs}
\usepackage{algorithmic}
\usepackage{algorithm}

\definecolor{sgray}{rgb}{0.8,0.8,0.8}
\definecolor{sred}{rgb}{0.8,0.,0.}
\definecolor{sred0}{rgb}{1.,1,0.}
\definecolor{sred1}{rgb}{1.,0.6875,0.}
\definecolor{sred2}{rgb}{1.,0.375,0.}
\definecolor{sred3}{rgb}{1.,0.0625,0.}
\definecolor{sred4}{rgb}{0.75,0.,0.}

\newcommand{\cO}{\mathcal{O}}

\newcommand{\model}{{\omega}}
\newcommand{\modelm}{\bar{\omega}}

\newcommand{\Model}{{\Omega}}
\newcommand{\Modelm}{\bar{\Omega}}
\newcommand{\bappearance}{{\boldsymbol{\theta}_{\boldsymbol{\mathit{a}}}}}
\newcommand{\appearance}{{\theta_{\mathit{a}}}}

\newcommand{\bspatial}{{\boldsymbol{\theta}_{\boldsymbol{g}}}}
\newcommand{\spatial}{{\theta_{g}}}

\newcommand{\Params}{{\Theta}}
\newcommand{\score}{{\mathrm{score}}}
\newcommand{\scorehat}{{\mathrm{\widehat{score}}}}
\DeclareMathOperator*{\supp}{supp}
\DeclareMathOperator*{\argmax}{arg\,max}

\definecolor{sblue}{rgb}{0.2,0.5,0.9}
\DeclareMathOperator\loss{score}

\newlength\savedwidth
\newcommand\whline{\noalign{\global\savedwidth\arrayrulewidth
\global\arrayrulewidth 1.1pt}%
\hline \noalign{\global\arrayrulewidth\savedwidth}}

\newcommand\wnhline{\noalign{\global\savedwidth\arrayrulewidth
\global\arrayrulewidth 0.9pt}%
\hline \noalign{\global\arrayrulewidth\savedwidth}}

\begin{document}
\title{Learning a Hierarchical Compositional Shape Vocabulary for Multi-class Object Representation}


\author{Sanja~Fidler,~\IEEEmembership{Member,~IEEE,}
        Marko~Boben, 
        and~Ale\v{s}~Leonardis,~\IEEEmembership{~Member,~IEEE}
\IEEEcompsocitemizethanks{\IEEEcompsocthanksitem S. Fidler is with University of Toronto, Canada, M. Boben
and A. Leonardis are with University of Ljubljana, Slovenia. A. Leonardis is also affiliated with University of Birmingham, UK.
\protect\\
E-mail: fidler@cs.toronto.edu, \{marko.boben,ales.leonardis\}@fri.uni-lj.si}
}

%


\IEEEcompsoctitleabstractindextext{%
\begin{abstract}
Hierarchies allow feature sharing between objects at multiple levels
of representation, can code exponential variability in a very
compact way and enable fast inference. This makes them potentially
suitable for learning and recognizing a higher number of object
classes. However, the success of the hierarchical approaches so far
has been hindered by the use of hand-crafted features or
predetermined grouping rules. This paper presents a novel framework
for \emph{learning} a hierarchical compositional shape vocabulary
for representing multiple object classes. The approach takes simple
contour fragments and learns their frequent spatial configurations.
These are recursively combined into increasingly more complex and
class-specific shape compositions, each exerting a high degree of
shape variability. At the top-level of the vocabulary, the
compositions are sufficiently large and complex to represent the
whole shapes of the objects. We learn the vocabulary layer after
layer, by gradually increasing the size of the window of analysis
and reducing the spatial resolution at which the shape
configurations are learned. The lower layers are learned jointly on
images of all classes, whereas the higher layers of the vocabulary
are learned incrementally, by presenting the algorithm with one
object class after another. The experimental results show that the
learned multi-class object representation scales favorably with the
number of object classes and achieves a state-of-the-art detection
performance at both, faster inference as well as shorter training
times.
\end{abstract}
}


\maketitle

\IEEEdisplaynotcompsoctitleabstractindextext

\ifCLASSOPTIONpeerreview
\begin{center} \bfseries EDICS Category: 3-BBND \end{center}
\fi
%
 \IEEEpeerreviewmaketitle

\section{Introduction}

\IEEEPARstart{V}{isual} categorization of objects has been an area
of active research in the vision community for decades. Ultimately,
the goal is to recognize and detect an increasing number of object
classes in images within an acceptable time frame. The problem
entangles three highly interconnected issues: the internal object
{\bf representation} which should compactly capture the high visual
variability of objects and generalize well over each class, means of
{\bf learning} the representation from a set of images with as
little supervision as possible, and an effective {\bf inference}
algorithm that robustly matches the object representation against
the image.

Using vocabularies of visual features has been a popular choice of
object class representation and has yielded some of the most
successful performances for object detection to
date~\cite{s:fergus07,s:torralba07,s:opelt08,s:shotton08,s:ferrari08,s:leibe08}.
However, the majority of these works are currently using flat coding
schemes where each object is represented with either no structure at
all by using a bag-of-words model or only simple geometry induced
over a set of intermediately complex object parts. In this paper,
our aim is to model the \emph{hierarchical compositional structure}
of the objects and do so for multiple object classes.

Modeling structure (geometry of objects) is important for several
reasons. Firstly, since objects within a class have usually
distinctive and similar shape, it allows for an efficient shape
parametrization with good generalization capabilities. It further
enables us to \emph{parse} objects into meaningful components which
is of particular importance in robotic applications where the task
is not only to detect objects but also to execute higher-level
cognitive tasks (manipulations, grasping, etc). Thirdly, by inducing
the structure over the features the representation becomes more
robust to background clutter.

Hierarchies incorporate structural dependencies among the features
at multiple levels: objects are defined in terms of parts, which are
further composed from a set of simpler constituents,
etc~\cite{s:triggs05,s:ullman06,s:fidler09,s:zhu06,s:fidler07,s:ommer07,Girshick09}.
Such architectures allow sharing of features between the visually
similar as well as dissimilar classes at multiple levels of
specificity~\cite{s:tsotsos90, s:fidler07,s:amit02,s:ommer07,s:fidler09c}. 
Sharing of features means sharing common computations and increasing
the speed of the joint detector~\cite{s:torralba07}. More
importantly, shared features lead to better
generalization~\cite{s:torralba07} and can play an important role of
regularization in learning of novel classes with few training
examples. Furthermore, since each feature in the hierarchy
recursively models certain variance over its parts, it captures a
high structural variability and consequently a smaller number of
features are needed to represent each class.

Learning of feature vocabularies without or with little supervision
is of primary importance in multi-class object representation. By
learning, we minimize the amount of time of human involvement in
object labeling~\cite{s:zhu06,s:hebert07} and avoid bias of
pre-determined grouping rules or manually crafted
features~\cite{s:dickinson06,s:sarkar94,s:gemanS02,s:felzenswalb07}.
Second, learning the representation statistically yields features
most shareable between the classes, which may not be well predicted
by human labelers~\cite{s:edelman02}. However, the complexity of
\emph{learning the structure} of a hierarchical representation
bottom-up and without supervision is enormous: there is a huge
number of possible feature combinations, the number of which
exponentially increases with each additional layer
--- thus an
effective learning algorithm must be employed. 

In this paper, the idea is to represent the objects with a
\emph{learned hierarchical compositional shape vocabulary} that has
the following architecture. The vocabulary at each layer contains a
set of hierarchical deformable models which we will call
\emph{compositions}. Each composition is defined recursively: it is
a hierarchical \emph{generative} probabilistic model that represents
a geometric configuration of a small number of parts which are
themselves hierarchical deformable models, i.e., compositions from a
previous layer of the vocabulary. We present a framework for
\emph{learning} such a representation for multiple object classes.
Learning is statistical and is performed bottom-up. The approach
takes simple oriented contour fragments and learns their frequent
spatial configurations. These are recursively combined into
increasingly more complex and class-specific shape compositions,
each exerting a high degree of shape variability. In the top-level
of the vocabulary, the compositions are sufficiently large and
complex to represent the whole shapes of the objects. We learn the
vocabulary layer after layer, by gradually increasing the size of
the window of analysis and the spatial resolution at which the shape
configurations are learned. The lower layers are learned jointly on
images of all classes, whereas the higher layers of the vocabulary
are learned incrementally, by presenting the algorithm with one
object class after another. We assume supervision in terms of a
positive
and a validation set of class images --- 
however, the structure of the vocabulary is learned in an
\emph{unsupervised} manner. That is, the number of compositions at
each layer, the number of parts for each of the compositions along
with the distribution parameters are inferred from the data without
supervision.

We experimentally demonstrate several important issues: {\bf 1.)}
Applied to a collection of natural images, the approach
\emph{learns} hierarchical models for various curvatures,
T- and L- junctions, i.e., features usually emphasized by the
Gestalt theory~\cite{s:wertheimer23}; {\bf 2.)} We show that these
generic compositions can be effectively used for object
classification; {\bf 3.)} For object detection we demonstrate a
competitive speed of detection with respect to the related
approaches already for a single class. {\bf 4.)} For multi-class
object detection we achieve a highly sub-linear growth in the size
of the hierarchical vocabulary at multiple layers and, consequently,
a scalable complexity of inference as the number of modeled classes
increases; {\bf 5.)} We demonstrate a competitive detection accuracy
with respect to the current state-of-the-art. Furthermore, the
learned representation is very compact --- a hierarchy modeling $15$
object classes uses only $1.6$Mb on disk.

The remainder of this paper is organized as follows. In
Sec.~\ref{sec:related_work} we review the related work.
Sec.~\ref{sec:representation} presents our hierarchical
compositional representation of object shape with recognition and
detection described in Sec.~\ref{sec:detection}. In
Sec.~\ref{sec:learning} our learning framework is
proposed. 
The experimental results are presented in Sec.~\ref{sec:results}.
The paper concludes with a summary and discussion in
Sec.~\ref{sec:discussion} and pointers to future work in
Sec.~\ref{sec:future_work}.

\section{Related work and contributions}
\label{sec:related_work}

{\bf Compositional hierarchies.} Several compositional approaches to
modeling objects have been proposed in the literature, however, most
of them relied on hand-crafted representations, pre-determined
grouping rules or supervised training with manually annotated object
parts~\cite{s:zhu06,s:sarkar94,s:gemanS02,s:ettinger87}. The reader
is referred to~\cite{s:zhu06,s:dickinson08} for a thorough review.

{\bf Unsupervised learning of hierarchical compositional vocabularies.}
Work on {\em unsupervised hierarchical learning}
has been relatively scarce. 
Utans~\cite{s:utans94} has been the first to address unsupervised
learning of compositional representations. The approach learned
hierarchical mixture models of feature combinations, and was utilized
on learning simple dot patterns.

Based on the Fukushima's model~\cite{s:fukushima83}, Riesenhuber and
Poggio~\cite{s:poggio99} introduced the HMAX approach which
represents objects with a 2-layer hierarchy of Gabor feature
combinations. The original HMAX used a vocabulary of pre-determined
features, while these have subsequently been replaced with randomly
chosen templates~\cite{s:serre07}. Since no statistical learning is
involved, as much as several thousands of features are needed to
represent the objects. 
An improved learning
algorithm has recently been proposed by Masquelier and
Thorpe~\cite{s:thorpe07}.

Among the neural network representatives, Convolutional
nets~\cite{s:lecun07,s:lecun09} have been most widely and
successfully applied to generic object recognition. The approach
builds a hierarchical feature extraction and classification system
with fast feed-forward processing. The hierarchy stacks one or
several feature extraction stages, each of which consists of filter
bank layer, non-linear transformation layers, and a pooling layer
that combines filter responses over local neighborhoods using an
average or max operation, thereby achieving invariance to small
distortions~\cite{s:lecun09}. A similar approach is proposed by
Hinton~\cite{s:hinton07} with recent improvements by Ng et
al.~\cite{s:ng09}.
One of the main drawbacks of these approaches, however, may be that
they do not explicitly model the spatial relations among the
features and are thus potentially less robust to shape variations.

Bouchard and Triggs~\cite{s:triggs05} proposed a $3-$layer hierarchy
(extension of the constellation model~\cite{s:fergus07}) and a
similar representation was proposed by Torralba et
al.~\cite{s:torralba07,s:torralba08}. For tractability, all of these
models are forced to use very sparse image information, where prior
to learning and detection, a small number (around $30$) of interest
points are detected. Using highly discriminative SIFT features might
limit their success in cluttered images or on structurally simpler
objects with little texture. The repeatability of the SIFT features
across the classes is also questionable~\cite{s:serre07}. On the
other hand, our approach is capable of dealing with several tens of
thousands of contour fragments as input. We believe that the use of
repeatable, dense and indistinctive contour fragments provide us
with a higher repeatability of the subsequent object representation
facilitating a better performance.

Epshtein and Ullman~\cite{s:ullman07a} approached the representation
from the opposite end; the hierarchy is built by \emph{decomposing}
object relevant image patches into recursively smaller entities. The
approach has been utilized on learning each class individually while
a joint multi-class representation has not been pursued. A similar
line of work was adopted by Mikolajczyk et
al.~\cite{s:mikolajczyk06} and applied to recognize and detect
multiple ($5$) object classes.

Todorovic and Ahuja~\cite{s:todorovic08} proposed a data-driven
approach where a hierarchy for each object example is generated
automatically by a segmentation algorithm. The largest repeatable
subgraphs are learned to represent the objects. Since bottom-up
processes are usually unstable, exhaustive grouping is employed
which results in long training and inference times.

Ommer and Buhmann~\cite{s:ommer07} proposed an unsupervised
hierarchical learning approach, which has been successfully
utilized for object classification. The features at each layer are
defined as histograms over a larger, spatially constrained area.
Our approach explicitly models the spatial relations among the features,
which should allow for a more reliable detection of objects with lower sensitivity to background clutter.

The learning frameworks most related to ours include the work
by~\cite{s:scalzo05,s:fleuret01} and just recently~\cite{s:lzhu08}.
However, all of these methods build \emph{separate} hierarchies for
each object class. This, on the one hand, avoids the massive number
of possible feature combinations present in diverse objects, but, on
the downside, does not exploit the shareability of features among the
classes.

Our approach is also related to the work on multi-class shape
recognition by Amit et al.~\cite{s:amit04,s:amit99,s:amit02}, and
some of their ideas have also inspired our approach. While the
conceptual representation is similar to ours, the compositions there
are designed by hand, the hierarchy
 has only three layers (edges, parts and objects), and the application is mostly targeted to reading
licence plates.

While surely equally important, the work on learning of visual
taxonomies~\cite{s:marszalek08,s:vangool08} of object classes
tackles the categorization/recognition process by hierarchical
cascade of classifiers. Our hierarchy is compositional and
generative with respect to object structure and does not address the
taxonomic organization of object classes. We proposed a model that is hierarchical both in structure as well as classes (taxonomy) in our later work~\cite{fidler10}.

Our approach is also related to the discriminatively trained grammars by Girshick et al.~\cite{Girshick09}, developed after
our original work was published. Like us, this approach models objects with deformable parts and subparts, the weights of which are trained using structure prediction. This approach has achieved impressive results for object detection in the past years. Its main drawback, however, is that the structure of the grammar needs to be specified by hand which is what we want to avoid doing here.

Compositional representations are currently not at the level of performance of supervised deep convolutional networks~\cite{krizhevsky12} which discriminatively train millions of weights (roughly three orders of magnitude more weights than our approach). To perform well these networks typically need millions of training examples, which is three or four orders of magnitude more than our approach. While originally designed for classification these networks  have recently been very successful in object detection when combined with bottom-up region proposals~\cite{girshick2014rcnn}. We believe that these type of networks can also benefit from the ideas presented here.

{\bf Contour-based recognition approaches.} Since our approach deals with object shape, we also briefly
review the related work on contour-based object class detection.

Contour fragments have been employed in the earlier work by Selinger
and Nelson~\cite{s:selinger98} and a number of follow-up works have
used a similar approach. Opelt et al.~\cite{s:opelt08} learned
\emph{boundary fragments} in a boosting framework and used them in a
Hough voting
scheme for object class detection. 
Ferrari et al.~\cite{s:ferrari07,s:ferrari08} extracted kPAS
features which were learned by combining $k$ roughly straight
contour fragments. The learned kPAS features resemble those obtained
in the lower hierarchical layers by our approach. 
Fergus et al.~\cite{s:fergus05,s:fergus07} represent objects as
constellations of object parts and propose an approach that learns
the model from images without supervision and in a scale invariant
manner. The part-based representation is similar to ours, however in
contrast, our approach learns a  hierarchical part-based
representation.

To the best of our knowledge, we are the first to 1.) \emph{learn}
generic shape structures from images at multiple hierarchical layers
and without supervision, which, interestingly, resemble those
predicted by the Gestalt theory~\cite{s:wertheimer23}, 2.) learn a
multi-class generative compositional representation in a bottom-up
manner from simple contour fragments, 3.) demonstrate scalability of
a hierarchical generative approach when the number of classes
increases --- in terms of the speed of inference, storage, and
training times.

This article conjoins and extends several of our conference papers
on learning of a hierarchical compositional vocabulary of object
shape~\cite{s:fidler09a,s:fidler07,s:fidler06,s:fidler08}.
Specifically, the original model has been reformulated
and more details have been included which could
not be given in the conference papers due to page constraints.
Further, more classes have been used in the experiments, the
approach has been tested on several standard recognition benchmarks
and an analysis with respect to the computational behavior of the
proposed approach has been carried out.

\section{A hierarchical compositional object representation}
\label{sec:representation}

The complete hierarchical framework addresses three major
issues: the representation, inference and learning. 
In this Section we present our hierarchical compositional object representation.

\def\IH{3.3cm}
\begin{figure}[t!]
\centering
\begin{tikzpicture}[style=thick, scale=1]
\pgftext[at=\pgfpoint{-2.7cm}{1.cm}]{\hspace{-1mm}\includegraphics[width=0.72\linewidth]{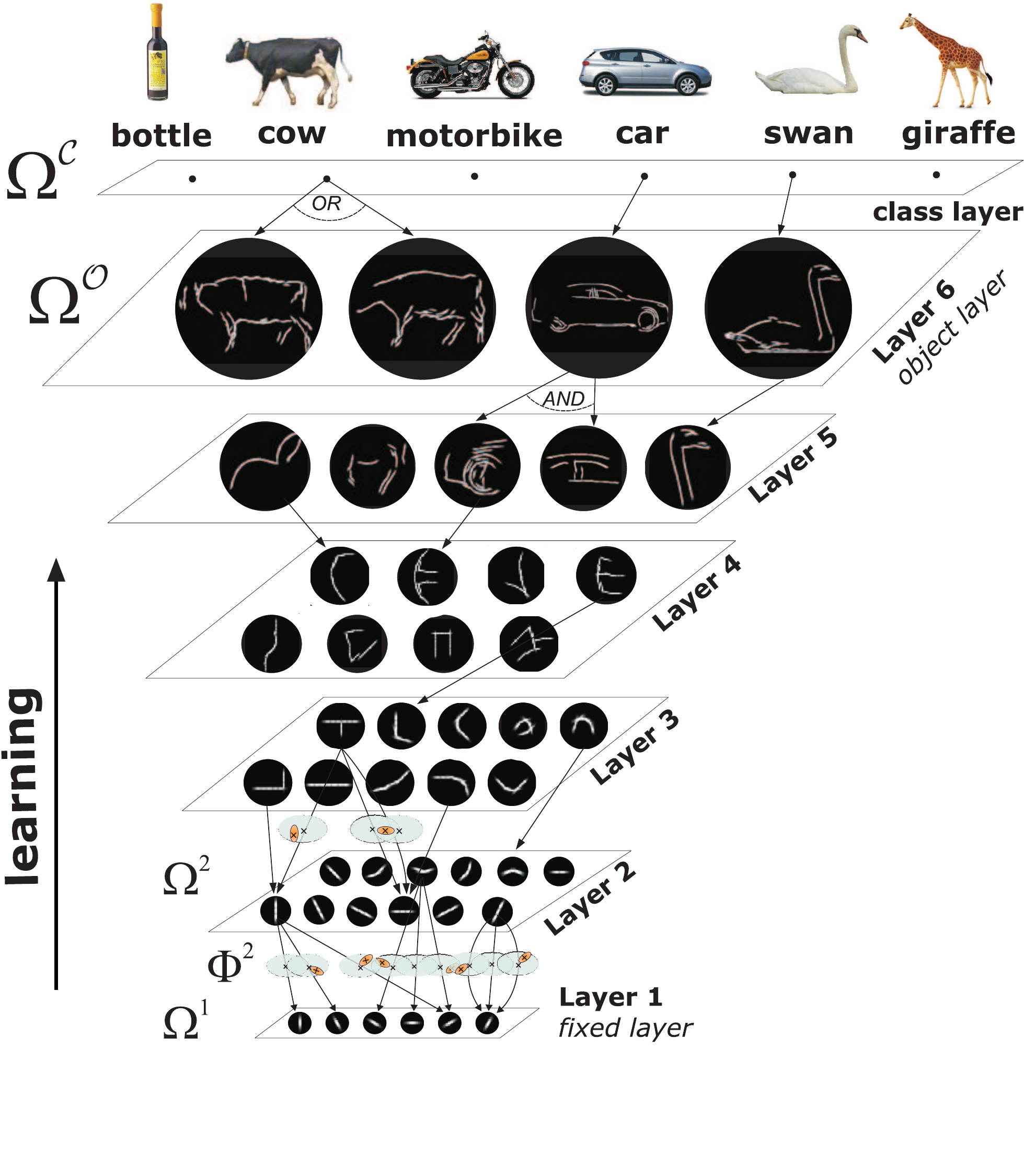}}
\pgftext[at=\pgfpoint{1.2cm}{0.1cm}]{\includegraphics[height=\IH]{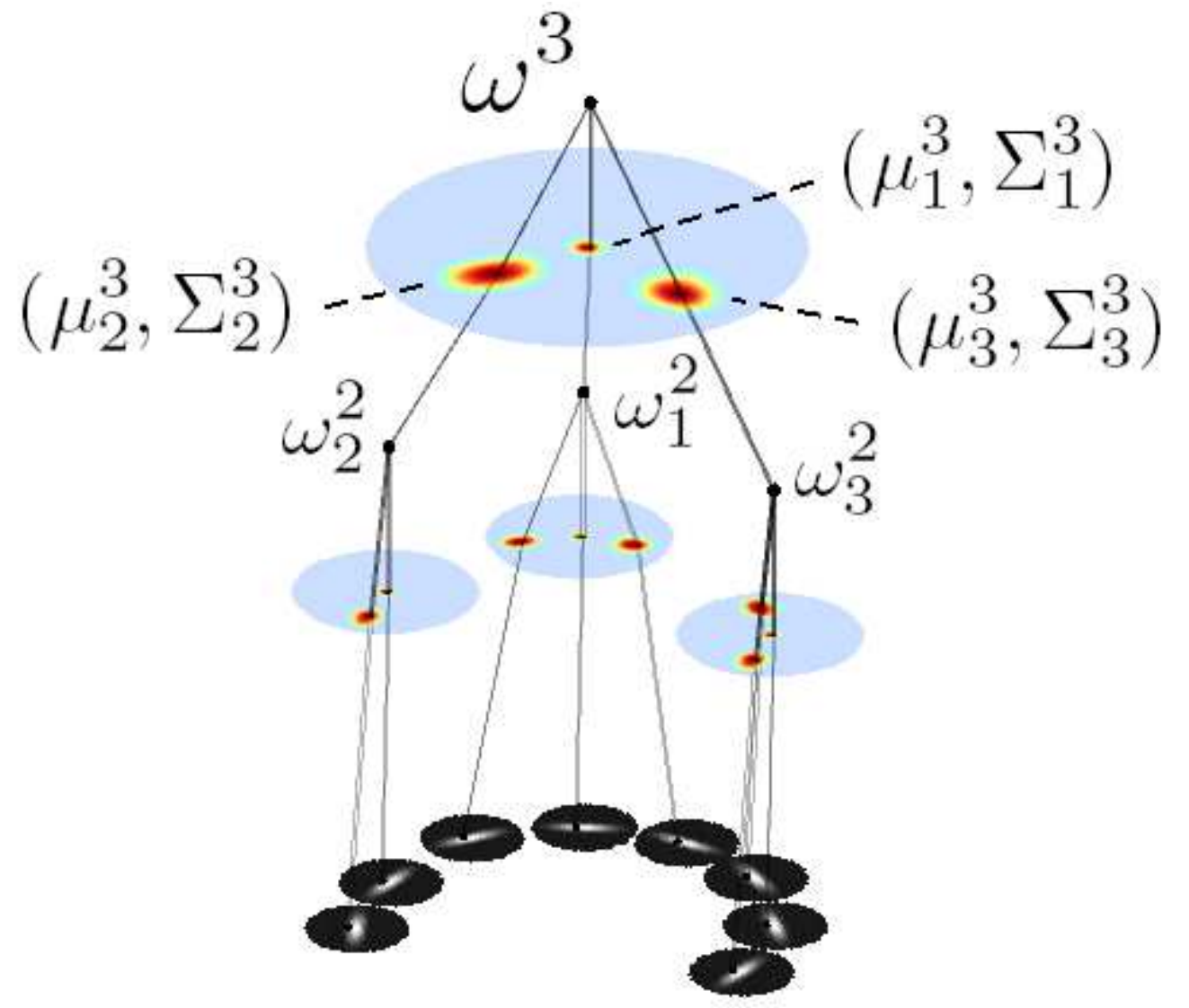}}

\draw[thin,dotted,->] (-2.2,0.4) --  (0.6,1.47);
\draw[thin,dotted,->] (-2.55,-0.44) .. controls (-0.9,-0.28) and (-0.0,-0.02) .. (1.02,0.42);
\draw[thin,dotted,->] (-2.7,-1.44) .. controls (-0.9,-1.39) and (-0.1,-1.22) .. (0.28,-1.14);
\draw[thin,dotted,->] (-3,-1.54) .. controls (-0.9,-1.7) and (-0.1,-1.56) .. (1.45,-1.45);
 \end{tikzpicture}

\vspace{-7mm}

 \caption{{\small {\bf Left}: Illustration of the
hierarchical vocabulary. The shapes depict the compositions
$\model^\ell$ and the links between them denote the compositional
relations between them. Note that there are no actual images stored
in the vocabulary -- the depicted shapes are \emph{mean} shapes represented by the compositions. {\bf Right}: An example of a complete composition
model from layer $3$. The blue patches are the limiting sizes within
which the model can generate shapes, while the red ellipses denote
the Gaussians representing the spatial relations between the
parts.}}
 \label{fig:composition}
\end{figure}

Our aim is to model the distribution of object \emph{shapes} in a
given class and we wish to do so for multiple object classes in a
computationally efficient way. The idea is to represent the objects
with a \emph{single learned hierarchical compositional shape vocabulary}
that has the following architecture. The vocabulary at each layer
contains a set of hierarchical deformable models which we will call
\emph{compositions}. Each composition is defined recursively: it represents a
geometric configuration of a small number of parts which are compositions
from the previous layer of the vocabulary. 
Different compositions can
share models for the parts, which makes the vocabulary efficient in
size and results in faster inference. 

The geometric
configuration of parts is modeled by relative spatial relations
between each of the parts and one part called a \emph{reference
part}. The hierarchical ``topology'' of a composition is assumed to be a
tree and is learned in a way to ensure that
its (sub)parts do not overlap or overlap minimally.

Note that a particular shape can be composed in multiple different ways, e.g.
a rectangle can be formed by two sets of parallel lines or four properly aligned right angles.
Thus some of the compositions may describe the same shape but have
very different topologies (different tree structures).
To deal with this issue, such compositions are grouped
 into OR nodes which act as mixture models (similarly as in~\cite{s:felzenswalb10}). This is done based on the geometric similarity of the shape they represent.

The structure of the compositions will be learned in an \emph{unsupervised manner} from images, along
with the corresponding parameters. The number of compositions that
make a particular layer in the vocabulary will also not be set in
advance but determined through learning. We will learn the layers $2$ and $3$ without any
supervision, from a pool of unlabeled images. To learn the higher layers, we will assume
bounding boxes with object labels, i.e., the set of positive and
validation boxes of objects for each class needs to be given. The
amount of supervision the approach needs is further discussed in
Sec.~\ref{sec:multiclass}.

At the lowest (first) layer, the hierarchical vocabulary consists of
a small number of base models that represent short contour fragments
at coarsely defined orientations. The number of orientations is
assumed given. At the top-most layer the compositions will represent
the shapes of the whole objects. For a particular object class, a
number of top-layer compositions will be used to represent the
whole distribution of shapes in the class, i.e., each top-layer
composition will model a certain aspect (3D view or an articulation)
of the objects within the class. The left side of
Fig.~\ref{fig:composition} illustrates the representation.

Due to the recursive definition, a composition at any layer of the
vocabulary is thus an ``autonomous'' deformable model. As such, a
composition at the top layer can act as a detector for an object
class, while a composition at an intermediate layer acts as a
detector for a less complex shape (for example, an L junction).
Since we are combining deformable models as we go up in the
hierarchy, we are capturing an increasing variability into the
representation while at the same time preserving the spatial
arrangements of the shape components.

We will use the following notation. Let $\Model$ denote the set and
structure of all compositions in the vocabulary and $\Params$ their
parameters. Since the vocabulary has a hierarchical structure we
will write it as $\Model = \Model^1 \cup \Modelm^1 \cup \Model^2 \cup \Modelm^2 \cup \cdots \cup
\Model^\cO\cup \Model^C$ where $\Model^\ell = \{\model_i^\ell\}_i$
is a set of compositions at layer $\ell$ and $\Modelm^\ell = \{\modelm_i^\ell\}_i$
is a set of OR-compositions from layer $\ell$, more specifically, 
$\modelm_i^\ell \subseteq \Model^\ell$.
Whenever it is clear from
the context, we will omit the index $i$. With $\Model^\cO$ we denote
the top, \emph{object layer} of compositions (we will use $\cO=6$ in
this paper, and this choice is discussed in
Sec.~\ref{sec:multiclass}), which roughly code the whole shapes
of the objects. The final, \emph{class layer} $\Model^C$, is an OR layer 
of object layer compositions ($\Model^\cO$), i.e., each class layer composition pools all of the corresponding object layer compositions from $\Model^\cO$ for each class separately.

The definition of a composition with respect to just one layer below is similar to that of the constellation model~\cite{s:fergus05}.
A \emph{composition} $\model^\ell$, where $\ell>1$, consists of $P$ \emph{parts} ($P$ can differ across compositions), with \emph{appearances}
$\bappearance^{\ell}=[\appearance_{j}^\ell]_{j=1}^P$ and 
\emph{geometric parameters} $\bspatial^{\ell}=[\spatial_{j}^\ell]_{j=1}^P$. The parts are compositions from the previous OR layer $\Modelm^{\ell - 1}$. 
One of the parts that forms $\model^\ell$ is taken as a reference and the positions of other 
parts are defined relative to its position. We will refer to such a part as a \emph{reference part}.

The appearance $\appearance_{j}^\ell$ is an $N^{\ell-1}$ dimensional vector, where $N^{\ell-1} = |\Modelm^{\ell - 1}|$ is the number
of OR-compositions at layer $\ell-1$, and $\appearance_{j}^\ell(k)$ represents the \emph{compatibility} between $\modelm_k^{\ell-1}$ and $j$-th part of $\model^\ell$. This vector will be sparse, thus only the non-zero entries need to be kept. 
For example, imagine composing an object model for the class \emph{car}. A car can have many different shapes for the trunk or the front of the car. Thus a car detector should ``fire'' if any part that looks like any of the trunks is present in the rear location and any front part in the front location of the detector. 
We allow $\appearance_j^\ell$ to have multiple positive entries only at the object level $\Modelm^\cO$, for other layers we consider $\appearance_j^\ell$ to be non-zero only at one entry (in this case, equal to $1$).

The geometric relations between the position of each of the parts
relative to the reference part are modeled by two-dimensional
Gaussians 
$\spatial_{j}^\ell=(\mu_{j}^\ell,
\Sigma_{j}^\ell)$. For the convenience of notation later on, we will
also represent the location of a reference part with respect to
itself with a Gaussian having zero mean and small variance,
$\epsilon^2 \mathrm{Id}$.

We allow for a composition to also have ``repulsive'' parts. These are the parts that the composition cannot consist of. For example, a handle is a repulsive
part for a cup since its presence indicates the class mug. We introduce the repulsive parts to deal with compositions that are supersets of one another. 

The models at the first layer of the vocabulary are defined over the
space of image features, which will in this paper be $n$-dimensional
Gabor feature vectors (explained in Sec.~\ref{sec:features}). 

\begin{figure}
\begin{minipage}{0.32\linewidth}
\includegraphics[width=\linewidth,height=2.2cm,trim=40 60 90
60,clip=true]{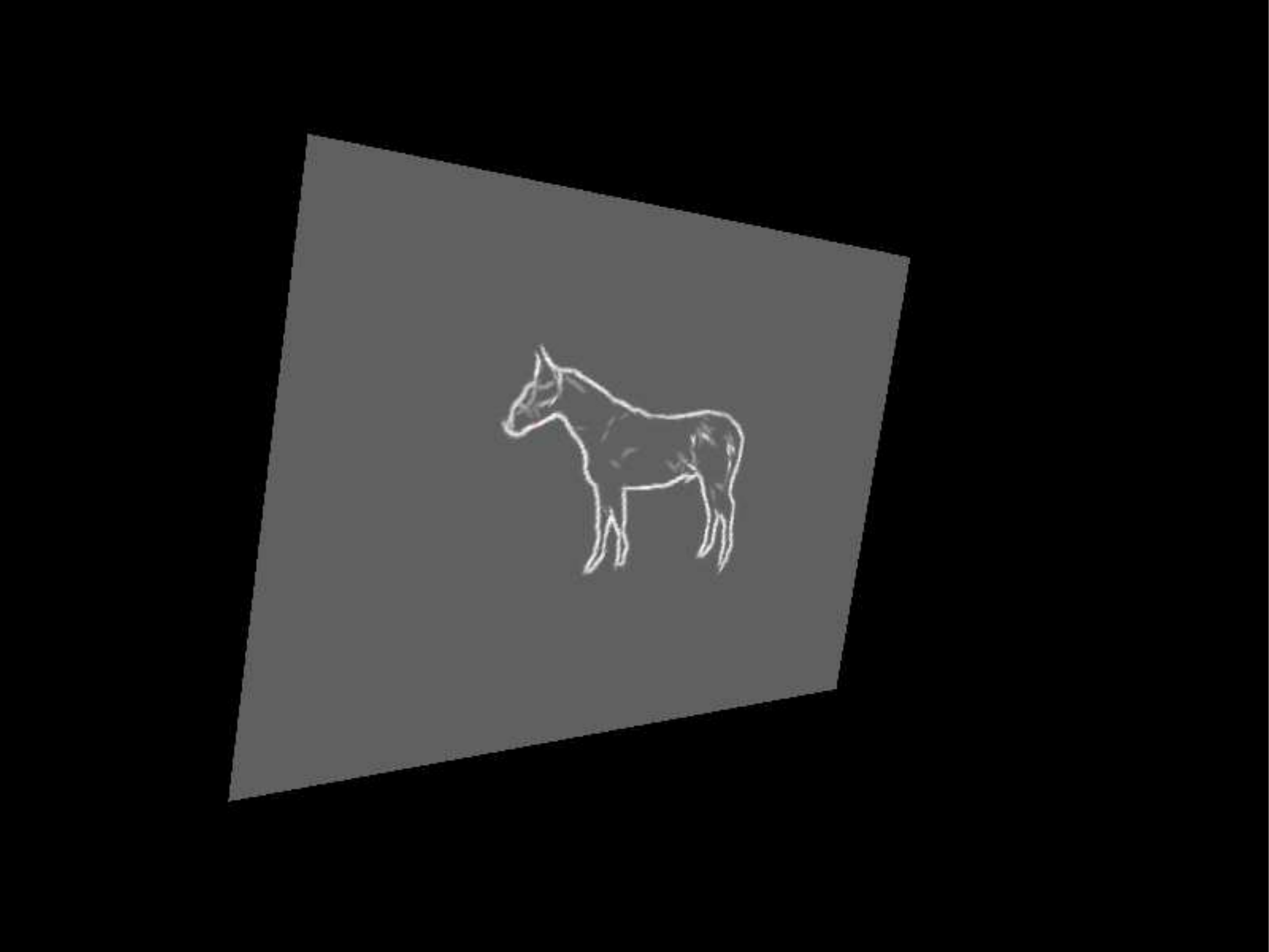}\\[-5.8mm]
\hspace*{1.6cm}{\setlength\fboxsep{1.1pt}\setlength\fboxrule{0pt}\colorbox{sblue}{\framebox[1.15cm][l]{\footnotesize
$\,$Layer 1}}}
\end{minipage}
\begin{minipage}{0.32\linewidth}
\includegraphics[width=\linewidth,height=2.2cm,trim=0 15 35
34,clip=true]{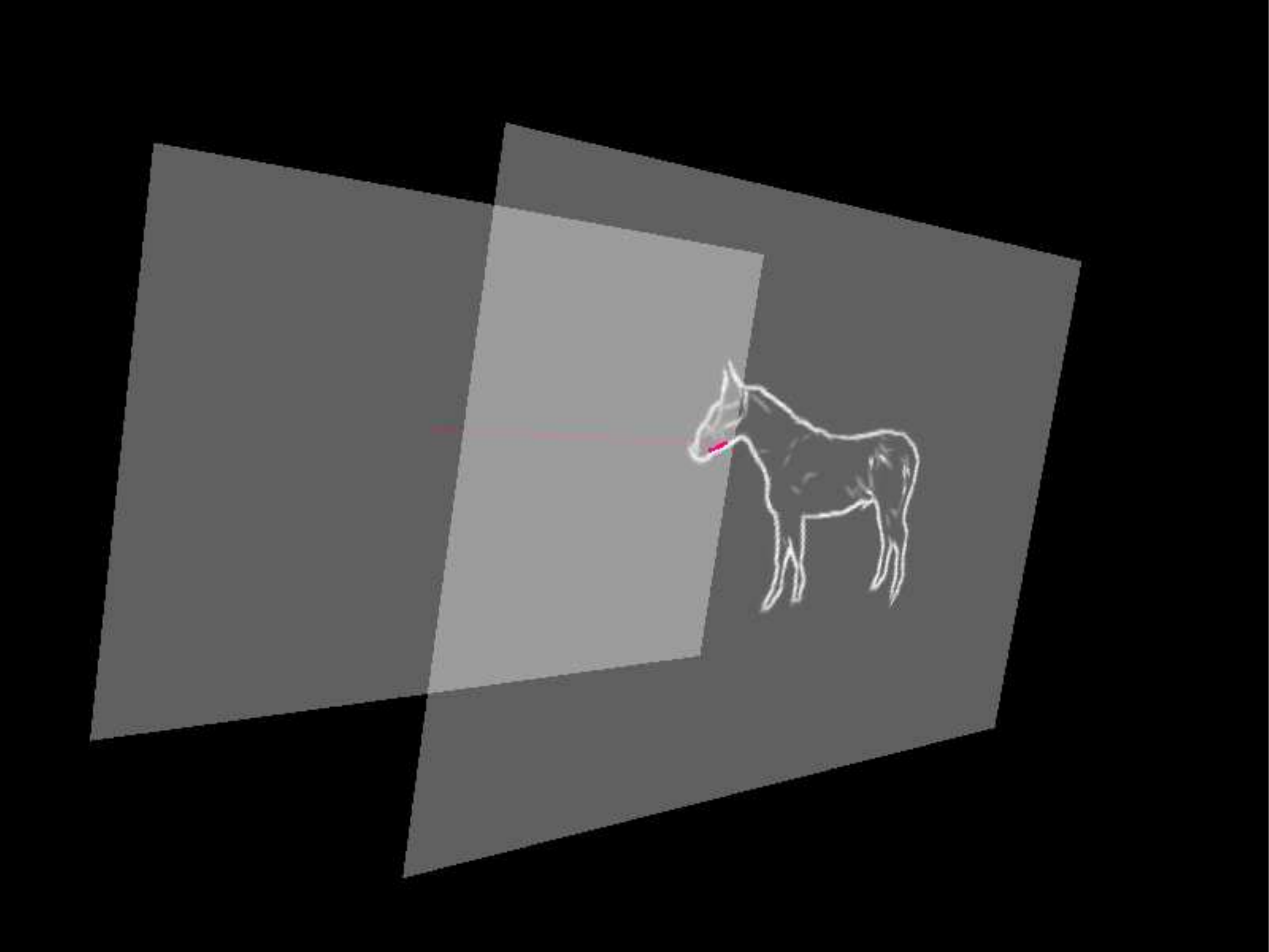}\\[-5.8mm]
\hspace*{1.6cm}{\setlength\fboxsep{1.1pt}\setlength\fboxrule{0pt}\colorbox{sblue}{\framebox[1.15cm][l]{\footnotesize
$\,$Layer 2}}}
\end{minipage}
\begin{minipage}{0.32\linewidth}
\includegraphics[width=\linewidth,height=2.2cm,trim=66 18 0 68,clip=true]{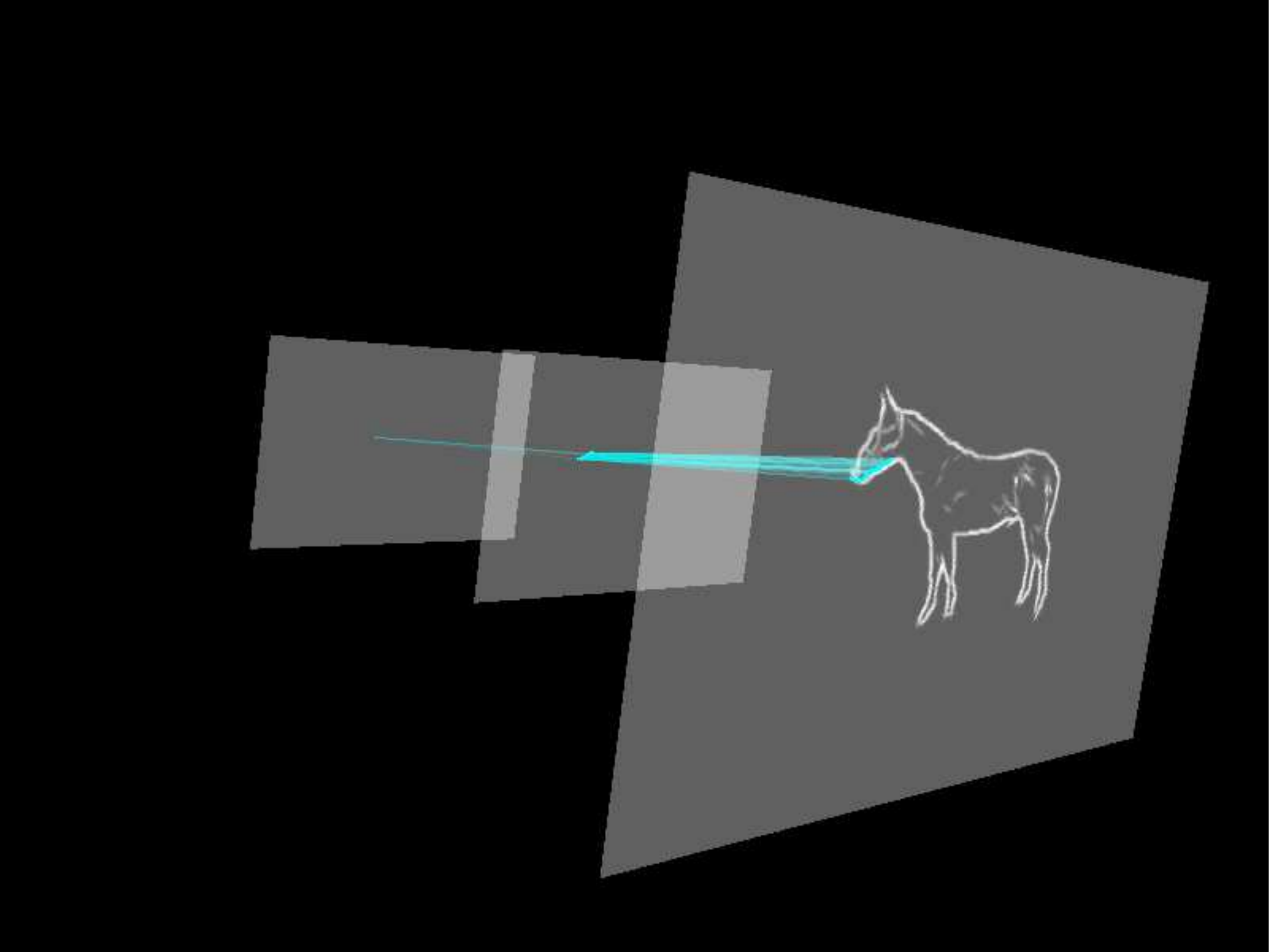}\\[-5.8mm]
\hspace*{1.6cm}{\setlength\fboxsep{1.1pt}\setlength\fboxrule{0pt}\colorbox{sblue}{\framebox[1.15cm][l]{\footnotesize
$\,$Layer 3}}}
\end{minipage}\\[1mm]
\begin{minipage}{0.32\linewidth}
\includegraphics[width=\linewidth,height=2.2cm,trim=155 38 0 118,clip=true]{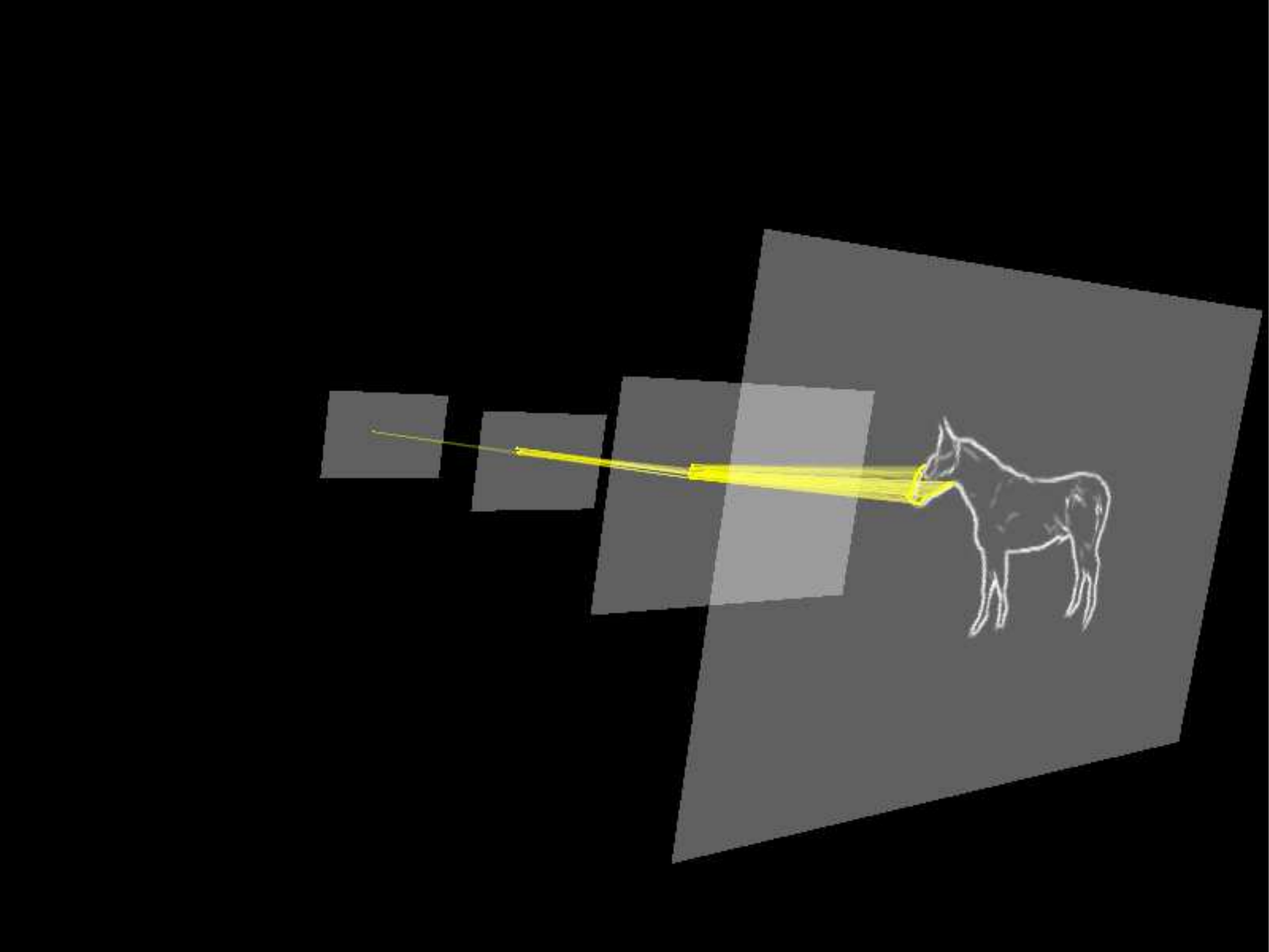}\\[-5.8mm]
\hspace*{1.6cm}{\setlength\fboxsep{1.1pt}\setlength\fboxrule{0pt}\colorbox{sblue}{\framebox[1.15cm][l]{\footnotesize
$\,$Layer 4}}}
\end{minipage}
\begin{minipage}{0.32\linewidth}
\includegraphics[width=\linewidth,height=2.2cm,trim=190 50 0 135,clip=true]{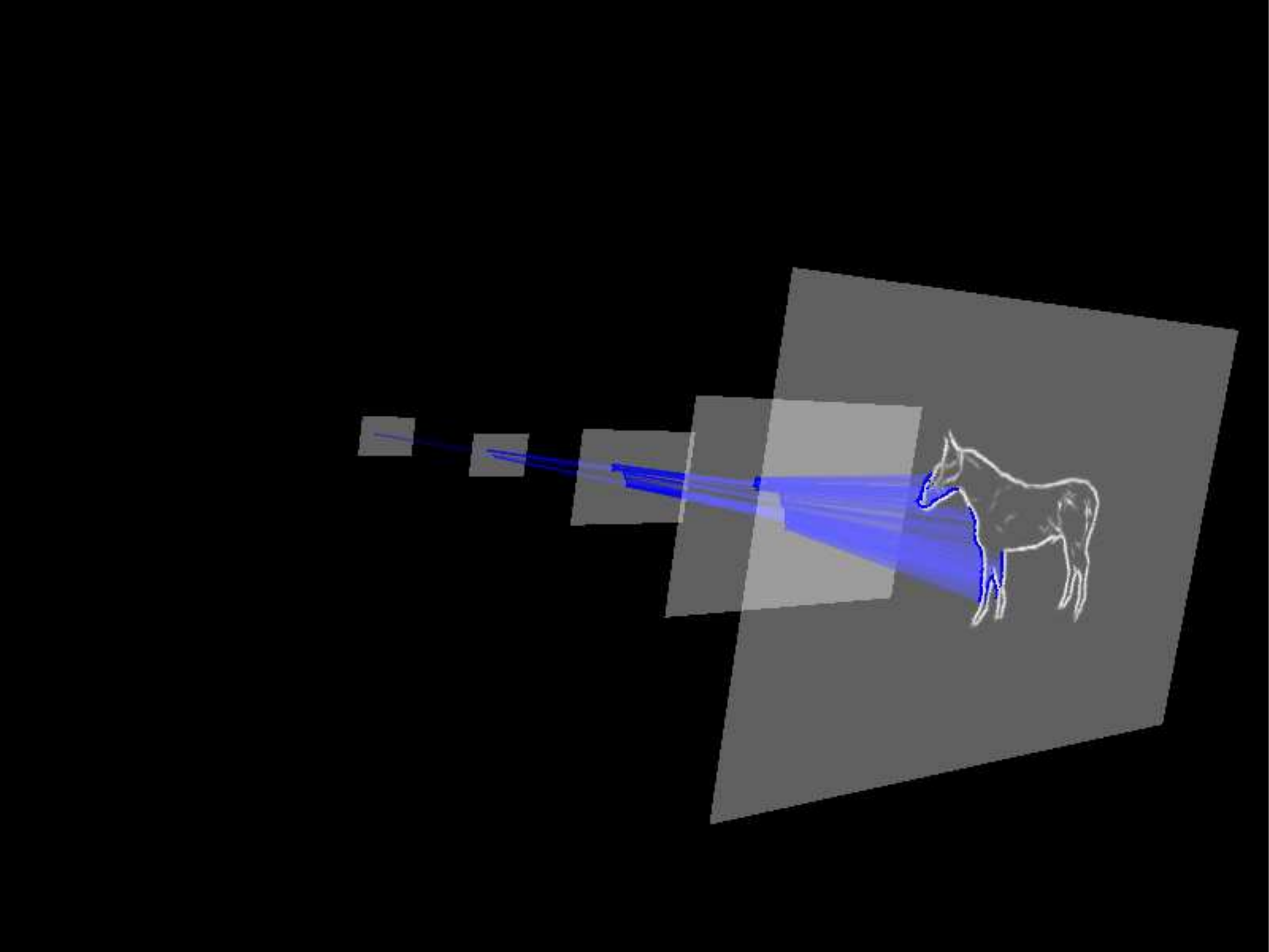}\\[-5.8mm]
\hspace*{1.6cm}{\setlength\fboxsep{1.1pt}\setlength\fboxrule{0pt}\colorbox{sblue}{\framebox[1.15cm][l]{\footnotesize
$\,$Layer 5}}}
\end{minipage}
\begin{minipage}{0.32\linewidth}
\includegraphics[width=\linewidth,height=2.2cm,trim=210 80 2 150,clip=true]{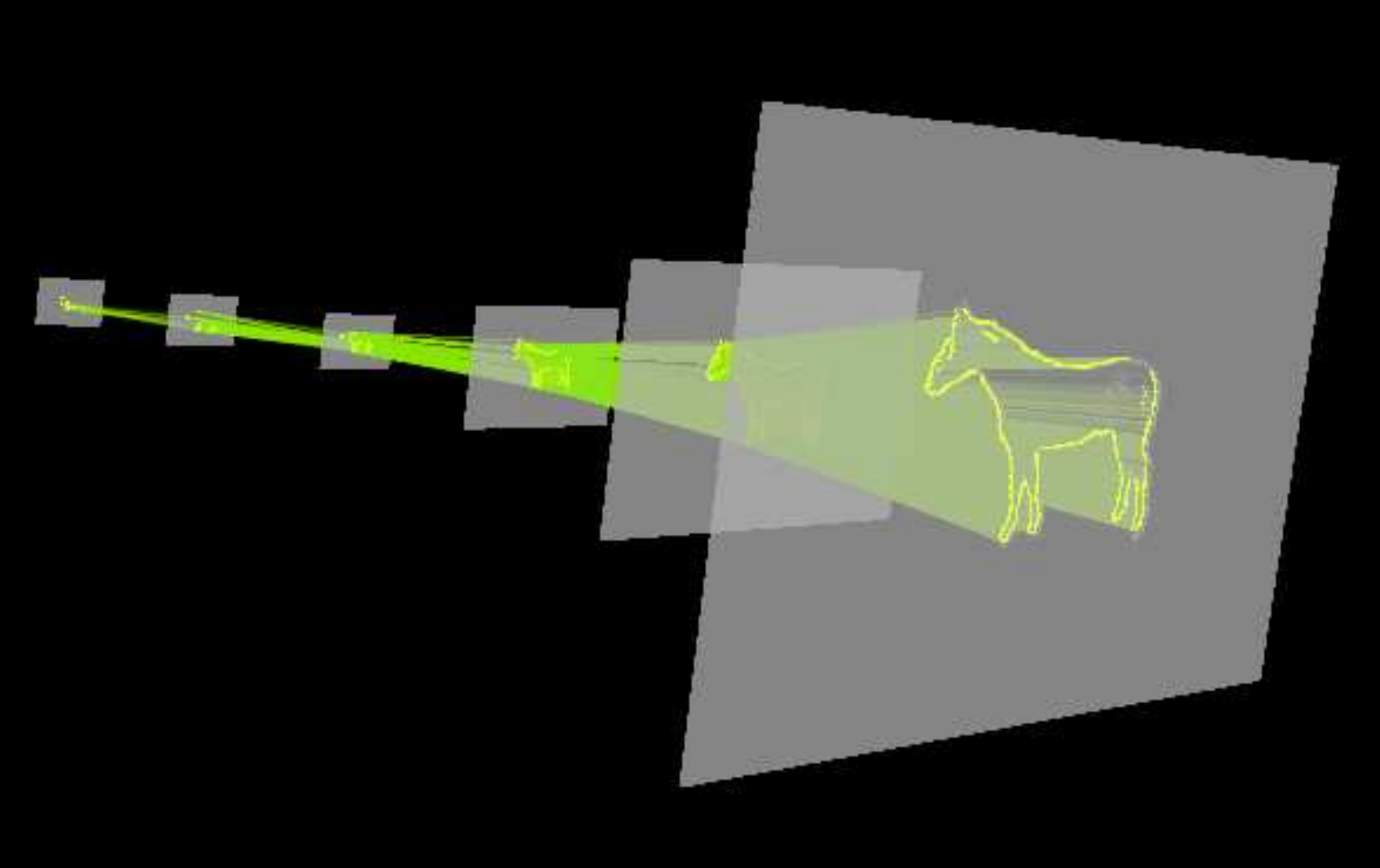}\\[-5.8mm]
\hspace*{1.6cm}{\setlength\fboxsep{1.1pt}\setlength\fboxrule{0pt}\colorbox{sblue}{\framebox[1.15cm][l]{\footnotesize
$\,$Layer 6}}} 
\end{minipage}
\caption{Fig. shows inferred parse graphs
(Sec.~\ref{sec:inference}) for compositions from
different layers of the hierarchy. The complexity and the size of
shapes explained by the compositions gradually increases with the
level of hierarchy.}
 \label{fig:sizerec}
\end{figure}

Note that each composition can only model shapes within a limited
spatial extent, that is, the window around a modeled shape is of
limited size and the size is the same for all compositions at a
particular layer $\ell$ of the vocabulary. We will denote it with
$r^\ell$. The 
size $r^\ell$ increases 
with the level of hierarchy. Fig.~\ref{fig:sizerec} depicts the gradual
increase in complexity and spatial scope of the compositions in the
hierarchy.

\section{Recognition and detection}
\label{sec:detection}

The model is best explained by first considering inference. Thus,
let us for now assume that the representation is already known and
we are only concerned with recognizing and detecting objects in a
query image given our hierarchical vocabulary. How we learn the representation will be
explained in Sec.~\ref{sec:learning}.

Sec.~\ref{sec:features} explains the image features we use, which
form a set of observations, and how we extract them
from an image. 
Sec.~\ref{sec:inference} describes the inference of the hidden
states of the model, while Sec.~\ref{sec:effcomp} shows how
to perform it in a computationally efficient way.

\subsection{Extracting image features}
\label{sec:features}

Let $I$ denote a query image. The features which we extract from $I$
should depend on how we define the base models at the first layer
$\Model^1$ of the vocabulary. In this paper we choose to use
oriented edge fragments, however, the learning and the recognition
procedures are general and independent of this particular choice.

In order to detect oriented edges in an image we use a Gabor filter
bank: 
\begin{gather*}
\ g_{\lambda,\varphi,\gamma,\sigma}(x,y,\psi)=\
e^{-\tfrac{u^2+\gamma^2
v^2}{2\sigma^2}}\cos\big(\frac{2\pi u}{\lambda}+\varphi\big)\\
u=x\cos\psi-y\sin\psi,\ \ v=x\sin\psi+y\cos\psi,
\end{gather*}
where $(x,y)$ represents the center of the filter's domain, 
and the parameters in this paper are set to
$(\lambda,\gamma,\sigma)=(6,0.75,2)$.
A set of two filter banks is used, one with even, $\varphi=0$, and
the other with odd, $\varphi=-\tfrac{\pi}{2}$, Gabor kernels defined
for $n$ equidistant orientations, $\psi=i\tfrac{\pi}{n},\
i=0,1,\dots,n-1$. For the experiments in this paper we use $n=6$.

We convolve the image $I$ with both filter banks and compute the
total energy for each of the orientations~\cite{s:petkov95}:
\begin{equation}
\mathcal E(x,y,\psi)=\sqrt{r_0^2(x,y,\psi)+r_{-\pi/2}^2(x,y,\psi)},
\end{equation}
where $r_{0}(x,y,\psi)$ and $r_{-\pi/2}(x,y,\psi)$ are the
convolution outputs of even and odd Gabor filters at location
$(x,y)$ and orientation $\psi$, respectively. We normalize $\mathcal
E$ to have the highest
value in the image equal to $1$. 
We further perform a non-maxima suppression over the total energy
$\mathcal E$ to find the locations of the local maxima for each of
the orientations. This is similar to performing the Canny operator
and taking the locations of the binary responses. At each of these
locations $(x,y)$, the set of which will be denoted with
$\mathbf{X}$, we extract Gabor features $\mathbf f=\mathbf
f(x,y)$ which are $n$-dimensional vectors containing the orientation
energy values in $(x,y)$, $\mathbf{f}(x,y)=\big[\mathcal
E(x,y,i\tfrac{\pi}{n})\big]_{i=0}^{n-1}$. Specifically, the feature
set $\mathbf F$ is the set of all feature vectors extracted in the
image, $\mathbf{F}=\{\mathbf{f}(x,y), (x,y)\in \mathbf{X}\}$. The
number of features in $\mathbf F$ is usually around $10^4$ for an
image of average size and texture. The image features $(\mathbf
F,\mathbf X)$ serve as the \emph{observations} to the recognition
procedure. Figure~\ref{fig:filtered_image} shows an example of
feature extraction.

\def\IH{2.1cm}
\begin{figure}[t!]
\centering
\includegraphics[height=\IH,width=2.85cm]{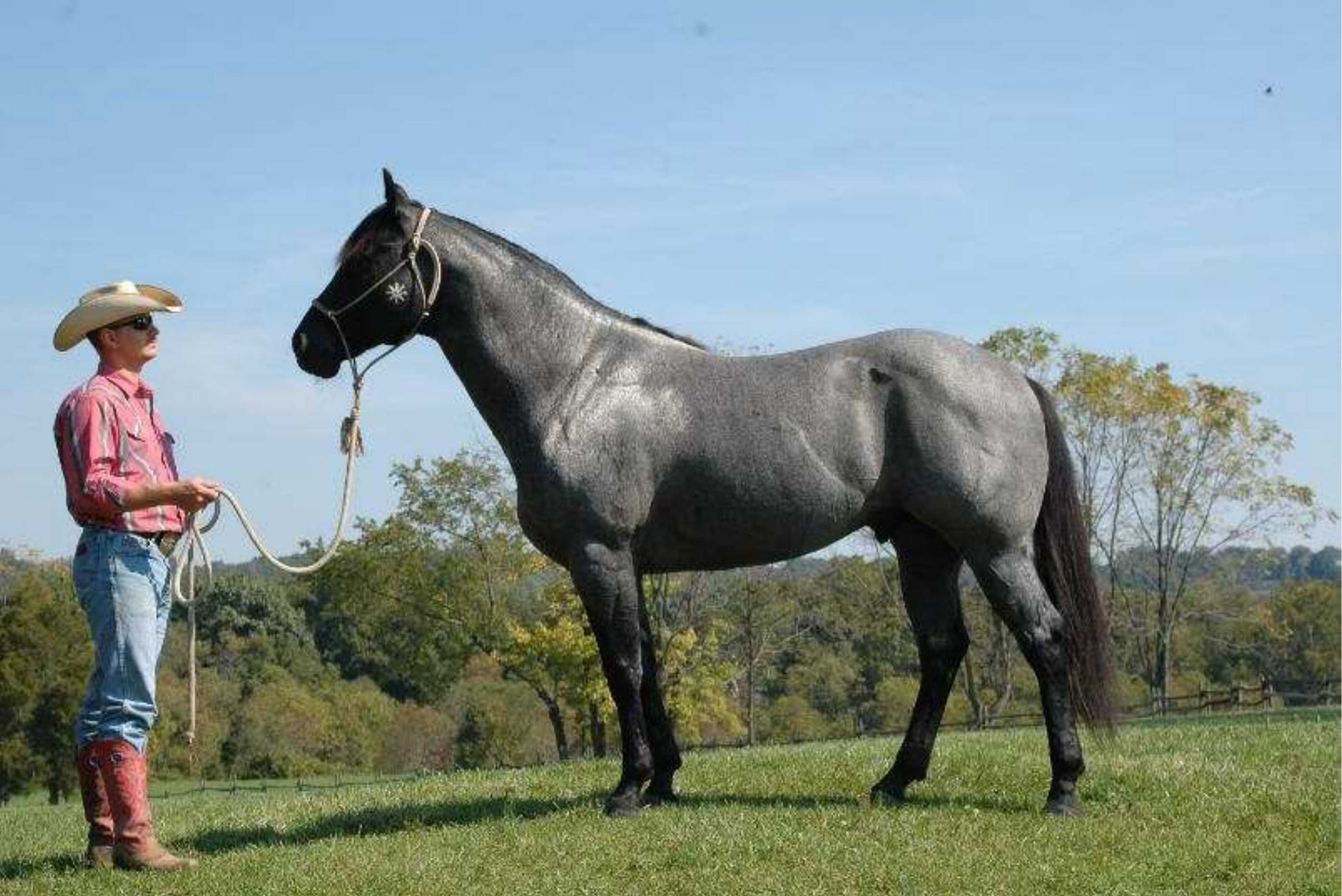}
\includegraphics[height=\IH,width=2.85cm,trim=3 29 3
              12,clip=true]{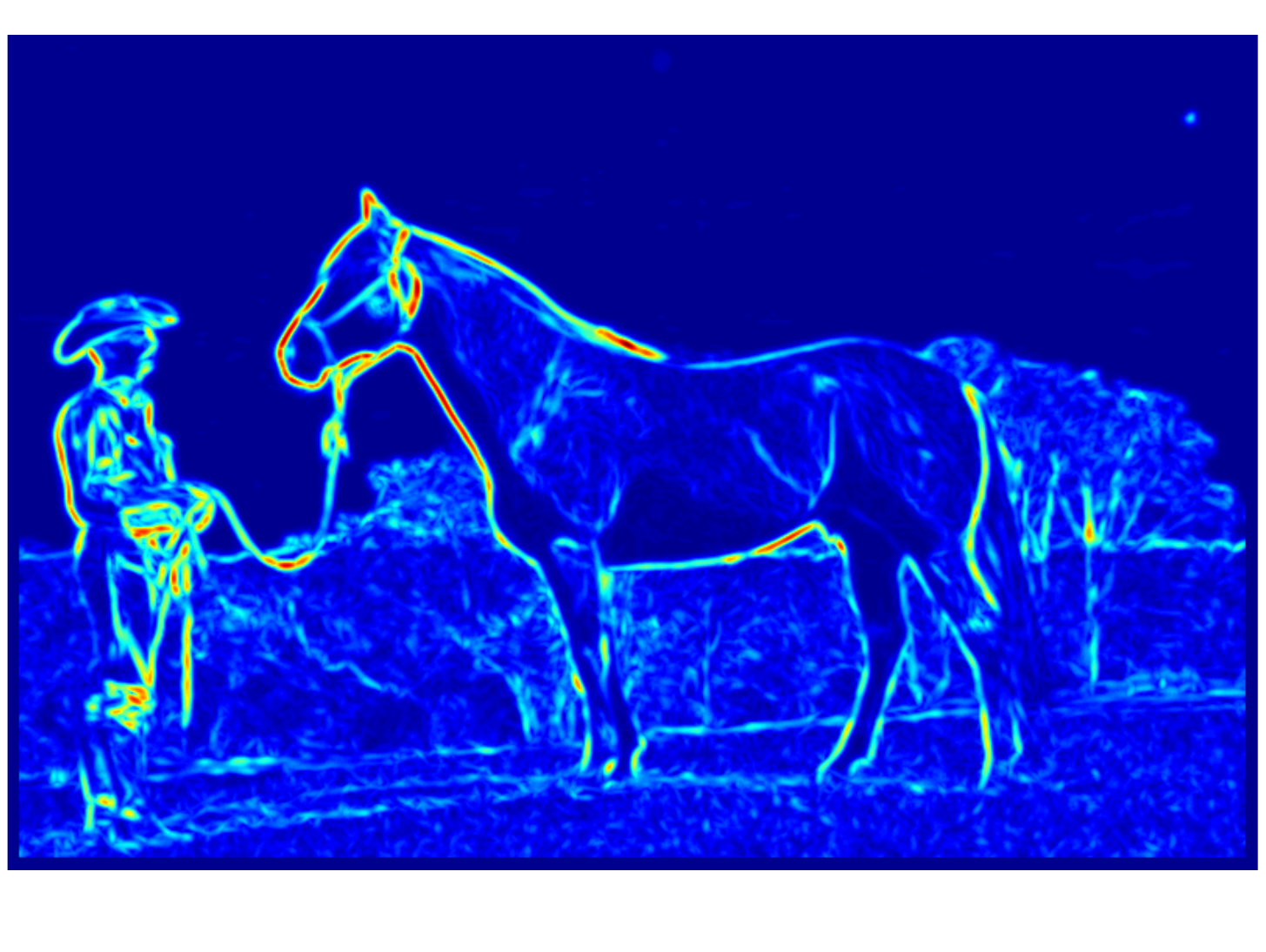}
\includegraphics[height=\IH,width=2.85cm,trim=3 29 3 12,clip=true]{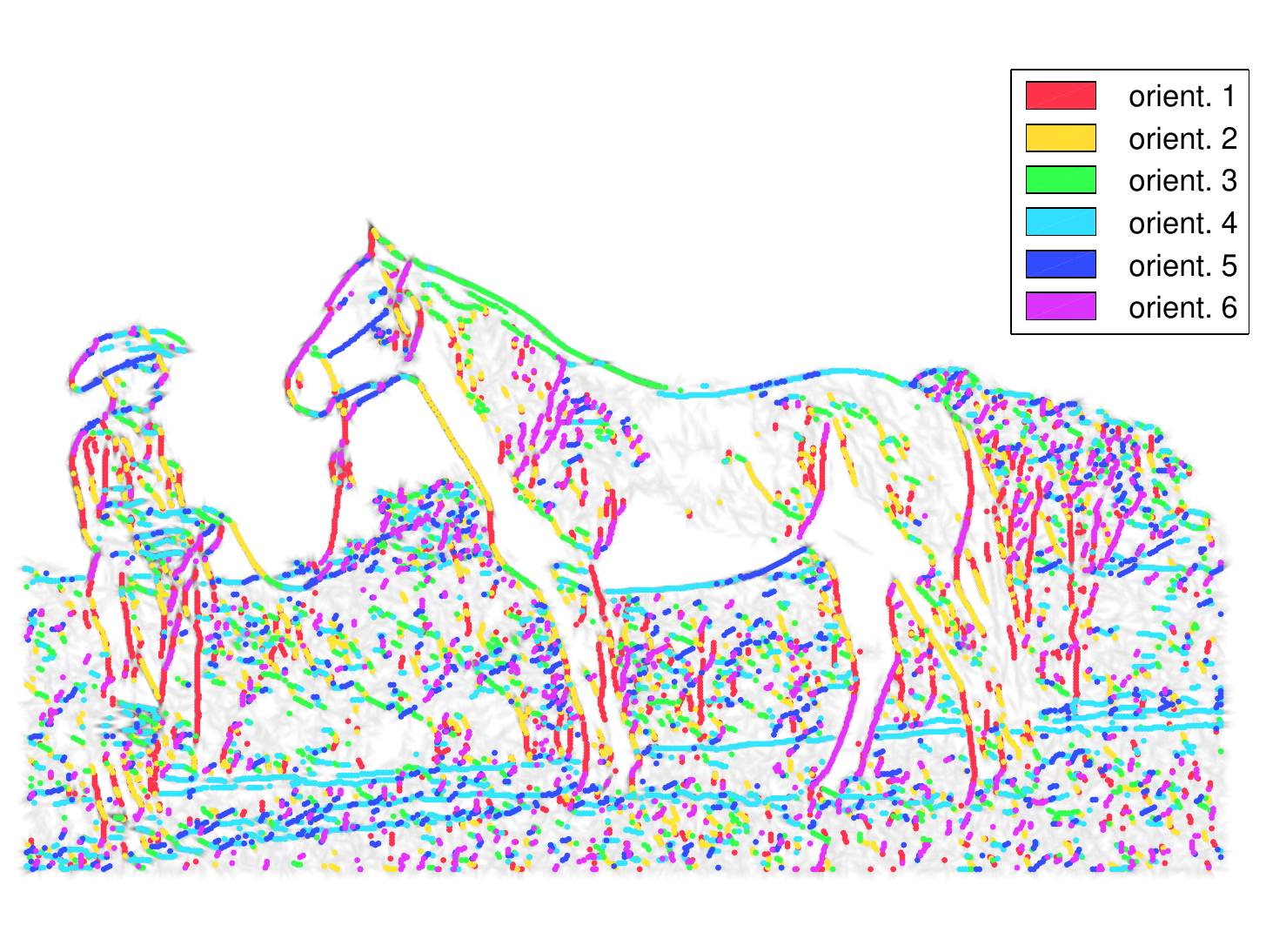}
\caption[Extracting oriented line fragments in an image]{{\bf Left}:
Original image. {\bf Middle}: Maximum over orientation energies,
$\argmax_{\psi} \mathcal E(x,y,\psi)$. {\bf Right}: Points in which
$n$-dimensional Gabor feature vectors $\mathbf f$ are extracted.
Different colors denote different dominant orientations of features
$\mathbf f$.}
 \label{fig:filtered_image}
\end{figure}

\emph{Scale.} The feature extraction as well as the recognition
process is performed at several scales of an image: we use a
Gaussian pyramid with two scales per octave and process each scale
separately. That is, feature extraction and recognition up to the
object level $\mathcal O$ are performed at every scale and
independently of other scales. For detection of object classes we
then consider information from \emph{all} scales. 
Not to overload the notation, we
will just refer to one scale of $I$.

\subsection{Inference}
\label{sec:inference}

Performing inference in a query image with a given vocabulary
entails inferring a hierarchy of hidden states from the observations
(which are the image features $(\mathbf F,\mathbf X)$). The hidden
states at the first layer will be the only ones receiving input
directly from the observations, whereas all the other states will
receive input from a hidden layer below.

Each hidden state $z^\ell = (\model^\ell, x^\ell)$ has two components: $\model^{\ell}$ denotes a composition from the
vocabulary and $x^\ell$ a location in an image (to abbreviate
notation from here on, we will use $x$ instead of $(x,y)$ to denote
the location vector). 
The notation $x^{\ell}$ is used to emphasize
that $x$ is from a spatial grid with resolution corresponding to
hidden layer $\ell$, i.e., the spatial resolution of the locations
of the hidden states will be
increasingly coarser with the level of hierarchy (discussed in Sec.~\ref{sec:effcomp}). 

To each hidden state $z^\ell$ we associate its \emph{score}, $\score(z^\ell)$. The score captures the deformations and compatibility of the parts.
It is recursively computed as follows:
\begin{align} \label{eq:scoredef}
\score(z^\ell)=\prod_{p=1}^{P(\model^\ell)} \max_{z_p^{\ell-1}} \Big(&{\scorehat}(z_p^{\ell-1})  \cdot D(x_j^{\ell-1}-
x^{\ell}\mid
\mu_j^{\ell},\Sigma_j^{\ell})\notag\\ 
&\cdot \mathrm{Comp}(\model^\ell,\modelm_j^{\ell-1},p)\Big),
\end{align}
where, in general, ${\scorehat}(z_p^{\ell-1}):={\score}(z_p^{\ell-1})$, and $D$ represents the deformation score function. We define it as an unnormalized Gaussian
\begin{equation} \label{eq:Ddef}
D(x\mid\mu,\Sigma) = \exp\big(-0.5\cdot(x-\mu)^T\Sigma^{-1}(x-\mu)\big)
\end{equation}
Note that $D$ is always from the interval $[0,1]$, and is $1$ when the part $z^{\ell-1}$ is located at its expected relative position $\mu$. The term $\mathrm{Comp}(\model^\ell,\modelm_j^{\ell-1},p)$ represents
the compatibility between the $p$-th part of $\model^\ell$ and composition $\modelm_j^{\ell-1}$. We use
$\mathrm{Comp}(\model^\ell,\modelm_j^{\ell-1},p) = \appearance_{p}^\ell(j)$. Our scores will always be from $[0,1]$.

Note that a composition can also have repulsive parts. For these we take $D$ to equal a constant $\alpha$ (which we set to
$0.1$ in this paper) and we define ${\scorehat}(z_p^{\ell-1}):=1-{\score}(z_p^{\ell-1})$. 
As an example, assume we have in our vocabulary a composition of three short horizontal edges as well as a composition of just two. 
The scoring function is set up in a way to prefer the larger composition if the image (local patch in the image) contains three strong horizontal edges while
it will prefer the smaller composition if one of the edges is much weaker. 

Score $\score(\modelm^\ell,x)$ of hidden state corresponding to an OR-composition is taken simply as maximum over all OR-ed compositions
\begin{equation} \label{eq:scoremdef}
\score(\modelm^\ell,x) = \max_{\model^\ell \in \modelm^\ell} \score(\model^\ell, x)
\end{equation}

Due to recursive definition of the score and the fact that each composition is a tree we can use dynamic programming to compute the scores efficiently. Note however, that since the number
of compositions in the vocabulary can be relatively large (order of hundreds), exhaustive computation would run too slow. We thus perform
hypothesis pruning at each level and focus computation to only promising parts of the image. We explain this process in more detail 
in the next section.

\subsection{Efficient computation}
\label{sec:effcomp}

The process of inference in image $I$ involves building an \emph{inference graph}
$\mathcal G=\mathcal G_I=(Z,E)$ where the nodes $Z$ of the graph are the (``promising'')
hypotheses $z^\ell$. Graph $\mathcal G$ has
a hierarchical structure resembling that of the vocabulary, thus vertices $Z$ are partitioned into
vertex layers $1$ to $\cO$, that is, $Z=Z^1\cup \bar{Z}^1 \cup Z^2\cup \bar{Z}^2 \cdots \cup
Z^\cO$. Vertices $Z^\ell$ correspond to composition layer $\Model^\ell$ and vertices of $\bar{Z}^\ell$ to 
OR layer $\Modelm^\ell$.

Assume that the hypotheses up to layer $\ell-1$, $\bar{Z}^\ell$ have already been computed. To get the next layer
we visit each hypothesis $\bar{z}_i^{\ell-1}$. For a particular $\bar{z}_i^{\ell-1} = (\modelm_i^{\ell - 1}, x^{\ell - 1})$ we first \emph{index} into the vocabulary to retrieve
potential compositions $\model^\ell$ to be matched. As the candidate list we retrieve all $\model^\ell$ for which  
$\modelm_i^{\ell - 1}$ is one of its parts, i.e.\ $\appearance^\ell(i)$ is non-zero. This typically prunes the number of compositions
to be matched significantly. 

The matching process for a retrieved composition $\model^\ell$ entails computing the score defined in Eq.~\eqref{eq:scoredef}. If the hypothesis scores higher
than a threshold $\tau^\ell$ we add it to $Z^{\ell-1}$, otherwise we prune it.  During the process of training (until the hierarchy is built up to
the level of objects), these thresholds are fixed and set to very
small values (in this paper we use $\tau=0.05$). Their only role is
to prune the very unlikely hypotheses and thus both, speed-up
computation as well as minimize the memory/storage requirements (the
size of $\mathcal G$). After the class models (the final, layer
$\Model^C$ of the vocabulary) are learned we can also learn the
thresholds in a way to optimize the speed of inference while
retaining the detection accuracy, as will be explained in
Sec.~\ref{sec:thresholds}. Note that due to the thresholds, we can avoid some of the computation: we do not need to consider
the spatial locations of parts that are outside the $\tau^\ell$ Gaussian radius. 

\emph{Reduction in spatial resolution.} 
After each layer $Z^{\ell}$ is built, we perform \emph{spatial
downsampling}, where the locations $x^{\ell}$ of the hidden states
$z^{\ell}$ are downsampled by a factor $\rho^\ell < 1$. Among the
states that code the same composition $\model^{\ell}$ and for which
the locations (taken to have integer values) become
equal, only the state with the highest score is kept. 
We use $\rho^1=1$ and $\rho^{\ell} = 0.5$, $\ell>1$. This step is
mainly meant to reduce the computational load: by reducing the
spatial resolution at each layer, we bring the far-away
(location-wise) hidden states closer. This, indirectly, keeps the scales of the Gaussians relatively small across
all compositions in the vocabulary and makes inference faster.

The OR layer $\bar{Z}^\ell$ pools all compositions $z^\ell  = (\omega^\ell, x^\ell) \in Z^\ell$ for which 
$\{\bar{z}^\ell = (\modelm^\ell, x^\ell) : \model^\ell \in\modelm^\ell \}$ into one node, which further reduces the hypotheses space.

Edges of $\mathcal{G}$ connect vertices of $Z^\ell$ to vertices of $\bar{Z}^{\ell - 1}$ and vertices of $\bar{Z}^{\ell}$ to 
vertices of $Z^{\ell}$. Each $z^\ell = (\model^\ell, x^\ell)$ is connected to vertices $\bar{z}_p^{\ell - 1}$ that yield the max
score for the part $j$ in Eq.~\eqref{eq:scoredef} and each vertex  $\bar{z}^\ell = (\modelm^\ell, x^\ell)$ connects to $z^\ell$ giving
the max value in Eq.~\eqref{eq:scoremdef}. The subgraph of $\mathcal{G}$ on vertices reachable from vertex $z$ via downward edge paths is called a
\emph{parse graph} and denoted by $\mathcal{P}(z)$. A set of vertices of $\mathcal{P}(z)$ from $Z^1$, i.e.\ layer 1 vertices reachable from $z$ through edges in $\mathcal G$, are called \emph{support} of $z$ and denoted with $\supp(z)$. See Fig.~\ref{fig:sizerec} for examples of parse graphs and supports of vertices from different layers. 
The definition of support is naturally extended to a set of nodes as $\supp(S) = \cup_{z \in S} \supp(z)$.

\section{Learning}
\label{sec:learning}

This section addresses the problem of learning the representation,
which entails the following:
\begin{enumerate}
\item {\bf Learning the vocabulary of compositions.} Finding an appropriate set of compositions to represent our data,
learning the structure of compositions (the number of
parts and a (rough) initialization of the parameters) and learning the OR-compositions.
\item {\bf Learning the parameters of the representation.} Finding
the parameters for spatial relations and appearance for each composition as
well as the thresholds to be used in the approximate inference
algorithm.
\end{enumerate}

In the learning process our aim is to build a vocabulary of compositions which tries to meet the following two conditions: 
(a) Detections of compositions on the object layer, $\Omega^\mathcal{O}$, give good performance on the train/validation set.
(b) The vocabulary is minimal, optimizing the inference time.
Intuitively, it is possible to build a good composition on the object layer if the previous layer $\Model^{\mathcal{O} - 1}$ produces a good ``coverage'' of training examples. By coverage we mean that the supports of hypotheses in $Z^{\mathcal{O} - 1}$ contain most of the feature locations $\bf X$. Hence, we would like to build our vocabulary such that the compositions at each layer cover the feature location $\bf X$ in the training set sufficiently well and is, according to (b), the most compact representation. 

To find a good vocabulary according to the above constraints is a hard problem. Our strategy is to learn one layer at a time, in a bottom up fashion, and at each layer learn a small compact vocabulary while maintaining its good coverage ratio. Bottom-up layer learning is also the main strategy adopted to learn the convolutional networks~\cite{s:lecun07,s:hinton07}.

We emphasize that our learning algorithm uses several heuristics and is not guaranteed to find the optimal vocabulary. Deriving a more mathematically elegant learning approach is part of our future work.

\begin{figure*}[t!]
\centering
\includegraphics[width=\linewidth]{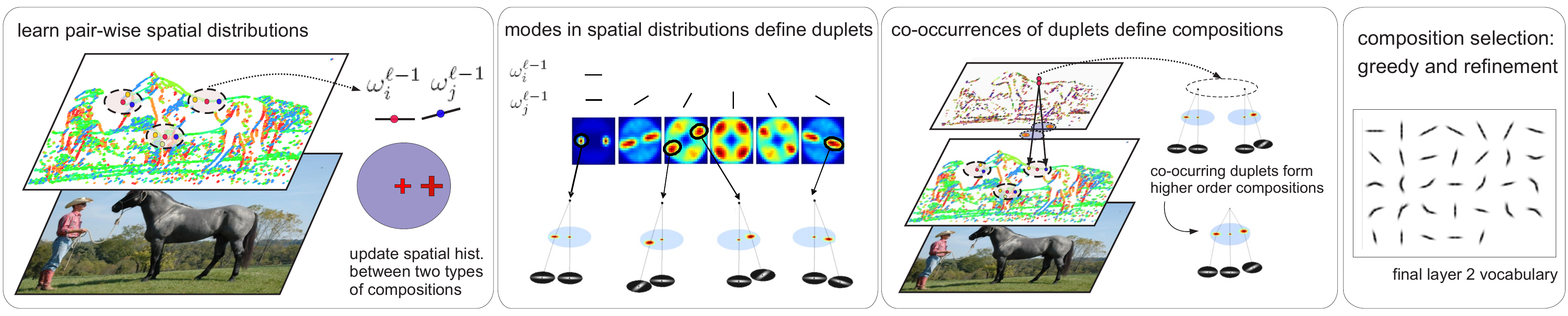}
\caption{Approach to learning the structure and the parameters of
the vocabulary at a particular layer.} \label{fig:learning}
\end{figure*}

\subsection{Learning the structure}
\label{sec:structure}

Here we explain how we learn the initial set of compositions, their structure and initial 
values of the parameters (which are re-estimated as described in Sec.~\ref{sec:parameters}).

The first layer, $\ell=1$, is fixed thus the first layer that we learn is $\ell=2$.  Since learning the vocabulary is recursive we will describe how to learn layer $\ell$ from $\ell-1$. We will assume that for each training image $I$ we have the
inference graph $\mathcal G_I=(Z,E)$ built
up to layer $\ell-1$, which is obtained as described in
Sec.~\ref{sec:inference}.

Define a spatial neighborhood of radius $r^\ell$ around the node $z^{\ell-1}=(\modelm^{\ell-1},x^{\ell-1})$ with $\mathcal N(z^{\ell-1})$. We also define the \emph{support} of $\mathcal N(z^{\ell-1})$ by re-projecting the circular area to the original image resolution and finding all image features whose locations fall inside the projected circular area. Our goal in learning is to find a vocabulary that best explains all of the spatial neighborhoods $\mathcal N(z^{\ell-1})$, where $z^{\ell-1}\in Z^{\ell-1}$:

\begin{align}
\label{eq:learning}
\Model_*^\ell = \argmax_{\Model^\ell}\  \Big(&-C\sum_{\model_i^\ell\in \Model^\ell} P(\model_i^\ell) + \\
&+\sum_{I_j} \sum_{k\, \mid \, z_k^{\ell-1}\in Z_{I_j}^{\ell-1}} \loss(\mathcal N(z_k^{\ell-1}),\Model^\ell)\Big)\nonumber
\end{align}

Here $P(\model_i^\ell)$ is the number of parts that defines a composition $\model_i^\ell$, and $C$ is a weight that balances the two terms. We define the score term as follows:
\begin{align}
\label{eq:loss}
\loss(\mathcal N(z_k^{\ell-1}),\Model^\ell)=\max_{z^\ell} \big(&\loss_{coverage}(\mathcal N(z_k^{\ell-1}), z^\ell)\, +\\
&\loss_{tree}(z^\ell)\big),\nonumber
\end{align}
where
\begin{equation}
\label{eq:loss_coverage}
\loss_{coverage}(\mathcal N(z_k^{\ell-1}), z^\ell) = \frac{|\supp(z^\ell)\cap \supp(\mathcal N(z_k^{\ell-1}))|}{|\supp(z^\ell)\cup\supp(\mathcal N(z_k^{\ell-1}))|}
\end{equation}
and
\begin{equation}
\label{eq:loss_tree}
\loss_{tree}(z^\ell) = \begin{cases} -\infty\quad \text{if supports of any two pairs of}\\ \ \ \quad\quad \text{parts of } z^\ell \text{ overlap more than }  0.2\\0 \,\qquad\text{otherwise}\end{cases}
\end{equation}

Our learning strategy will be the following:
\begin{enumerate}
\item First learn the relative spatial relations between all possible pairs of compositions from layer $\ell-1$.
\item Detect the modes in these distributions which will define two-part compositions called \emph{duplets}.
\item Find a set of all compositions, where each composition is a set of (frequently) co-occurring duplets that form a tree.
\item Select a subset of compositions that tries to optimize the scoring function $\eqref{eq:learning}$.
\item Learn the OR layer by finding clusters of compositions that detect similar shapes.
\end{enumerate}

The overall learning procedure is illustrated in
Fig.~\ref{fig:learning}.

\subsubsection{Learning spatial correlations between parts}
\label{sec:spatial}

We start by learning the spatial relations among 
\emph{pairs} of compositions
$(\modelm_i^{\ell-1},\modelm_j^{\ell-1})\in \Modelm^{\ell-1}\times
\Modelm^{\ell-1}$. The first composition $\modelm_i^{\ell-1}$ in the
pair plays the role of a \emph{reference}. This means that the
spatial relations will be learned relative to it. We will learn these relations by visiting
all pairs of states in the inference graphs $\bar Z^{\ell-1}$ that are within $r^\ell$ distance of each other (the choice of $r^\ell$ is described below). We learn the
spatial relations by means of two-dimensional histograms
$h_{ij}$ of size $[-r^\ell,r^\ell]\times[-r^\ell,r^\ell]$.

During training, each histogram $h_{ij}^\ell$ is updated at location $x'^{\ell-1} - x^{\ell-1}$ for each pair
of hidden states $(z^{\ell-1},z'^{\ell-1})$, where $z^{\ell-1}=(\modelm_i^{\ell-1},x^{\ell-1})$ and
$z'^{\ell-1}=(\modelm_j^{\ell-1},x'^{\ell-1})$, satisfying:
\begin{enumerate}
\item $|x^{\ell-1}-x'^{\ell-1}| \leq r^\ell$, that is, $z'^{\ell-1}$ lies in the circular area around $z$ with radius $r^\ell$.

\item $z^{\ell-1}$ and $z'^{\ell-1}$ have disjoint supports. We enforce this to ensure that the resulting composition will have a tree topology, i.e.\ parts
in each composition do not overlap spatially (or overlap minimally). The overlap of the supports is calculated as: 

$o(z^{\ell-1}, z'^{\ell-1})=\tfrac{|\supp(z^{\ell-1})\cap \supp(z'^{\ell-1})|}{|\supp(z^{\ell-1})\cup \supp(z'^{\ell-1})|}$

We
allow for a small overlap of the parts, i.e., we select $(z^{\ell-1},z'^{\ell-1})$ for which $o(z^{\ell-1},z'^{\ell-1})<0.2$.
\end{enumerate}
The histograms are updated for all graphs $\mathcal G_I$
and all the admissible pairs of hypotheses at layer $\bar Z^{\ell-1}$.

\begin{figure}[t!]
\centering
\includegraphics[width=0.174\linewidth,trim=90 28 94 27,clip=true]{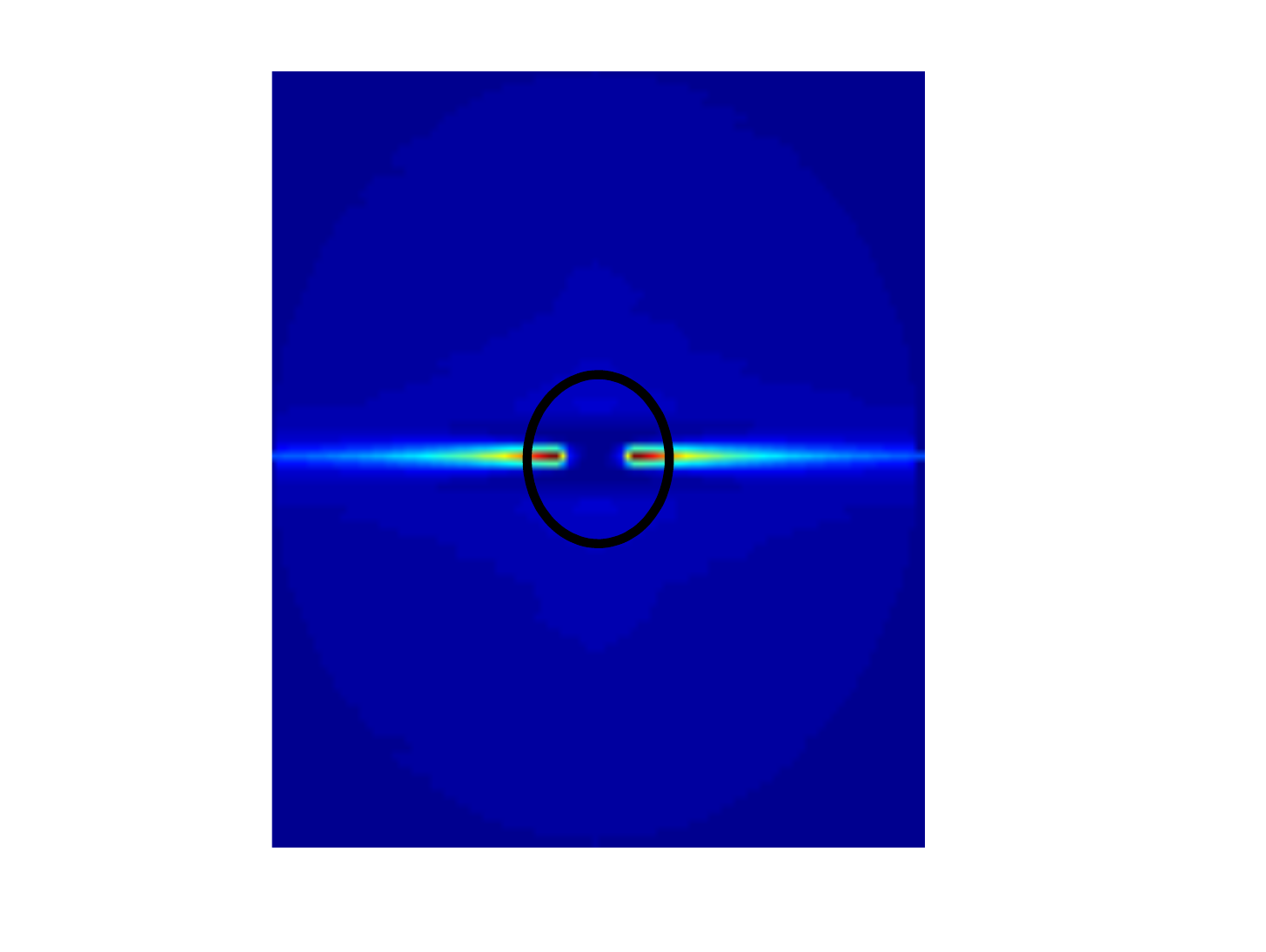}\hspace{-1.8mm}
\includegraphics[width=0.174\linewidth,trim=90 28 94 27,clip=true]{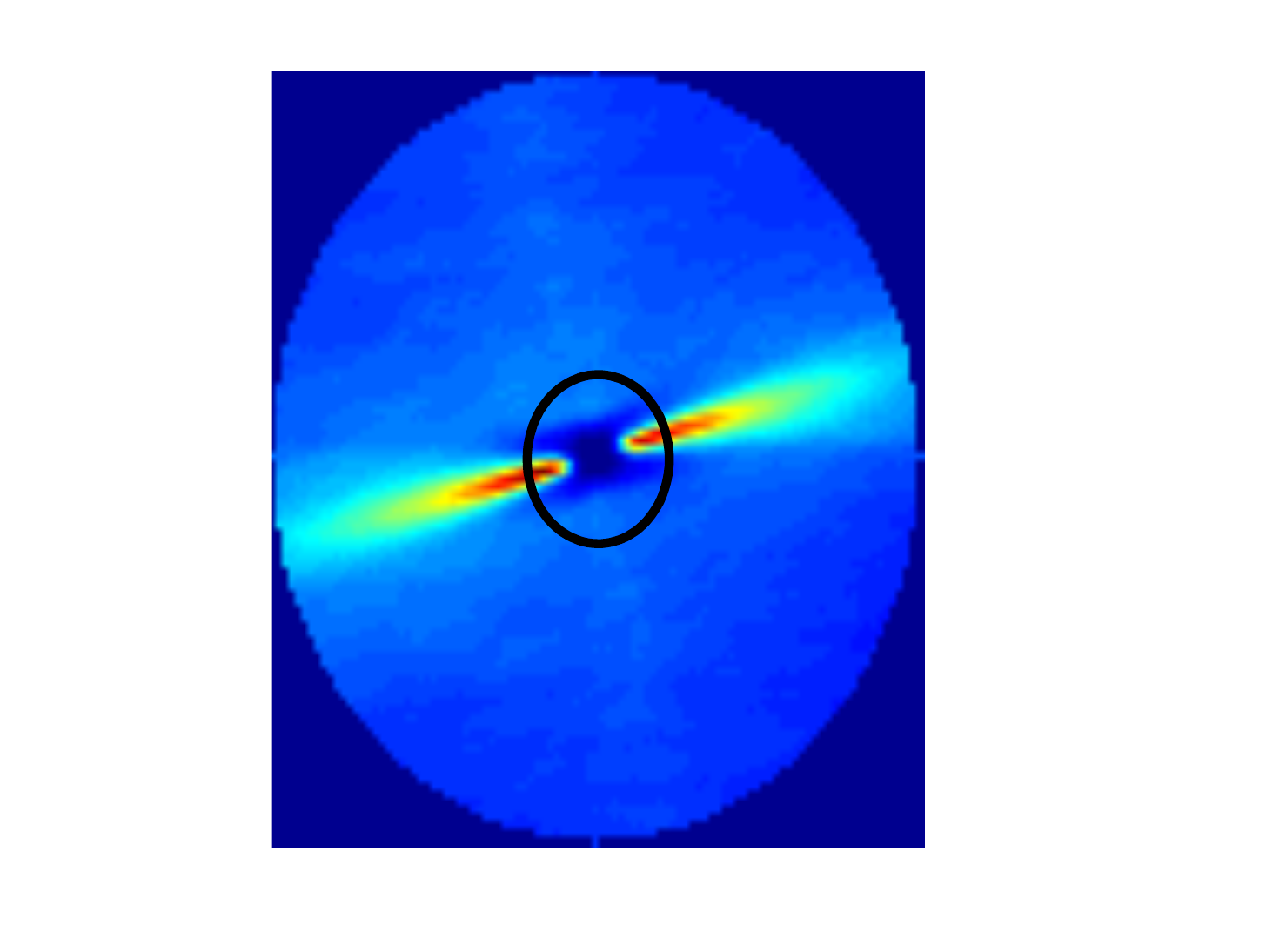}\hspace{-1.8mm}
\includegraphics[width=0.174\linewidth,trim=90 28 94 27,clip=true]{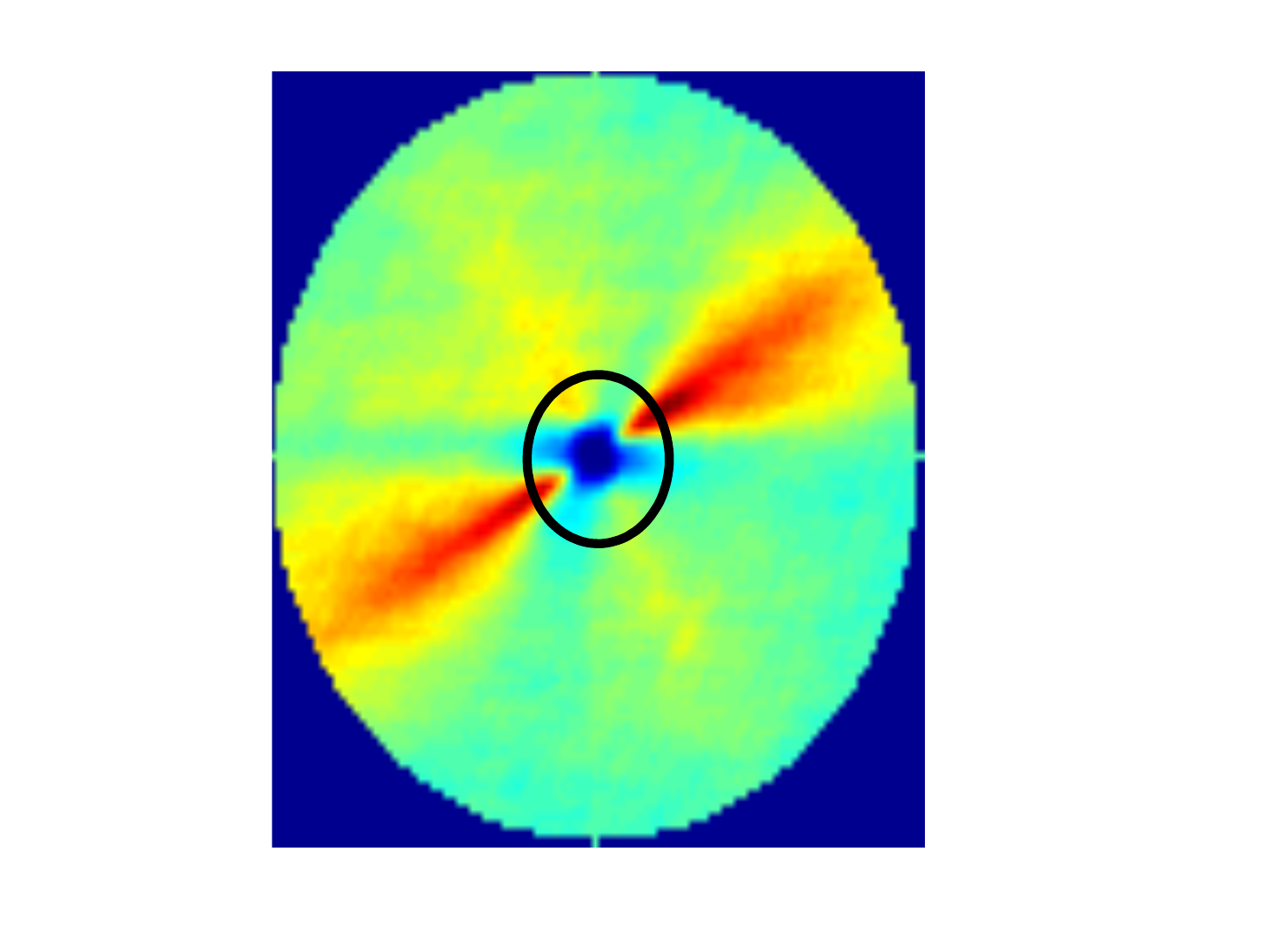}\hspace{-1.8mm}
\includegraphics[width=0.174\linewidth,trim=90 28 94 27,clip=true]{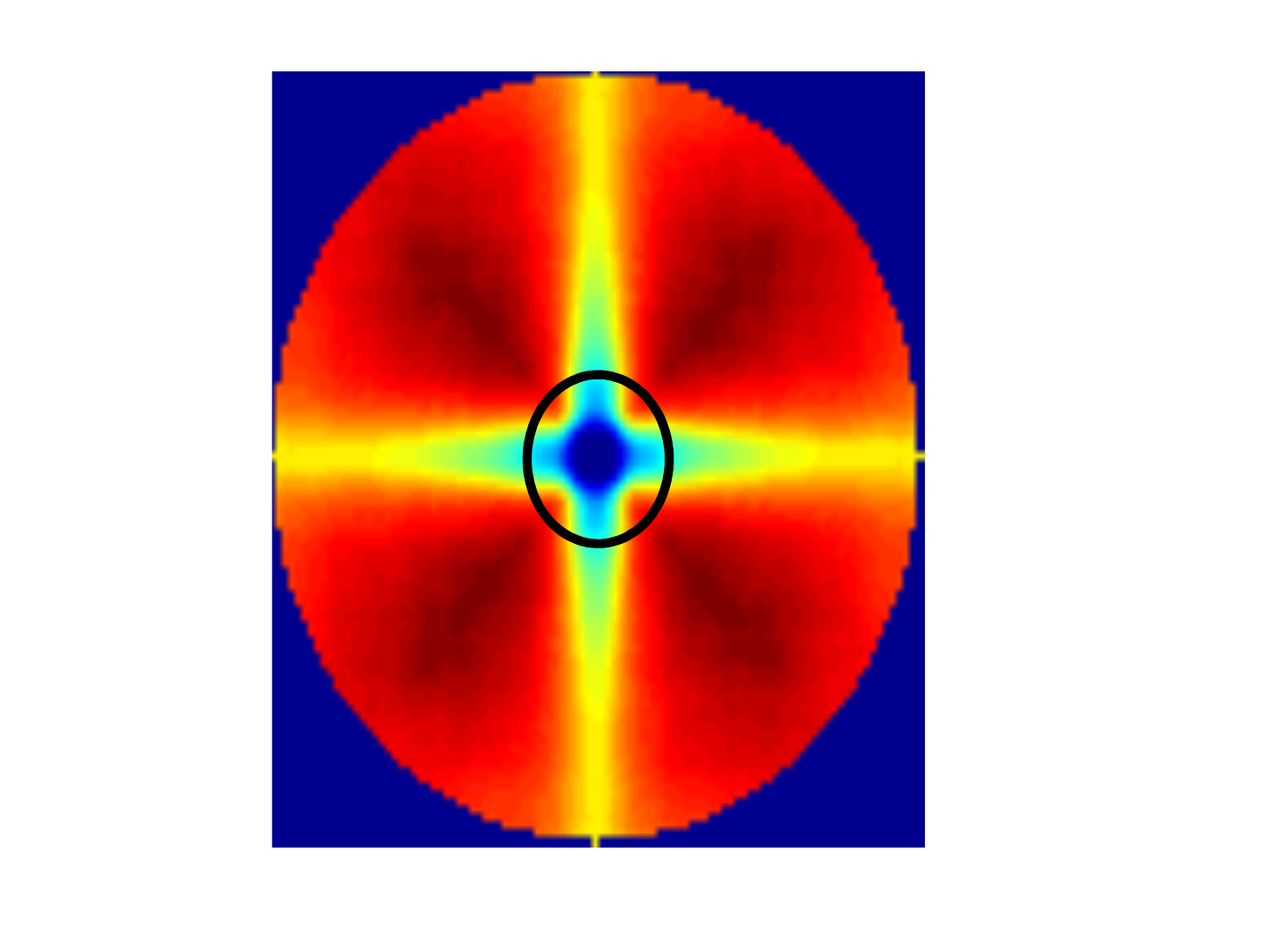}\hspace{-1.8mm}
\includegraphics[width=0.174\linewidth,trim=90 28 94 27,clip=true]{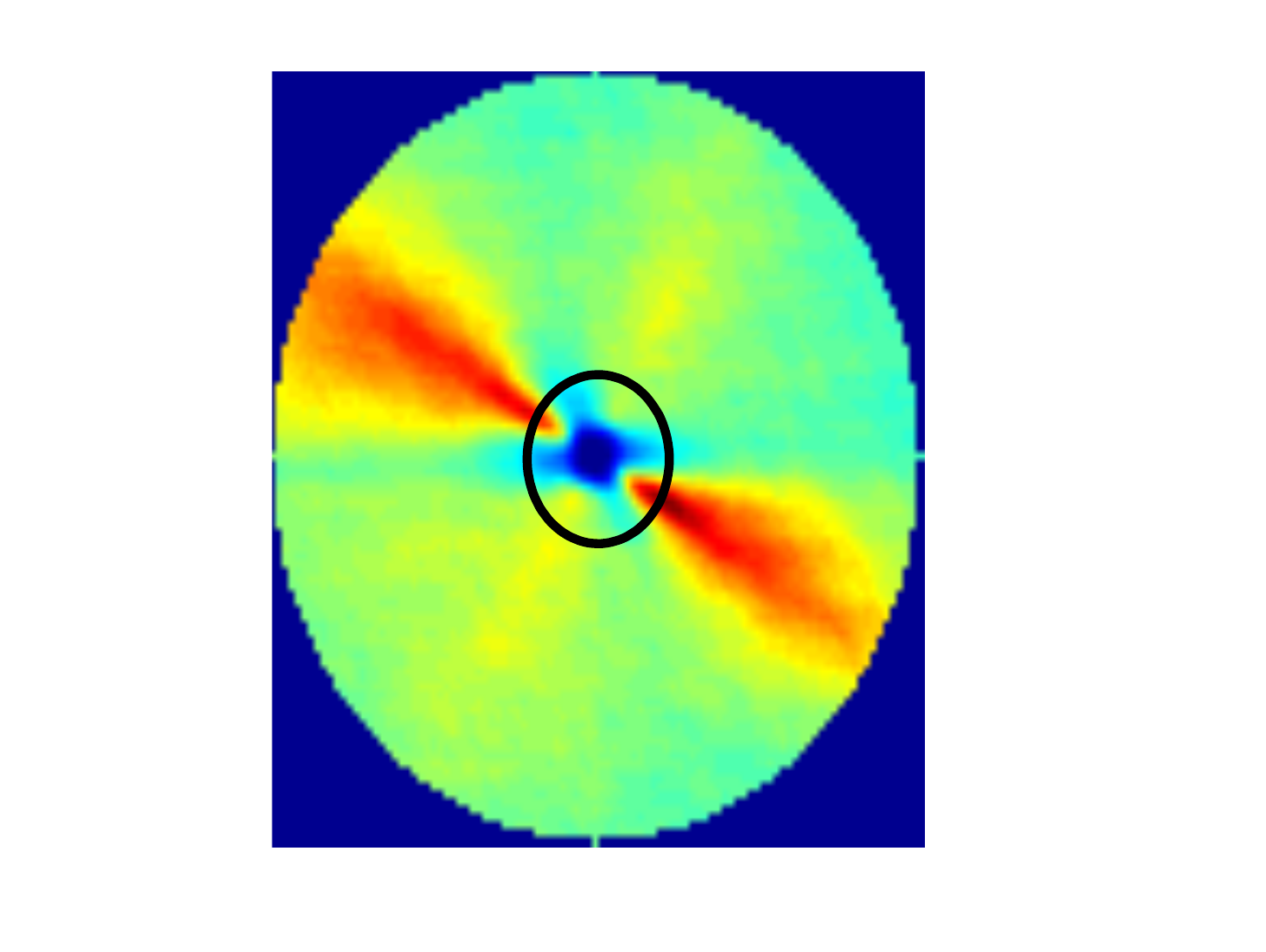}\hspace{-1.8mm}
\includegraphics[width=0.174\linewidth,trim=90 28 94 27,clip=true]{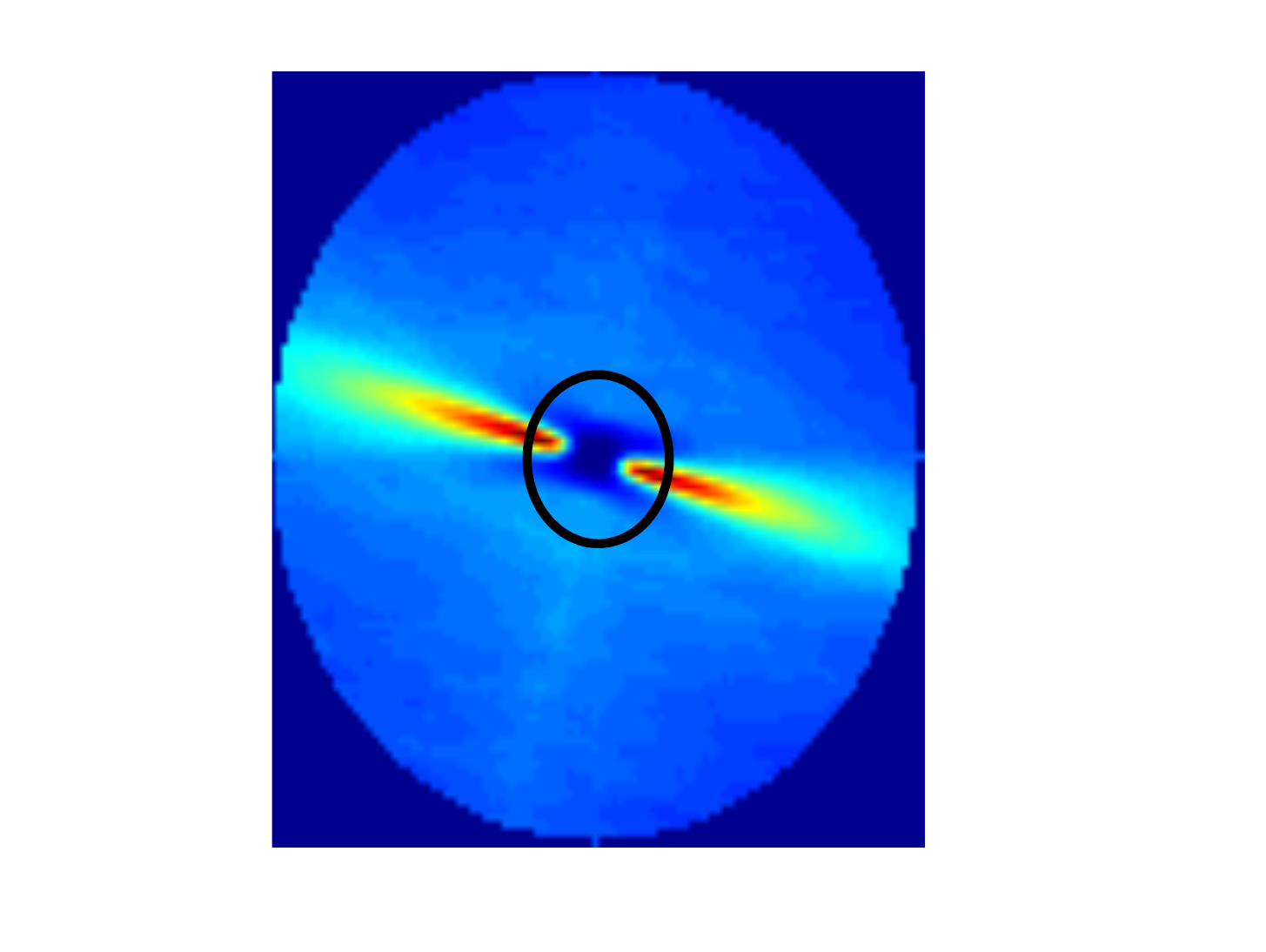}\hspace{-1.8mm}
\caption{{\small Examples of learned histograms $h_{ij}^{\ell=2}$
for very large radius, $r^\ell=50$. The circle shows the chosen
radius, $r^\ell=8$.}}
 \label{fig:size}
\end{figure}

\emph{The choice of $r^\ell$.} The correlation between the relative
positions of the hypotheses is the highest at relatively small
distances as depicted in Fig.~\ref{fig:size}. In this paper we set the radii $r^\ell$ to $8$ for layer $2$, $12$ for the higher layers, and $15$
for the final, object layer, but this choice in general depends on the
factors of the spatial downsampling (described in Sec.~\ref{sec:effcomp}). Note that the circular regions are
defined at spatial resolution of layer $\ell-1$ which means that the
area, if projected back to the original resolution (taking into account the downsampling factors), covers a much
larger area in an image.

\emph{Finding the initial vocabulary: {\bf duplets}}. From the computed pair-wise histograms we form
\emph{duplets} by finding the modes of the spatial distributions. Fig.~\ref{fig:maps_sizes}
illustrates the clustering procedure. For each mode we estimate the
mean $\mu^\ell$ and variance $\Sigma^\ell$ by fitting a Gaussian
distribution around it. Each of these modes thus results in a
two-part composition with initial parameters: $P=2$,
$\appearance_1^\ell(\model_i^\ell)=1$,
$\appearance_2^\ell(\model_j^\ell)=1$ and
$\spatial_2^{\ell}=(\mu^\ell,\Sigma^\ell)$ (the first, reference
part, in a composition is always assigned the default parameters as
explained in Sec.~\ref{sec:representation}). For the higher layers,
$\ell > 3$, where the co-occurrences are sparser, we additionally
smooth the histograms prior to clustering, to regularize the
data.

\subsubsection{From duplets to compositions}
\label{sec:compositions}

Once we have the duplets our goal is to combine them into possibly higher-order (multi-part) \emph{compositions}. We can form a composition whenever two or more different duplets share the same part. For example, we can combine a duplet defined by 
$\appearance_1^\ell(i)=1$,
$\appearance_2^\ell(j)=1$ and
$\spatial_2^{\ell}=(\mu_1^\ell,\Sigma_1^\ell)$, and a duplet defined by $\appearance_1^\ell(i)=1$,
$\appearance_2^\ell(k)=1$ and
$\spatial_2^{\ell}=(\mu_2^\ell,\Sigma_2^\ell)$, into a composition with $P=3$, $\appearance_1^\ell(i)=1$,
$\appearance_2^\ell(j)=1$, $\appearance_3^\ell(k)=1$, and $\spatial_2^{\ell}=(\mu_1^\ell,\Sigma_1^\ell)$, $\spatial_3^{\ell}=(\mu_2^\ell,\Sigma_2^\ell)$. We will also allow a composition to only consist of a single duplet. We have a very loose restriction on the upper bound of the number of parts, i.e., we do not consider compositions of more than $P=10$ parts.
We find the compositions by
matching the duplets to each of the neighborhoods and keeping count of co-occurring duplets. Since each duplet is defined as one part spatially related to another, reference, part, we will only count co-occurrences whenever two or more duplets share the same reference part (state $z_1^{\ell-1}$ in this case). We next describe this procedure in more detail.

We first perform inference with our vocabulary of duplets. We match the duplets in each neighborhood $\mathcal N(z_k^{\ell-1})$
by restricting that the matched reference part for each duplet  to be $z_k^{\ell-1}$ and match the other part of the duplet to the remaining states whose locations fall inside $\mathcal N(z_k^{\ell-1})$. We keep all matches for which the score computed as in Eq.\eqref{eq:scoredef} is above a threshold $\tau^\ell$. Denote a duplet match with $\hat z_{j}^\ell$ which has parts $z_k^{\ell-1}=(\bar\model_k^{\ell-1},x_k^{\ell-1})$ and $z_j^{\ell-1}=(\bar\model_j^{\ell-1},x_j^{\ell-1})$.

For each $\mathcal N(z_k^{\ell-1})$ we then find all subsets $z_j^\ell=\{\hat z_{j_p}^\ell\}_{p=1}^P$ of matched duplets
which form trees, i.e., overlap (measured as intersection over union) of supports between each pair $(z_{j_p}^{\ell-1}, z_{j_q}^{\ell-1})$ in the subset is lower than a threshold ($0.2$). Note that the reference part $z_k^{\ell-1}$ is common to all duplet matches (due to our restriction) and thus we are only checking the overlaps between the second parts of duplets in $z_j^\ell$. We compute $\loss_{coverage}(\mathcal N(z_k^{\ell-1}), z_j^\ell)$ for each $z_j^\ell$ as defined in Eq.~\eqref{eq:loss_coverage}.

Each $z_j^\ell=\{\hat z_{j_p}^\ell\}_{p=1}^P$ defines a higher-order composition $\model_j^\ell$ with $P+1$ parts, $\appearance_1^\ell(k)=1$, $\{\appearance_{p+1}^\ell(j_p)=1\}_{p=1}^P$, $\{\spatial_{p+1}^{\ell}=(\mu_{j_p}^\ell,\Sigma_{j_p}^\ell)\}_{p=1}^P$. We do not make a difference between different permutations of parts thus we always order $\{\model_{j_p}\}$ by increasing index values. 

We take a pass over all
neighborhoods $\mathcal N(z_k^{\ell-1})$ in all training images and keep track of all different $\model_j^\ell$ and update their score values. Note that some combinations of duplets will never occur, and is thus better to keep a sorted list of all $\model_j^\ell$ that occur at least once, and add to it if a new combination appears. Denote with $\Modelm_{0}^{\ell}$ the large set of compositions obtained in this step. In the next Subsec. we discuss how we select our  vocabulary $\Model^{\ell}$ from this large pool of compositions. 

\subsubsection{Composition selection}
\label{sec:selection}

The final vocabulary at layer $\ell$ is formed using a two-stage procedure. We first use a simple greedy selection: the composition $\model_j^\ell$ with the highest score (minus $C$ times its number of parts) is added to $\Model^\ell$ first. We next visit all neighborhoods $\mathcal N(z_k^{\ell-1})$ for which $\model_j^\ell$ was among the tree matches, and recompute the scores of all other matching tree compositions $z_{j'}^\ell$ as $\loss_{coverage}'(\mathcal N(z_k^{\ell-1}), z_{j'}^\ell)=\loss_{coverage}(\mathcal N(z_k^{\ell-1}), z_{j'}^\ell) -\loss_{coverage}(\mathcal N(z_k^{\ell-1}), z_j^\ell)$. We remove all $z_{j'}^\ell$ for which the new score is lower than $\epsilon$. We use the $\epsilon$ slack for efficiency -- the larger our current vocabulary is the less neighborhoods we need to update. We again select the composition with the highest new score and repeat the procedure. We stop when the highest score is lower than a threshold (set empirically).

We next employ a stochastic MCMC algorithm to get the final
vocabulary at layer $\ell$, $\Model^{\ell}$. The first state of the Markov chain is
the vocabulary $\Model_{g}^{\ell}$ obtained with the greedy
selection. Let $\Model_t^{\ell}$ denote the vocabulary at the
current state of the chain. 
At each step we either exchange/add/remove one
composition from $\Model_t^{\ell}$ with another one from
$\Model_{0}^{\ell}\setminus \Model_t^{\ell}$ to get the vocabulary
$\Model_{t+1}^{\ell}$. The vocabulary $\Model_{t+1}^{\ell}$ is
accepted as the next state of the Markov chain with probability
\begin{equation}
\min(1,\ \beta^{\loss(\Model_{t}^{\ell}) - \loss(\Model_{t+1}^{\ell})}),\ \ \beta > 1
\end{equation}
according to the Metropolis-Hastings algorithm.

The vocabulary at layer $\ell$ is finally defined as the
$\Model^{\ell}$ with maximal value of $\loss(\Model^\ell)$,
after running several iterations of the M-H algorithm. We usually
perform $100$ iterations. This vocabulary refinement step has shown to slightly improve results~\cite{s:fidler09a}.

\def\IH{1.48cm}
\def\IHB{2.35cm}
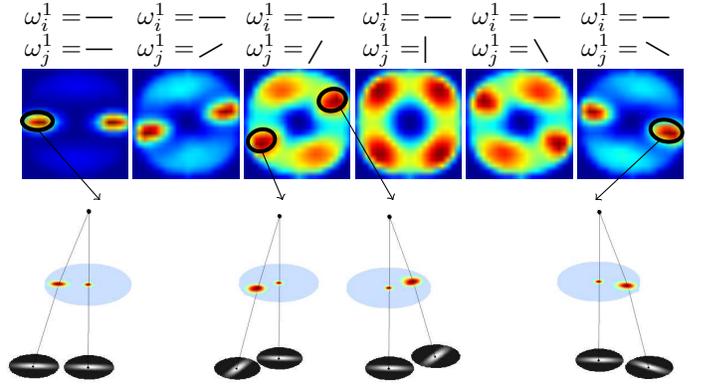
\begin{figure}[t!]
\centering
\begin{tikzpicture}[style=thick, scale=1]
\pgftext[at=\pgfpoint{0cm}{-0cm}]{ \hspace{-4.5mm}
\includegraphics[height=\IH,trim=92 30 85 30,clip=true]{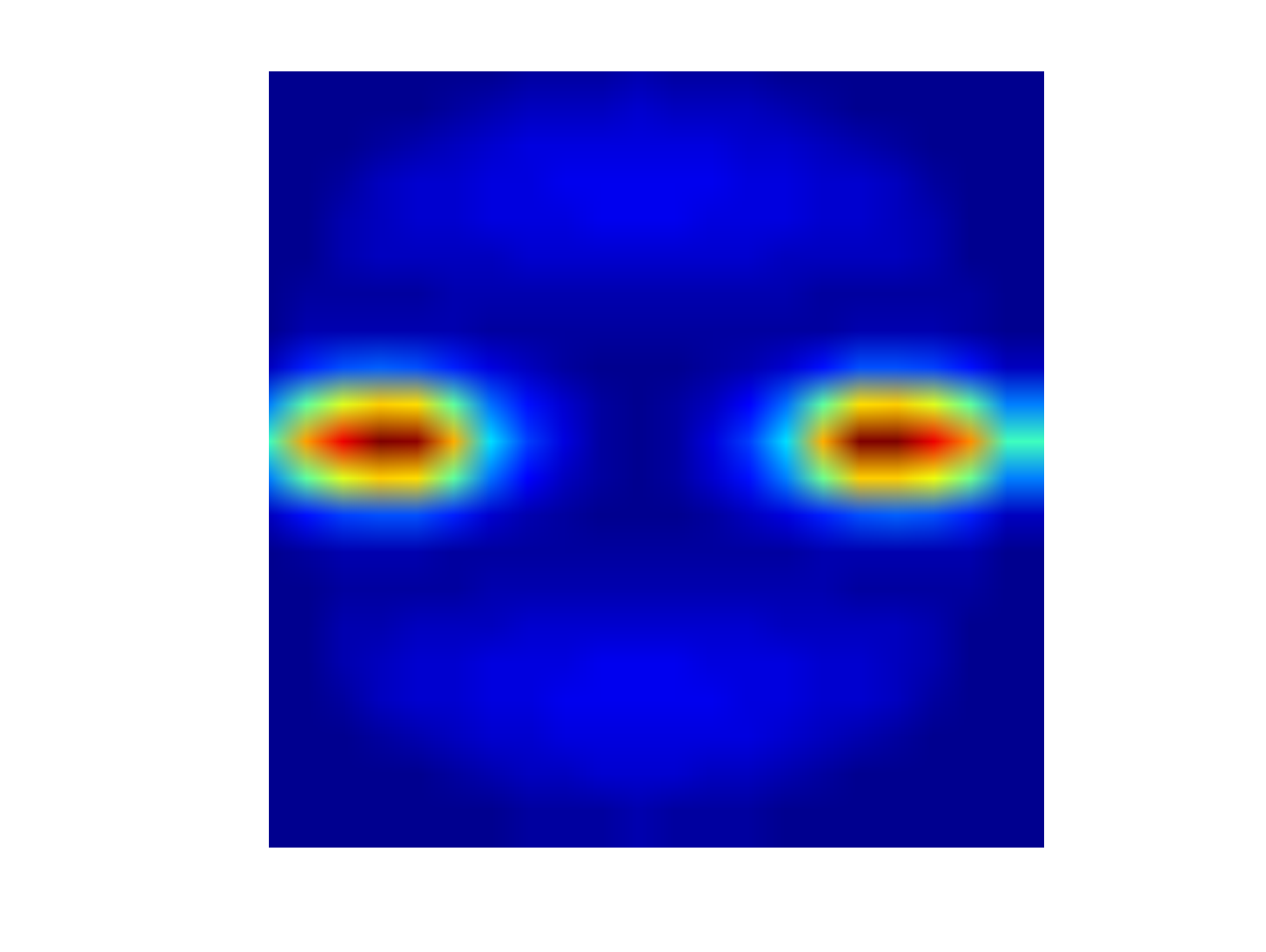}\hspace{-0.5mm}
\includegraphics[height=\IH,trim=92 30 85 30,clip=true]{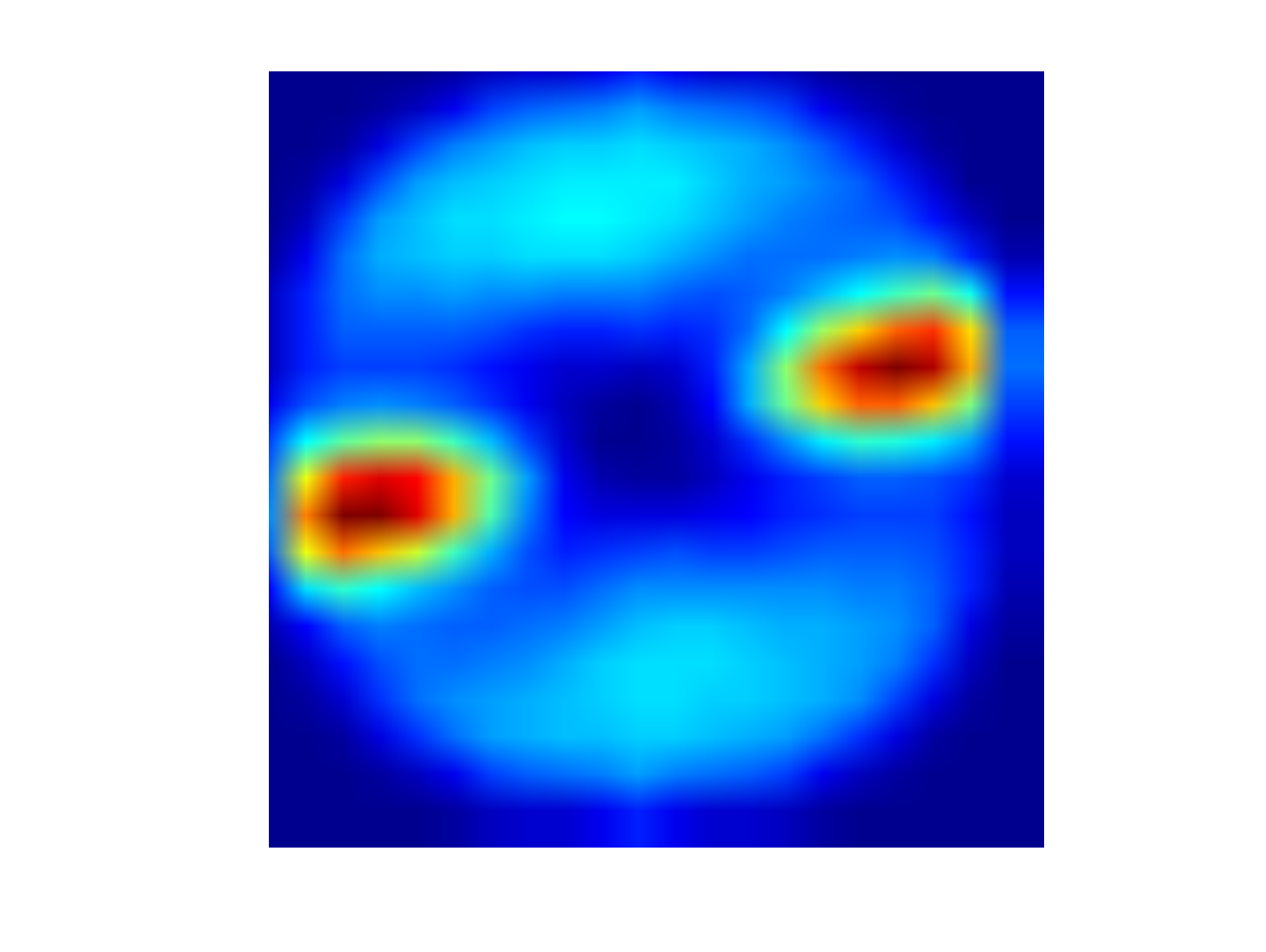}\hspace{-0.5mm}
\includegraphics[height=\IH,trim=92 30 85 30,clip=true]{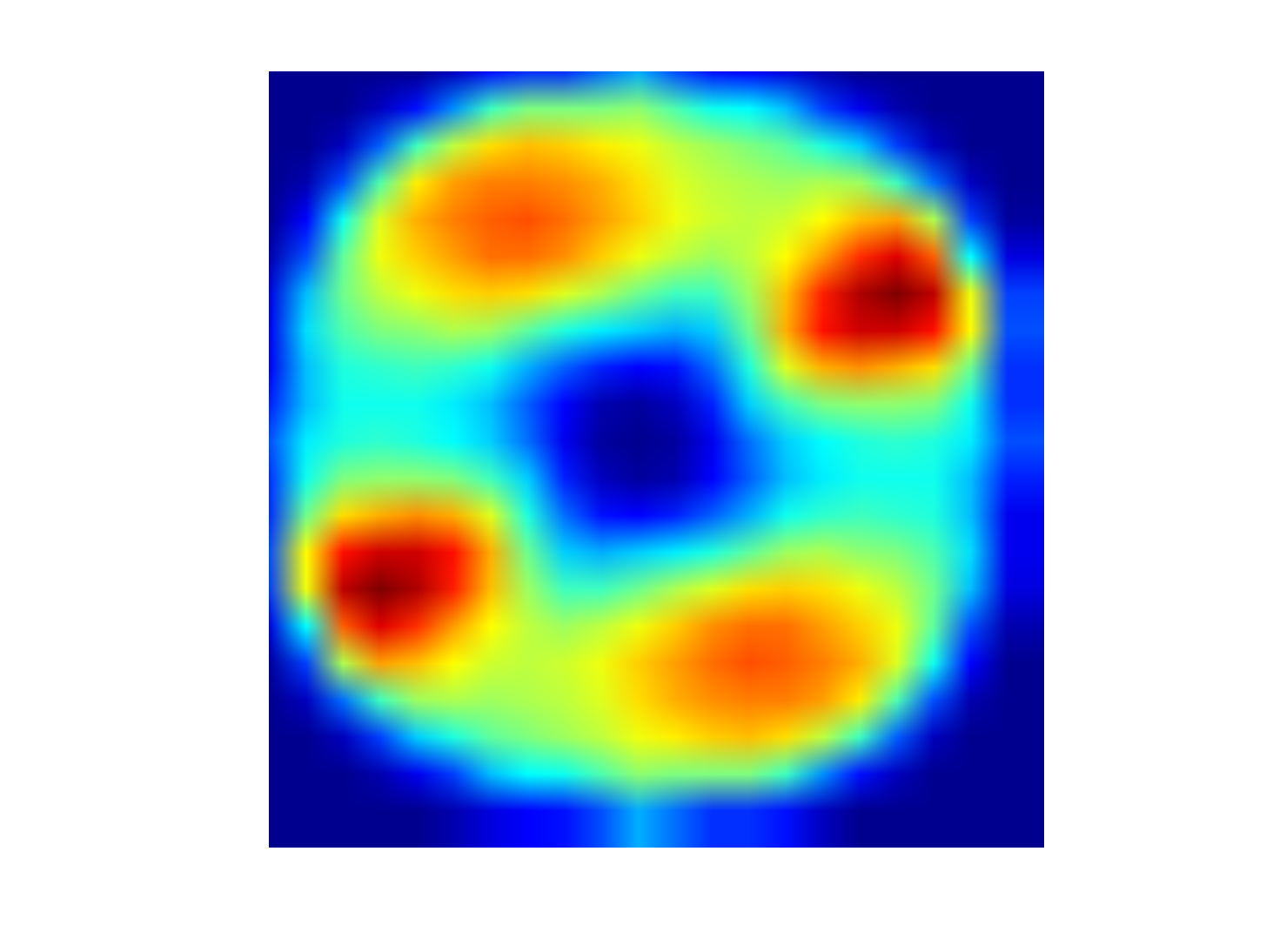}\hspace{-0.5mm}
\includegraphics[height=\IH,trim=92 30 85 30,clip=true]{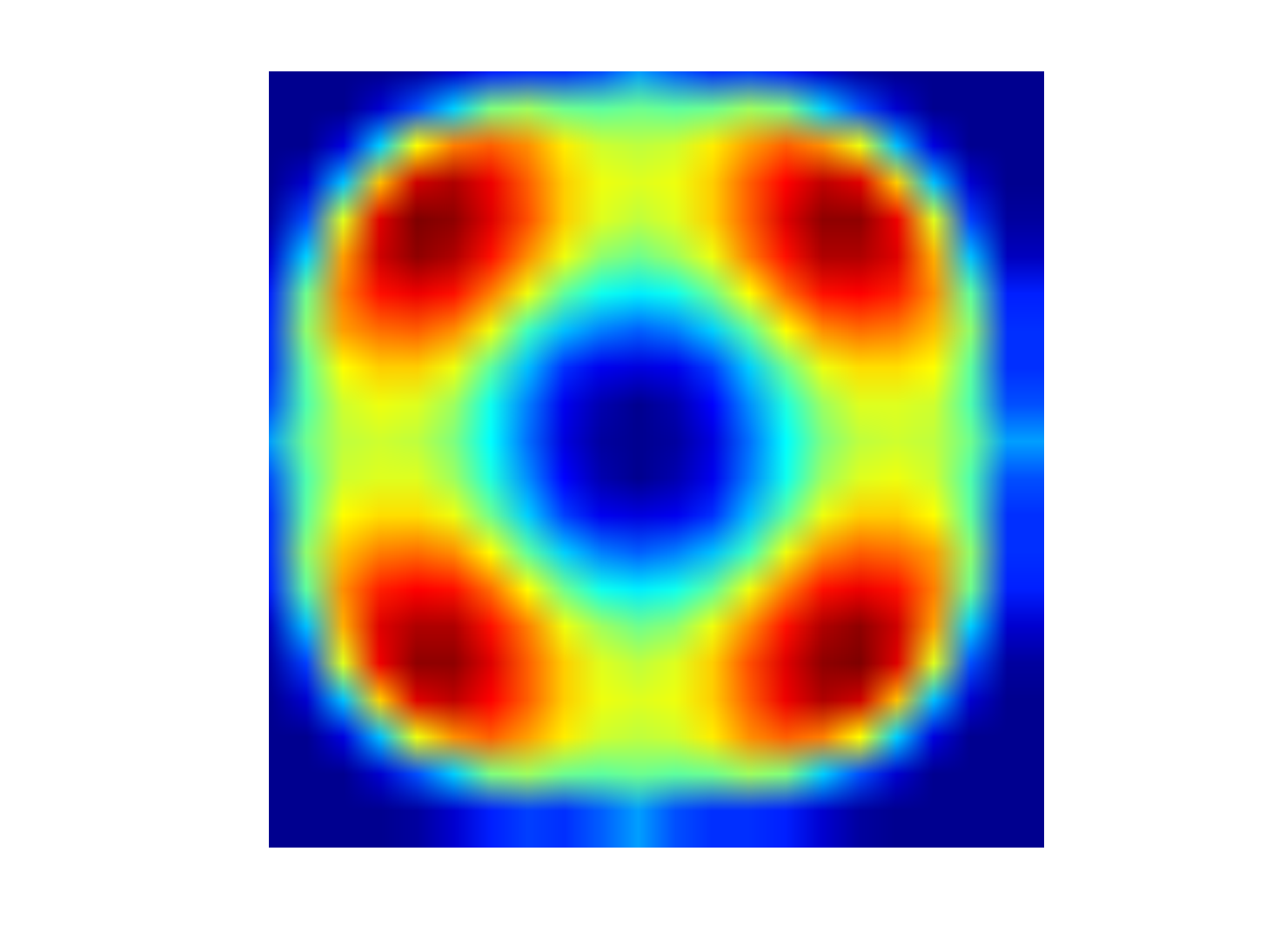}\hspace{-0.5mm}
\includegraphics[height=\IH,trim=92 30 85 30,clip=true]{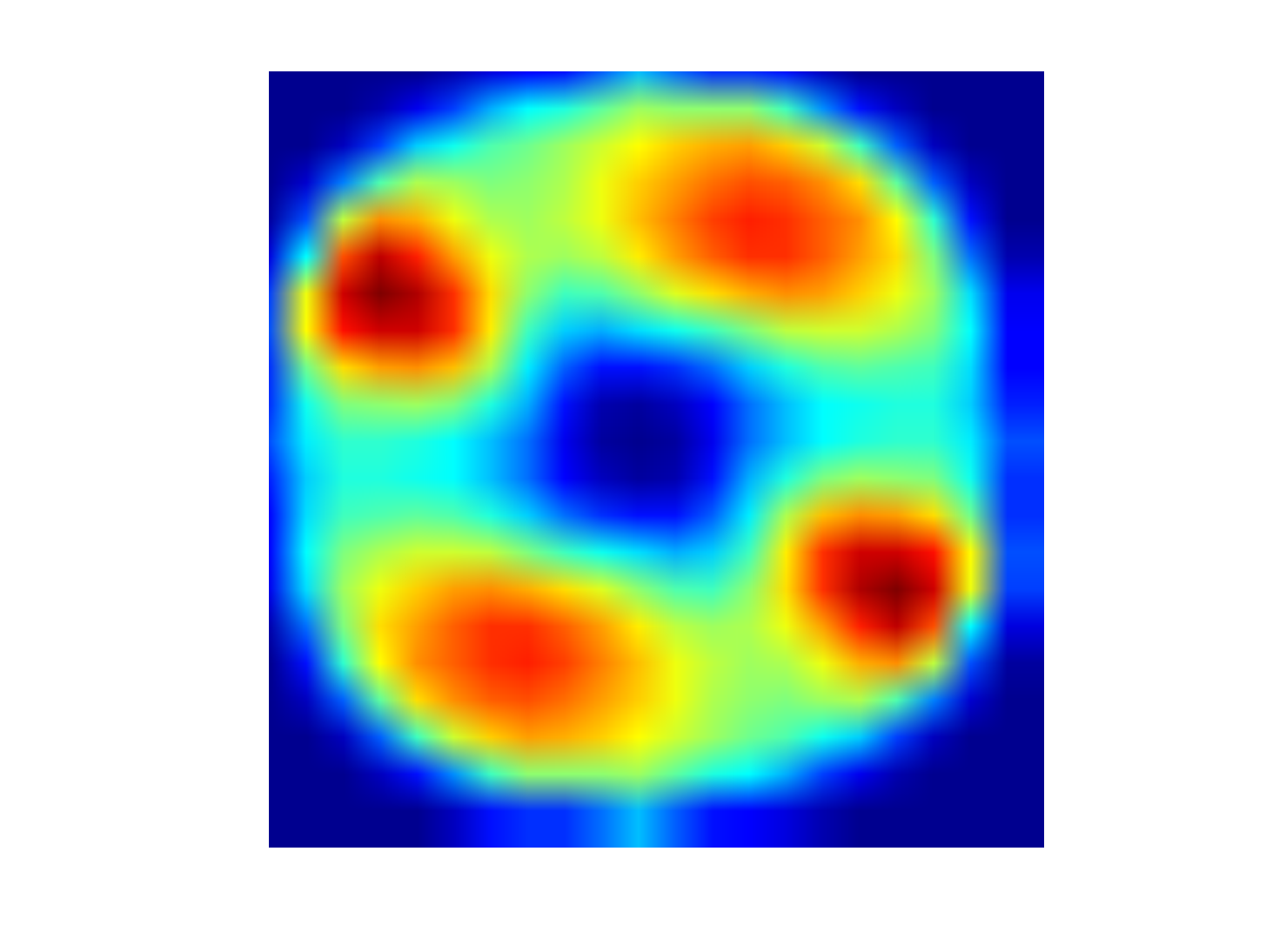}\hspace{-0.5mm}
\includegraphics[height=\IH,trim=92 30 85 30,clip=true]{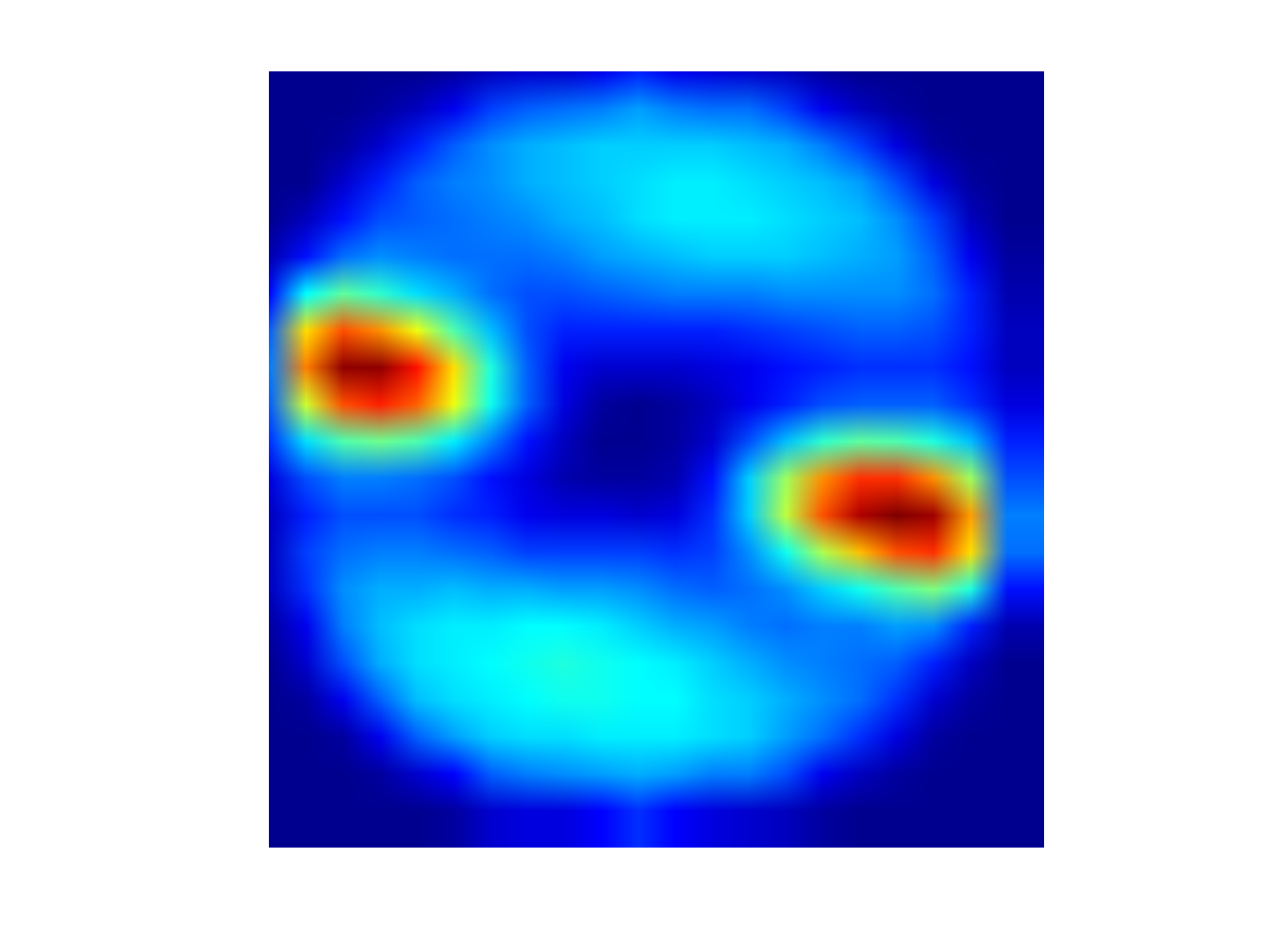}\hspace{-0.5mm}
}

\pgftext[at=\pgfpoint{0cm}{-2.25cm}]{ \hspace{-7.5mm}

\includegraphics[height=\IHB]{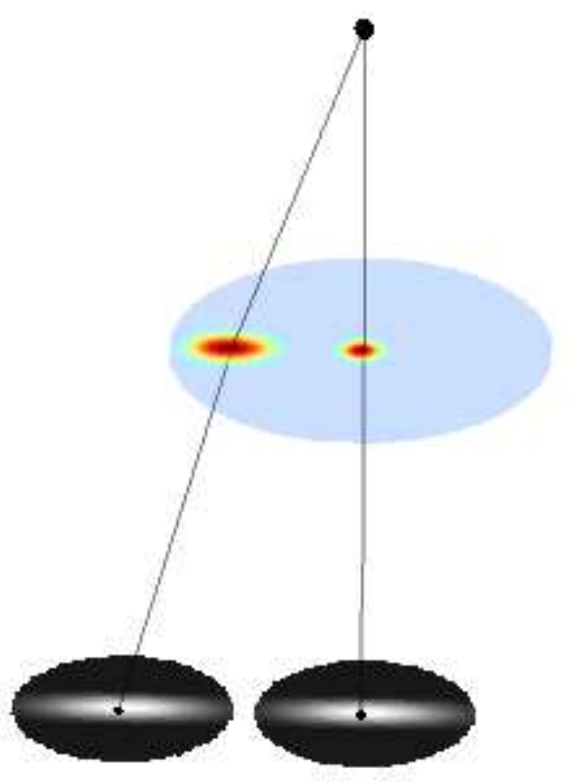}\hspace{7.5mm}
\includegraphics[height=\IHB]{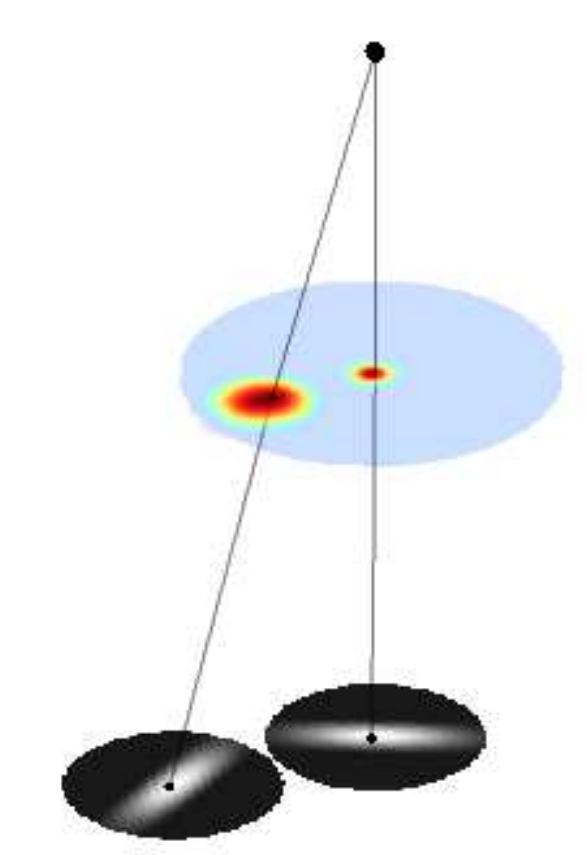}\hspace{0.6mm}
\includegraphics[height=\IHB]{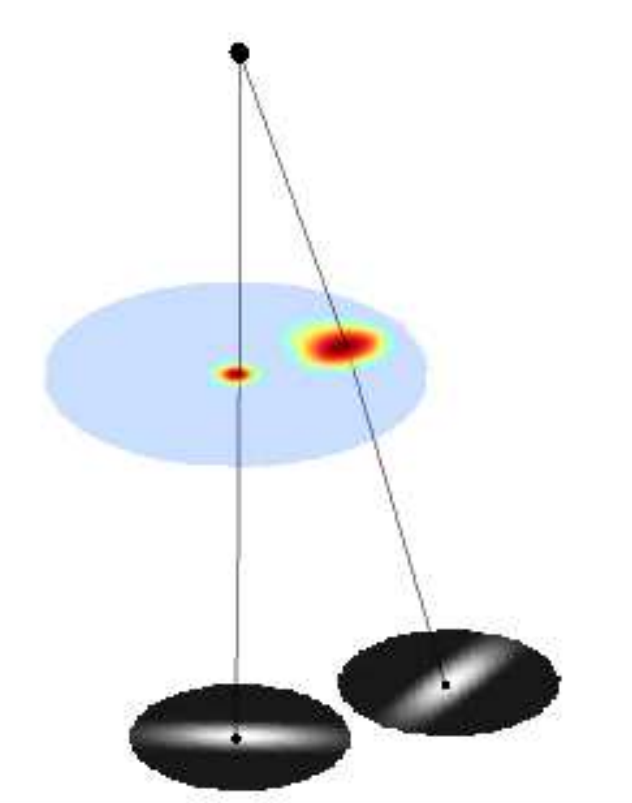}\hspace{9.mm}
\includegraphics[height=\IHB]{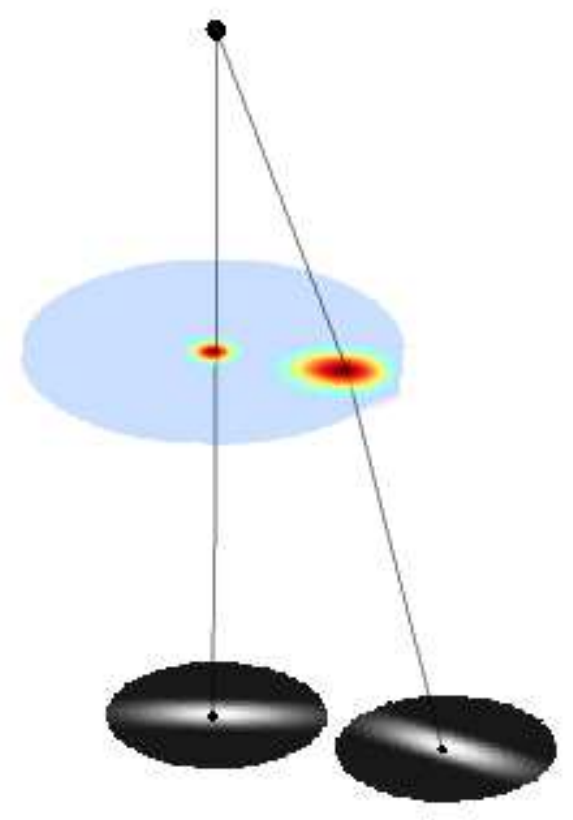}\hspace{-0.5mm}
}

\draw (-3, 1.2) node[text width=3.cm]  (1) {$\model_i^1=\,$---\\
$\model_j^1=\,$---};

\draw (-1.5, 1.2) node[text width=3.cm]  (1) {$\model_i^1=\,$---\\
$\model_j^1=\,$\raisebox{-0.25em}{\rotatebox{30}{---}}};

\draw (-0.05, 1.2) node[text width=3.cm]  (1) {$\model_i^1=\,$---\\
$\model_j^1=\,$\raisebox{-0.4em}{\rotatebox{60}{---}}};

\draw (1.5, 1.2) node[text width=3.cm]  (1) {$\model_i^1=\,$---\\
$\model_j^1=\,$\raisebox{-0.25em}{\rotatebox{90}{---}}};

\draw (2.95, 1.2) node[text width=3.cm]  (1) {$\model_i^1=\,$---\\
$\model_j^1=\,$\raisebox{-0.1em}{\rotatebox{120}{---}}};

\draw (4.4, 1.2) node[text width=3.cm]  (1) { $\model_i^1=\,$---\\
$\model_j^1=\,$\raisebox{0.3em}{\rotatebox{-30}{---}}};

\draw[line width=1.6pt,draw=black!100!blue,fill=black,fill
opacity=0.0] (-4.33, 0.05) ellipse (5.8pt and 3.4pt);

\draw[rotate=15,line width=1.6pt,draw=black!100!blue,fill=black,fill
opacity=0.0] (-0.31, 0.425) ellipse (5.2pt and 4.1pt);
\draw[rotate=15,line width=1.5pt,draw=black!100!blue,fill=black,fill
opacity=0.0] (-1.36, 0.15) ellipse (5.2pt and 4.1pt);

\draw[rotate=-12,line
width=1.5pt,draw=black!100!blue,fill=black,fill opacity=0.0] (3.96,
0.76) ellipse (6.1pt and 4.1pt);

\draw[->,thin] (-4.34,-0.05) to  (-3.5, -1);

\draw[->,thin] (-0.32,0.25) to  (0.4, -1);

\draw[->,thin] (-1.35,-0.34) to  (-1.07, -1);

\draw[->,thin] (3.96,-0.19) to  (3.09, -1);
 \end{tikzpicture}
 \caption{{\small {\bf Top}: Examples of histograms $h_{ij}^{\ell=2}$. {\bf Bottom}: \emph{Examples} of duplets defined by modes in $h_{ij}^{\ell=2}$. The number of all duplets here would be $2$ for the first two and the last histogram, and $4$ for the third, fourth and fifth histogram, with the total of $18$ duplets.}}
  \label{fig:maps_sizes}
\end{figure}

\def\IH{5.2cm}

To train the top, object-layer of compositions, we make use of the available supervision, similarly
as in the constellation model~\cite{s:weber00}. During the greedy selection of the compositions, we always
select the one with the highest detection accuracy (we use the F-measure).

\subsubsection{(Re-)Estimating the parameters}
\label{sec:parameters}

Subsec.~\ref{sec:selection} yields a vocabulary at layer $\ell$. However, notice that the parameters (spatial relations) have been inherited from the duplets and thus may not be optimal for each full composition. We thus make several EM-like iterations through the data to adjust the parameters: We first hold the current parameters fixed and do inference on all neighborhoods in all images. For each neighborhood we take the best scoring composition whose matching score is higher than $\tau^\ell$. For each such detection we store the locations of the parts. For each composition type we then re-estimate the Gaussian distributions using the locations of parts from all detections.

For the top, object layer we also update the appearance parameters $\appearance^{\hspace{-1mm}\cO}$ for each composition. We do this via co-occurrence in the following way. Suppose $z^\ell=(\model^\ell,x^\ell)$ is our top scoring composition for neighborhood $\mathcal N$, and $z_p^{\ell-1}$ is the best match for the $p$-th part. We then find all states $z_{q}^{\ell-1}=(\modelm_q^{\ell-1}, x^{\ell-1})$ for which the IOU overlap of the supports $\supp(z_p^{\ell-1})$ and $\supp(z_q^{\ell-1})$ exceeds $0.8$. We update a histogram $\bar \appearance^{\ell}(q)$. We take a pass through all neighborhoods, and finally normalize the histogram to obtain the final appearance parameters.

We only still need to infer the parameters for the first layer
models in the vocabulary. This is done by estimating the parameters
$(\mu_i^1,\Sigma_i^1)$ of a multivariate Gaussian distribution for
each model $\model_i^1$: Each Gabor feature vector $\mathbf f$ is
first normalized to have the dominant orientation equal to $1$. All
features $\mathbf  f$ that have the $i$-th dimension equal to $1$
are then used as the training examples for estimating the parameters
$(\mu_i^1,\Sigma_i^1)$.

\subsubsection{Getting the OR layers}
\label{sec:merging}

Different compositions can have different tree structures but can describe 
approximately the same shape. We want to form OR nodes (disjunctions or mixtures) of such compositions.
We do this by running inference with our vocabulary and selecting a number of different detections for each composition $\model^\ell$. For each detection $z^\ell=(\model^\ell,x^\ell)$ we compute the Shape Context descriptor~\cite{belongiesm} on the support $\supp(z^\ell)$. For each $\model^\ell$ we compute a prototype descriptor, one that is most similar to all other descriptors using the $\chi^2$ similarity measure. We perform agglomerative clustering on the prototypical descriptors which gives us the clusters of compositions. Each cluster defines an OR node at layer $\Modelm^\ell$.

\subsection{Learning a multi-class vocabulary}
\label{sec:multiclass}

In order to learn a single vocabulary to represent multiple object classes  we use the following strategy.
We learn layers $2$ and $3$ on a set of natural images containing scenes and objects without annotations.
This typically results in very generic vocabularies with compositions shared by many classes. To learn layer $4$ and higher we use annotations in the form of bounding boxes with class labels. We learn the layers incrementally, first using only images of one class, and then incrementally adding the compositions to the vocabulary on new classes.

When training the higher layers we scale each image such that the
diagonal of the bounding box is approximately $250$ pixels, and only learn our vocabulary within the boxes. Note that the size
of the training images has a direct influence on the number of
layers being learned (due to compositionality, the number of layers
needed to represent the whole shapes of the objects is logarithmic
in the number of extracted edge features in the image). In order to
learn a $6$-layer hierarchy, the $250$ pixel diagonal constraint has
proven to be a good choice.

Once we have learned the multi-class vocabulary, we could, in
principle, run the parameter learning algorithm (Sec.~\ref{sec:parameters}) to re-learn the parameters over the
complete hierarchy in a similar fashion as in the Hierarchical
HMMs~\cite{s:singer98}. 
However, we have not done so in this paper.

\subsubsection{Learning the thresholds for the \emph{tests}}
\label{sec:thresholds}

Given that the object class representation is known, we can learn
the thresholds $\tau_{\model^\ell}$ to be used in our (approximate)
inference algorithm (Sec.~\ref{sec:effcomp}).

We use a similar idea to that of Amit et al.~\cite{s:amit04} and
Fleuret and Geman~\cite{s:fleuret01}. The main goal is to learn the
thresholds in way that nothing is lost with respect to the accuracy
of detection, while at the same time optimizing for the efficiency
of inference.

Specifically, by running the inference algorithm on the set of class
training images $\{I_k\}$, we obtain the object detection scores 
$\score(z^\cO)$. For each composition $\model^\ell$ we find the smallest score 
it produces in any of the parse graphs of positive object detections $z^\cO$
over all train images $I_k$. Threshold for its score is then:
\begin{equation} \label{eq:thresholds}
\tau_{\model^\ell} = \min_k \min_{(\model^\ell, x) \in \mathcal{P}_{I_k}(z^\cO)} \score(\omega^\ell, x).
\end{equation}
For a better generalization, however, one can rather take a certain
percentage of the value on the right~\cite{s:fleuret01}.

\section{Experimental results}
\label{sec:results}

The evaluation is separated into two parts. We first show the
capability of our method to learn generic shape structures from
natural images and apply them to a object classification task.
Second, we utilize the approach for multi-class object detection on
$15$ object classes.

All experiments were performed on one core on a Intel Xeon-$4$ CPU
$2.66\,$Ghz computer. The method is implemented in C++. Filtering is
performed on CUDA.

\subsection{Natural statistics and object classification}
\label{sec:exp_classification}

\begin{figure*}[htb!]
\centering
\begin{minipage}{6.41cm}
\includegraphics[width=6.41cm,trim=20 18 800 18,clip=true]{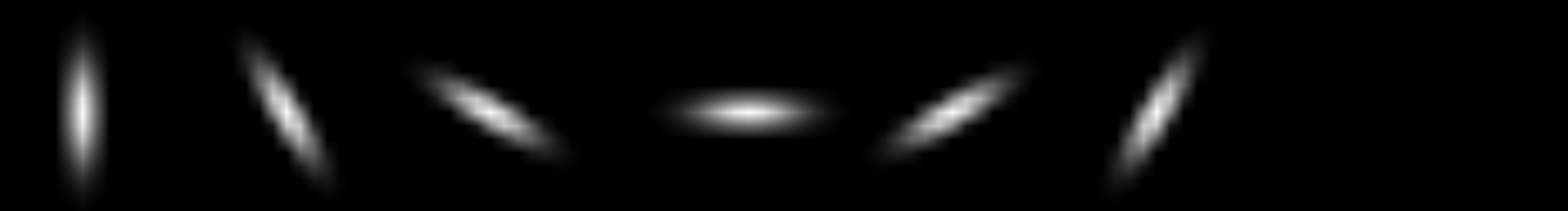}\\[-0.8mm]
{\setlength\fboxsep{1.85pt}\setlength\fboxrule{0pt}\colorbox{sgray}{\framebox[6.29cm][l]{\footnotesize Layer 1}}}\\[4.2mm]
\includegraphics[width=6.41cm,trim=55 86 40 74,clip=true]{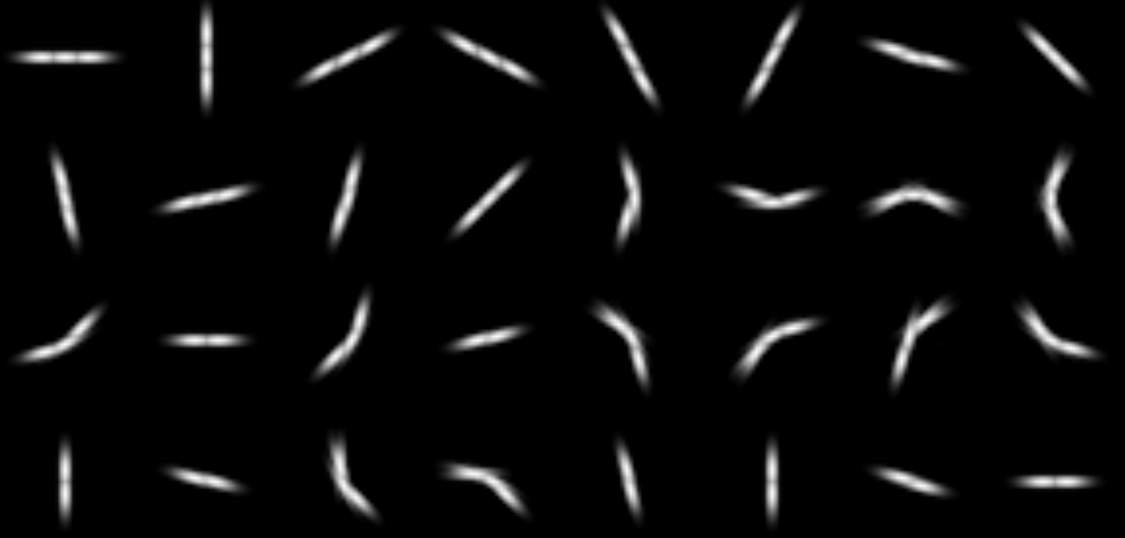}\\[-0.8mm]
{\setlength\fboxsep{1.85pt}\setlength\fboxrule{0pt}\colorbox{sgray}{\framebox[6.29cm][l]{\footnotesize
Layer 2}}}
\end{minipage}
\begin{minipage}{11.6cm}
\includegraphics[width=11.6cm,trim=55 92 39 84,clip=true]{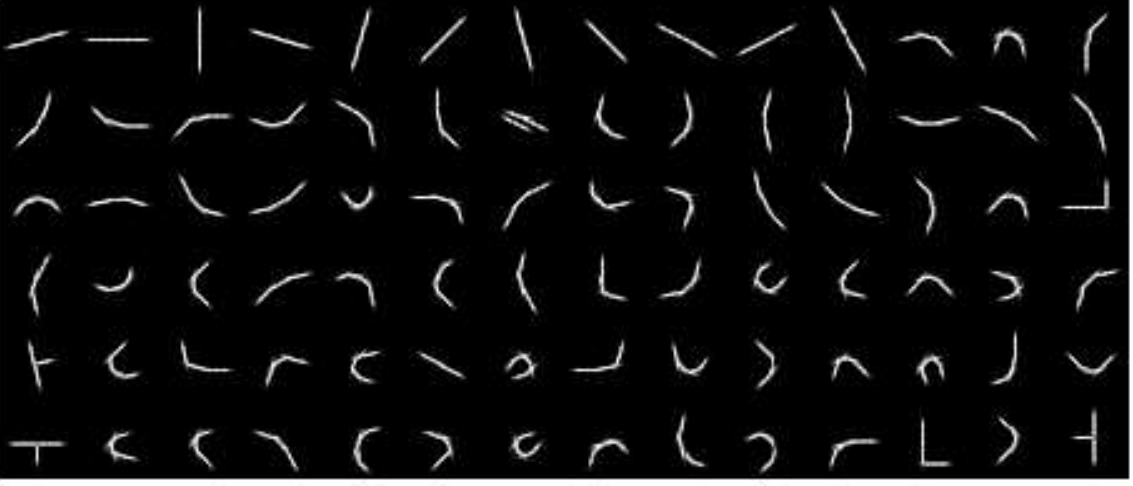}\\[-1.4mm]
{\setlength\fboxsep{1.85pt}\setlength\fboxrule{0pt}\colorbox{sgray}{\framebox[11.46cm][l]{\footnotesize
Layer 3}}}
\end{minipage}
\caption{Examples of learned vocabulary compositions (with the
exception of a fixed Layer $1$) learned on $1500$ natural images.
Only the mean of the compositions are depicted.}
 \label{fig:layers}
\end{figure*}

We applied our learning approach to a collection of $1500$ natural images.
Learning was performed only up to layer $3$, 
with the complete learning process taking roughly $5$ hours. The
learned hierarchy consisted of $160$ compositions on Layer $2$ and
$553$ compositions on Layer $3$. A few examples from both layers are
depicted in Fig.~\ref{fig:layers} (note that the images shown are
only the mean shapes modeled by the compositions). The
learned features include corners, end-stopped lines, various
curvatures, T- and L-junctions. Interestingly, these structures
resemble the ones predicted by the Gestalt
rules~\cite{s:wertheimer23}.

To put the proposed hierarchical framework in relation to other
categorization approaches which focus primarily shape, the learned $3-$layer vocabulary
was tested on the Caltech-$101$ database~\cite{s:feifei04}. The Caltech-$101$
dataset contains images of $101$ different object categories with
the additional background category. The number of images varies from
$31$ to $800$ per category, with the average image size of roughly
$300\times 300$ pixels.

Each image was processed at $3$ scales spaced apart by
$\sqrt 2$. 
The scores of the hidden states 
were combined with a linear one-versus-all SVM
classifier~\cite{s:joachims99}. This was done as follows: for each
image a vector of a dimension equal to the number of compositions at
the particular layer was formed. Each dimension of the vector, which
corresponds to a particular type of a composition (e.g. an
L-junction) was obtained by summing over all scores of the
hidden states coding this particular type of composition. To
increase the discriminative information, a radial sampling was used
similarly as in~\cite{s:malik06}: each image was split into $5$
orientation bins and $2$ distances from the image center, for which
separate vectors (as described previously) were formed and
consequently stacked together into one, high-dimensional vector.

The results, averaged over $8$ random splits of train and test
images, are reported in Table~\ref{table:caltech101} with
classification rates of existing hierarchical approaches shown for
comparison. For $15$ and $30$ training images we obtained a $60.5\%$
and $66.5\%$ accuracy, respectively, which is slightly better than
the most related hierarchical
approaches~\cite{s:ommer07,s:mutch06,s:lecun09}, comparable to those
using more sophisticated discriminative
methods~\cite{s:yu08,s:ahmed08}, and slightly worse than those using
additional information such as color or
regions~\cite{s:todorovic08a}.

\begin{figure}[htb!]
\centering 
\includegraphics[width=0.9\linewidth]{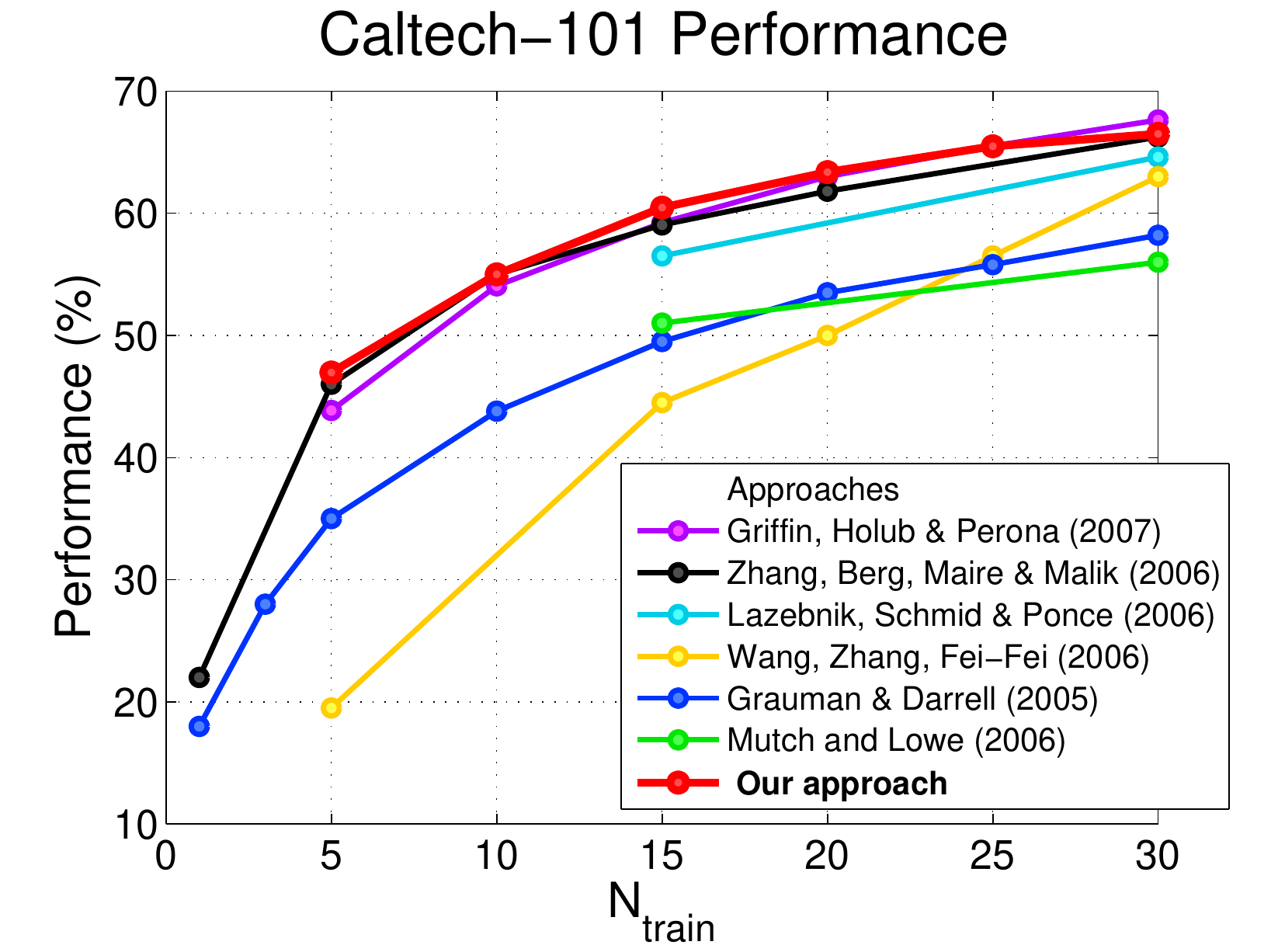}
\vspace{-2.2mm} \caption{Classification performance on
Caltech-$101$.} \label{fig:caltech101}
\end{figure}

We further tested the classification performance by varying the
number of training examples. For testing, $50$ examples were used
for the classes where this was possible and less otherwise. The
classification rate was normalized accordingly. In all cases, the
result was averaged over $8$ random splits of the data. The results
are presented and compared with existing methods in
Fig.~\ref{fig:caltech101}. We only compare to approaches that use
shape information alone and not also color and texture. Overall,
better performance has been obtained in~\cite{s:varma07}.

\begin{table}[htb!]
\addtolength{\tabcolsep}{5pt} \caption{Average classification rate
(in $\%$) on Caltech $101$.} \label{table:caltech101}
\centering\normalsize
\begin{small}
\vspace{-2.4mm}
\begin{tabular}{|l|r|r|}
\hline
 & $N_{train}=15$ & $N_{train}=30$\\
\wnhline
\hline
Mutch et al.~\cite{s:mutch06} &  $51$ & $56$ \\
\hline Ommer et al.~\cite{s:ommer07} & / & $61.3$\\
\hline Ahmed at al.~\cite{s:ahmed08} & $58.1$ & $67.2$ \\
\hline Yu at al.~\cite{s:yu08} & $59.2$ & $\mathbf{67.4}$ \\
\hline Lee at al.~\cite{s:ng09} & $57.7$ & $65.4$ \\
\hline Jarrett at al.~\cite{s:lecun09} & / & $65.5$ \\
\whline
\emph{\bf our approach} & $\mathbf{60.5}$ & $66.5$ \\
\hline
\end{tabular}
\end{small}
\vspace{-5mm}
\end{table}

\subsection{Object class detection}
\label{sec:exp_detection}

The approach was tested on $15$ diverse object classes from standard
recognition datasets. The basic information is given in
Table~\ref{table:dataset}. These datasets are known to be
challenging because they contain a high amount of clutter, multiple
objects per image, large scale differences of objects and exhibit a
significant intra class variability.

For training we used the bounding box information of objects or the
masks if they were available. Each object was resized so that its
diagonal in an image was approximately $250$ pixels. For testing, we
up-scaled each test image by a factor $3$ and used $6$ scales for
object detection. When evaluating the detection performance, a
detection is counted as correct, if the predicted bounding box
$b_{fg}$ coincides with the ground truth $b_{gt}$ more than $50\%$:
$\tfrac{area(b_{fg}\cap b_{gt})}{area(b_{fg}\cup b_{gt})} > 0.5$. On
the ETH dataset and INRIA horses this threshold is lowered to $0.3$
to enable a fair comparison with the related
work~\cite{s:ferrari07}. 
The performance is given either with
recall at equal error rate (EER) or positive detection rate at low FPPI,
depending on the type of results reported on these datasets
thus-far.

\subsubsection{Single class learning and performance}
\label{sec:single}

We first evaluated our approach to learn and detect each object
class individually. We report the training and inference times, and
the accuracy of detection, which will then be compared to the
multi-class case in Sec.~\ref{sec:multi} to test the scalability
of the approach.

\emph{Training time.} To train a class it takes on average $20-25$
minutes. For example, it takes $23$ minutes to train on Apple logo,
$31$ for giraffe, $17$ for swan, $25$ for cow, and $35$ for the
horse class. For comparison, in Shotton et al.~\cite{s:shotton08},
training on $50$ horses takes roughly $2$ hours (on a 2.2 GHz
machine using a C\#
implementation).

\emph{Inference time.} Detection for each individual class takes
from $2-4$ seconds per image, depending on the size of the image and
the amount of texture it contains. 
Other related approaches report approx. $5$ to $10$ times higher
times (in seconds): ~\cite{s:todorovic08}: $20-30$,~\cite{s:lzhu08}:
$16.9$,~\cite{s:fritz08}: $20$, ~\cite{s:fergus07}: $12-18$,
however, at slightly older hardware.

\emph{Detection performance.} The ETH experiments are performed in a $5$-fold
cross-validation obtained by sampling $5$ subsets of half of the
class images at random. The test set for evaluating detection consists of all the remaining images in the
dataset. The detection performance is given as the detection rate at
the rate of $0.4$ false-positives per image (FPPI), averaged over
the $5$ trials as in~\cite{s:ferrari07}. 
The detection performance is reported in Table~\ref{table:ferrari}.
Similarly, the results for the INRIA horses are given in a $5$-fold
cross-validation obtained by sampling $5$ subsets of $50$ class
images at random and using the remaining $120$ for testing. The test
set also includes $170$ negative images to allow for a higher FPPI
rate. With respect to~\cite{s:ferrari07}, we achieve a better
performance for all classes, most notably for giraffes ($24.7\%$).
Our method performs comparably to a discriminative framework by
Fritz and Schiele~\cite{s:fritz08}. Better performances have
recently been obtained by Maji and Malik~\cite{s:maji09} ($93.2\%$)
and Ommer and Malik~\cite{s:ommer09} ($88.8\%$) using Hough voting
and SVM-based verification. To the best of our knowledge, no other
hierarchical approach has been tested on this dataset so far.

For the experiments on the rest of the datasets we report the recall at EER.
The results are given in Table~\ref{table:ferrari}. Our approach achieves competitive
detection rates with respect to the state-of-the-art. 
Note also that~\cite{s:mikolajczyk06,s:leibe08} used $150$
training examples of motorbikes, while we only used $50$ (to enable
a fair comparison with~\cite{s:opelt08} on GRAZ).

While the performance in a single-class case is comparable to the current state-of-the-art, the main advantage of our approach are its computational properties when the number of classes is higher, as
demonstrated next.

\subsubsection{Multi-class learning and performance}
\label{sec:multi}

To evaluate the scaling behavior of our approach we have
incrementally learned $15$ classes one after another.

\begin{figure*}[htb!]
\centering
\begin{minipage}{0.487\linewidth}
\includegraphics[width=\linewidth,trim=55 75 39 63,clip=true]{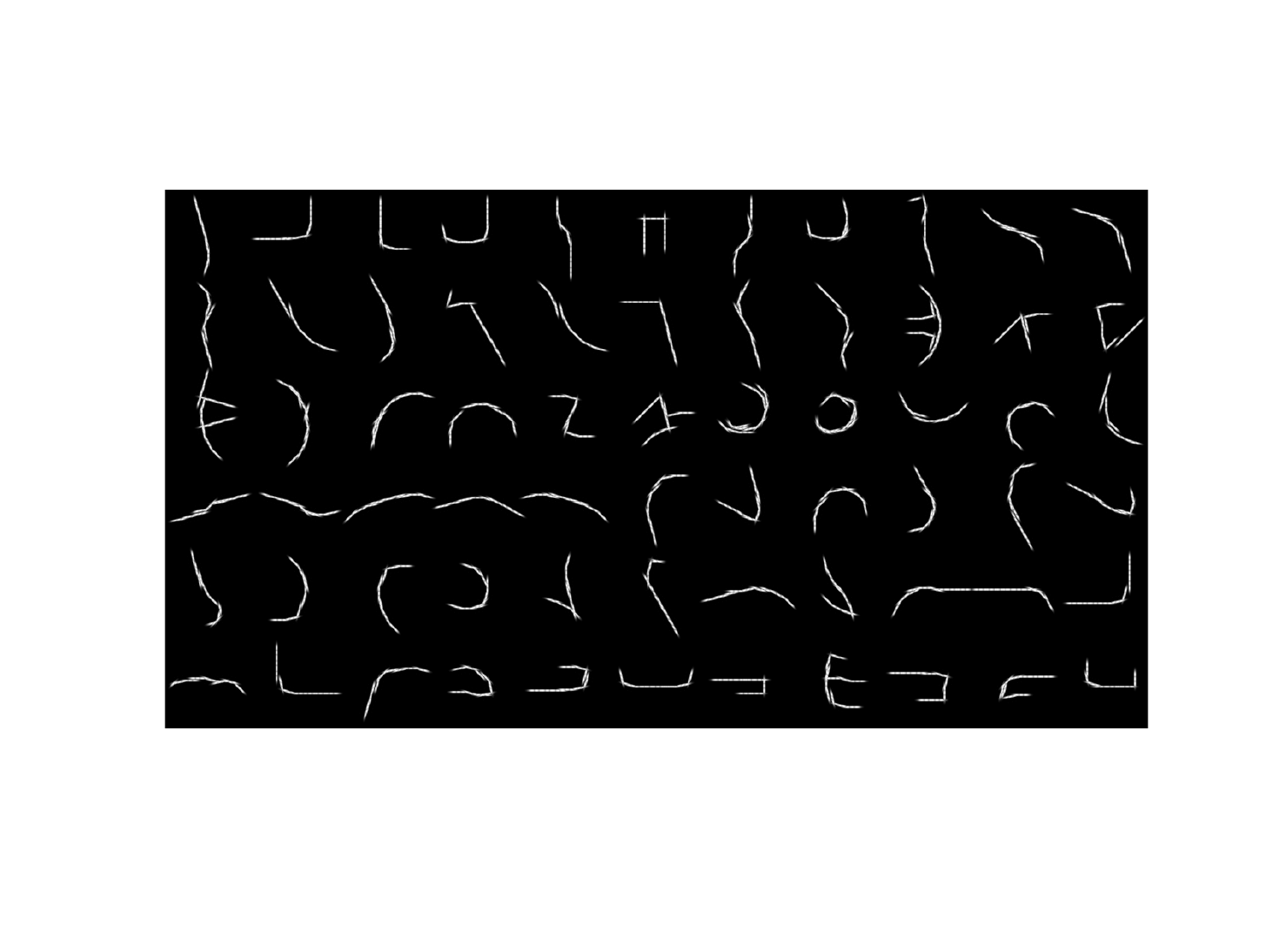}\\[-0.8mm]
{\setlength\fboxsep{1.85pt}\setlength\fboxrule{0pt}\colorbox{sgray}{\framebox[0.988\linewidth][l]{\footnotesize
Layer 4}}}
\end{minipage}
\begin{minipage}{0.506\linewidth}
\includegraphics[width=\linewidth]{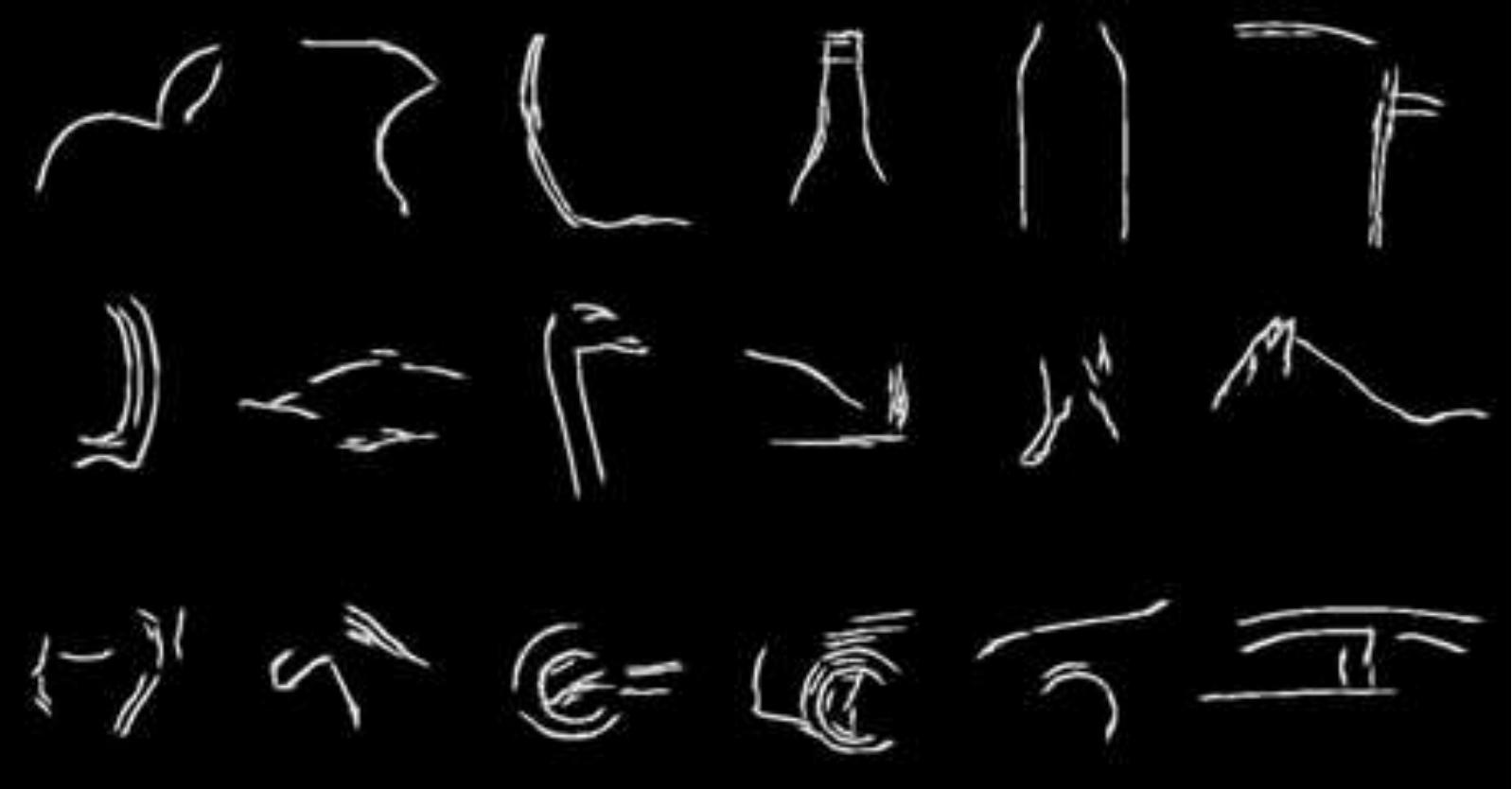}\\[-0.8mm]
{\setlength\fboxsep{1.85pt}\setlength\fboxrule{0pt}\colorbox{sgray}{\framebox[0.988\linewidth][l]{\footnotesize
Layer 5}}}
\end{minipage}

\vspace{1.9mm}
\begin{tikzpicture}[style=thick, scale=1]
\pgftext[at=\pgfpoint{-2.5cm}{1.cm}]{
\hspace{-1.4mm}\begin{minipage}{\linewidth}\includegraphics[width=\linewidth]{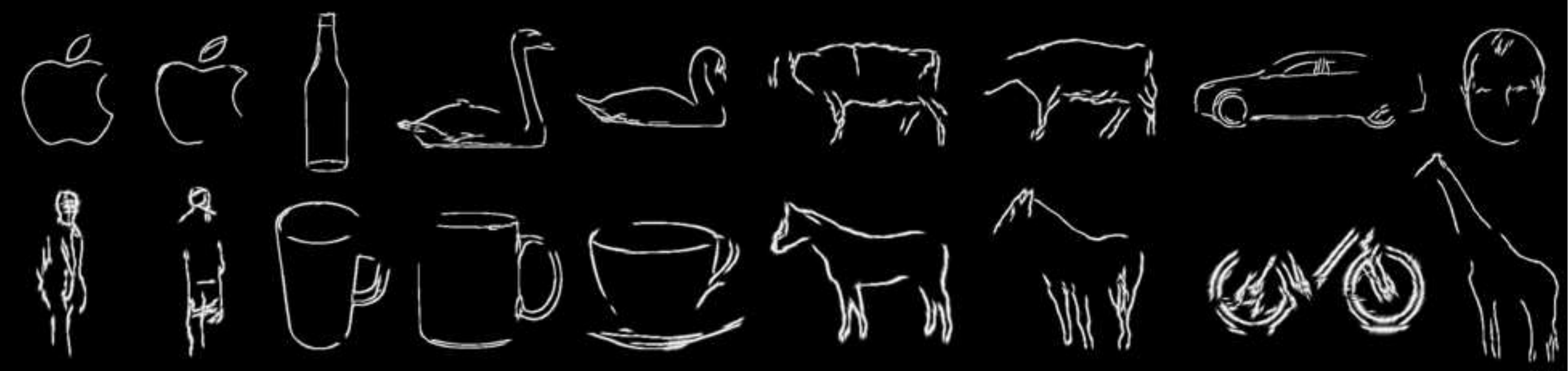}\\[-0.8mm]
{\setlength\fboxsep{1.85pt}\setlength\fboxrule{0pt}\colorbox{sgray}{\framebox[0.992\linewidth][l]{\footnotesize
Layer $6$ (object layer $\cO$)}}}
\end{minipage}
}

\vspace{0.mm}

\filldraw [blue!80!white] (-11.62, -1.9) rectangle +(0.21,0.21);

\draw (-10.5, -1.86) node (1) {\sf{Apple logo}};

\filldraw [red!65!black] (-9.33, -1.9) rectangle +(0.21,0.21);

\draw (-8.5, -1.88) node (2) {\sf{person}};

\filldraw [blue!52!red!75!pink] (-7.6, -1.9) rectangle +(0.21,0.21);

\draw (-6.89, -1.8) node  (3) {\sf{bottle}};

\filldraw [orange!80!red] (-6.11, -1.9) rectangle +(0.21,0.21);

\draw (-5.48, -1.86) node (4) {\sf{mug}};

\filldraw [yellow!90!white] (-4.78, -1.9) rectangle +(0.21,0.21);

\draw (-4.07, -1.8) node  (5) {\sf{swan}};

\filldraw [red] (-3.24, -1.9) rectangle +(0.21,0.21);

\draw (-2.65, -1.83) node (6) {\sf{cup}};

\filldraw [blue!20!red!65!pink] (-1.97, -1.9) rectangle
+(0.21,0.21);

\draw (-1.39, -1.8) node  (7) {\sf{cow}};

\filldraw [yellow!55!green] (-0.67, -1.9) rectangle +(0.21,0.21);

\draw (0.07, -1.78) node  (8) {\sf{horse}};

\filldraw [blue!75!green!50!cyan] (0.94, -1.9) rectangle
+(0.21,0.21);

\draw (1.49, -1.82) node  (9) {\sf{car}};

\filldraw [green!70!black] (2.14, -1.9) rectangle +(0.21,0.21);

\draw (2.98, -1.82) node  (10) {\sf{bicycle}};

\filldraw [orange!70!white] (3.91, -1.9) rectangle +(0.21,0.21);

\draw (4.53, -1.78) node  (11) {\sf{face}};

\filldraw [cyan] (5.22, -1.9) rectangle +(0.21,0.21);

\draw (5.99, -1.82) node  (11) {\sf{giraffe}};

\filldraw [blue!80!white] (-10.91,2.23) rectangle +(0.06,0.06);

\filldraw [blue!80!white] (-9.35,2.23) rectangle +(0.06,0.06);

\filldraw [blue!50!red!75!pink] (-7.89,2.23) rectangle +(0.06,0.06);

\filldraw [yellow!80!white] (-6.25,2.01) rectangle +(0.06,0.06);

\filldraw [yellow!80!white] (-4.09,2.22) rectangle +(0.06,0.06);

\filldraw [blue!20!red!65!pink] (-1.57,2.53) rectangle +(0.06,0.06);

\filldraw [blue!20!red!65!pink] (0.86,2.53) rectangle +(0.06,0.06);

\filldraw [blue!75!green!50!cyan] (3.52,2.3) rectangle +(0.06,0.06);

\filldraw [yellow!55!green] (-1.51,0.37) rectangle +(0.06,0.06);

\filldraw [yellow!55!green] (0.9,0.37) rectangle +(0.06,0.06);

\filldraw [orange!80!red] (-7.95,0.15) rectangle +(0.06,0.06);

\filldraw [orange!90!red] (-6.25,0.15) rectangle +(0.06,0.06);

\filldraw [red] (-4.1,0.15) rectangle +(0.06,0.06);

\filldraw [red!65!black] (-10.96,0.3) rectangle +(0.06,0.06);

\filldraw [red!65!black] (-9.31,0.3) rectangle +(0.06,0.06);

\filldraw [orange!70!white] (5.73,2.14) rectangle +(0.06,0.06);

\filldraw [green!70!black] (3.43,0.12) rectangle +(0.06,0.06);

\filldraw [cyan] (5.82,0.12) rectangle +(0.06,0.06);

 \end{tikzpicture}

 \vspace{-2.5mm}
\caption{Examples of samples from compositions at layers $4$, $5$,
and top, object layer learned on $15$ classes.}
 \label{fig:layers_higher}
\end{figure*}

\emph{The learned vocabulary.} A few examples of the learned shapes
at layers $4$ to $6$ are shown in Fig.\ref{fig:layers_higher} (only
samples from the generative model are depicted). 
An example of a complete object-layer composition is depicted in Fig.~\ref{fig:models}. 
It can be seen that the approach has learned the essential structure of
the class well, not missing any important shape information.

\emph{Degree of composition sharing among classes.} To see how
shared are the vocabulary compositions between classes of different
degrees of similarity, we depict the learned vocabularies for two
visually similar classes (motorbike and bicycle), two semi-similar
classes (giraffe and horse), and two dissimilar classes (swan and
car\_front) in Fig.~\ref{fig:sharing}. The nodes correspond to
compositions and the links denote the compositional relations
between them and their parts. The green nodes represent the
compositions used by both classes, while the specific colors denote
class specific compositions. If a spatial relation is shared as
well, the edge is also colored green. The computational advantage of
our hierarchical representation over flat
ones~\cite{s:opelt08,s:leibe08,s:torralba07} is that the
compositions are shared at \emph{several} layers of the vocabulary,
which significantly reduces the overall complexity of inference.
Even for the visually dissimilar classes, which may not have any
complex parts in common, our representation is highly shared at the
lower layers of the hierarchy.

To evaluate the shareability of the learned compositions
among the classes, we use the following measure: 
\begin{equation*}
\label{eq:sharing} \mathrm{deg\_share}(\ell) =
\frac{1}{|\Modelm^{\ell}|}\sum_{\modelm^{\ell}\in \Modelm^\ell}\frac{(\# \text{ of
classes that use } \modelm^{\ell})-1}{\# \text{ of all classes}-1},
\end{equation*}
defined for each layer $\ell$ separately. By ``$\model^{\ell}$ used
by class $c$'' it is meant that the probability of $\model^{\ell}$
under $c$ is not equal to zero. To give some intuition behind the
measure: $deg\_share=0$ if no composition from layer $\ell$ is
shared (each class uses its own set of compositions), and it is $1$
if each composition is used by all the classes.
Fig.~\ref{fig:inference_time} (b) plots the values for the learned
$15-$class representation. Beside the mean (which defines
$deg\_share$), the plots also show the standard deviation.

\emph{The size of the vocabulary.} It is important to test how the
size of the vocabulary (the number of compositions at each layer)
scales with the number of classes, since this has a direct influence
on inference time. We report the size as a function of the number of
learned classes in Fig.~\ref{fig:size_representation}. For the
``worst case'' we take the independent training approach: a
vocabulary for each class is learned independently of other classes
and we sum over the sizes of these separate vocabularies. One can
observe a logarithmic tendency especially at the lower layers. This
is particularly important because the complexity of inference is
much higher for these layers (because they contain less
discriminative compositions which are detected more numerously in
images). Although the fifth layer contains highly class specific
compositions, one can still observe a logarithmic increase in size.
The final, object layer, is naturally linear in the number of
classes, but it does, however, learn less object models than is the
number of all training examples of objects in each class.

We further compare the scaling tendency of our approach with the one
reported for a non-hierarchical representation by Opelt et
al.~\cite{s:opelt08}. The comparison is given in
Fig.~\ref{fig:inference_time} (a) where the worst case is taken as
in~\cite{s:opelt08}. We compare the size of the class-specific
vocabulary at layer $5$, where the learned compositions are of
approximately the same granularity as the features used
in~\cite{s:opelt08}. We additionally compare the overall size of the
vocabulary (a sum of the sizes over all layers). Our approach
achieves a substantially better scaling tendency, which is due to
sharing at multiple layers of representation. This, on the one hand,
compresses the overall representation, while it also attains a
higher variability of the compositions in the higher layers and
consequently, a lower number of them are needed to represent the
classes.

Fig.~\ref{fig:inference_time} (d) shows the storage demands as a
function of the number of learned classes. This is the actual size
of the vocabulary stored on a hard disk. Notice that the $15$-class
hierarchy takes only $1.6Mb$ on disk.

\emph{Inference time.} We further test how the complexity of
inference increases with each additional class learned. We randomly
sample ten images per class and report the detection times averaged
over all selected images. The times are reported as a function of
the number of learned classes. The results are plotted in
Fig.~\ref{fig:inference_time} (c), showing that the running times
are significantly faster than the ``worst case'' (the case of
independent class representation, in which each separate class
vocabulary is used to detect objects). It is worth noting, that it
takes only $16$ seconds per image to apply a vocabulary of all $15$
classes.

\emph{Detection performance.} We additionally test the multiclass
detection performance. The results are presented in
Table~\ref{table:comparison_multi} and compared with the performance
of the single class vocabularies. The evaluation was the following.
For the single class case, a separate vocabulary was trained on each
class and evaluated for detection independently of other classes.
For the multi-class case, a joint multi-class vocabulary was learned
(as explained in Sec.~\ref{sec:multi}) and detection was performed
as proposed in Sec.~\ref{sec:multiclass}, that is, we allowed for
competition among hypotheses explaining partly the same regions in
an image.

Overall, the detection rates are slightly worse in the multi-class
case, although in some cases the multi-class case outperformed the
independent case. The main reason for a slightly reduced performance
is due to the fact that the representation is generative and is not
trained to discriminate between the classes. Consequently, it does
not separate similar classes sufficiently well and a wrong
hypothesis may end up inhibiting the correct one. As part of the
future work, we plan to also incorporate more discriminative
information into the representation and increase the recognition
accuracy in the multi-class case. On the other hand, the increased
performance for the multi-class case for some objects is due to
feature sharing among the classes, which results in better
regularization and generalization of the learned representation.

\def\IH{3.cm}

\def\IH{3.cm}

\begin{table*}[htb!]
\addtolength{\tabcolsep}{-0.3pt}
\begin{small}
\centering
 \caption{Information on the datasets used to evaluate the detection performance.}
  \label{table:dataset}
  \vspace{-3mm}
\begin{tabular}{|c|r|r|r|r|r|r|r|r|r|}

\whline
 dataset & \multicolumn{5}{c|}{ETH shape} & UIUC & Weizmann & INRIA & TUD \\
 \cline{1-10}
class & Apple logo & bottle & giraffe & mug & swan & car\_side (multiscale) & horse (multiscale) & horse  & motorbike\\
\hline

$\#$ train im. & $19$ & $21$ & $30$ & $24$ & $15$ & $40$ & $20$ & $50$ & $50$ \\
\hline

$\#$ test im. & $21$ & $27$ & $57$ & $24$ & $17$ & $108$ & $228$ & $290$ & $115$ \\

\whline
\end{tabular}
\vspace{2mm}
\addtolength{\tabcolsep}{0.71pt}
\begin{tabular}{|c|r|r|r|r|r|r|r|r|r|r|r|r|} \whline
 dataset &  \multicolumn{11}{c|}{GRAZ}\\
 \cline{1-12}
class & bicycle\_side & bottle & car\_front & car\_rear & cow & cup & face & horse\_side & motorbike & mug & person\\
\hline

$\#$ train im. & $45$ & / & $20$ & $50$ & $20$ & $16$ & $50$ & $30$ & $50$ & / & $19$ \\

\hline

$\#$ test im. & $53$ & $64$ & $20$ & $400$ & $65$ & $16$ & $217$ & $96$ & $400$ & $15$ & $19$ \\

\whline
\end{tabular}
\end{small}
\vspace{-2.6mm}
\end{table*}

\begin{table*}[htb!]
\centering \caption{Detection results. On the ETH shape and INRIA
horses we report the detection-rate (in $\%$) at $0.4$ FPPI averaged
over five random splits train/test data. For all the other datasets
the results are reported as recall at equal-error-rate (EER).}
\label{table:ferrari}
\vspace{-2.2mm}
\begin{minipage}{0.55\linewidth}
\addtolength{\tabcolsep}{3.2pt}
\begin{tabular}{|c|l|r|r|r|} \whline
\multirow{8}{7.3mm}{ETH shape} & {\bf class} & {\bf
~\cite{s:ferrari07}} & {\bf ~\cite{s:fritz08}}
 & \multicolumn{1}{c|}{{\bf our approach}}\\
\hline

 & applelogo & $83.2\,(1.7)$ & $89.9\,(4.5)$ & $87.3\,(2.6)$ \\

\cline{2-5}

 & bottle & $83.2\,(7.5)$ & $76.8\,(6.1)$ & $\bf{86.2\,(2.8)}$ \\

\cline{2-5}

 & giraffe & $58.6\,(14.6)$ & $90.5\,(5.4)$ & $83.3\,(4.3)$ \\

\cline{2-5}

 & mug & $83.6\,(8.6)$ & $82.7\,(5.1)$ & $\bf{84.6\,(2.3)}$ \\

\cline{2-5}

 & swan & $75.4\,(13.4)$ & $84.0(\,8.4)$ & $78.2\,(5.4)$ \\

\cline{2-5}

 & {\bf average} & $76.8$ & $84.8$ & $83.7$ \\

\hline \hline

INRIA & horse & $84.8 (2.6)$ & / & $\mathbf{85.1(2.2)}$ \\

\whline
\end{tabular}
\vspace{1.1mm} \addtolength{\tabcolsep}{-3.7pt}
\begin{tabular}{|c|l|r|r|r|} \whline
 & {\bf class} & \multicolumn{2}{c|}{{\bf related work}} & {\bf our approach}\\

\hline

\multirow{1}{12.3mm}{UIUC} & car\_side, multiscale & $90.6$~\cite{s:mutch06} & $93.5$~\cite{s:todorovic08a} & $\bf{93.5}$ \\

 \hline

\multirow{1}{12.3mm}{Weizmann} &  horse\_multiscale  & $89.0$~\cite{s:shotton08} & $93.0$~\cite{s:shotton08a} & $\bf{94.3}$ \\

\hline

\multirow{1}{12.3mm}{TUD} & motorbike & $87$~\cite{s:leibe08} &
$88$~\cite{s:mikolajczyk06}
& $83.2$\\

\whline
\end{tabular}
\end{minipage}
\hspace{1mm}
\begin{minipage}{0.42 \linewidth}
\vspace{1.1mm} \addtolength{\tabcolsep}{2pt}
\begin{tabular}{|c|l|r|r|r|} \whline

    & {\bf class} & {\bf ~\cite{s:opelt08}} & {\bf ~\cite{s:shotton08}} & {\bf our approach}\\

 \hline

 & face & $96.4$ & $97.2$ & $94$ \\

\cline{2-5}

\multirow{9}{7.3mm}{GRAZ} & bicycle\_side & $72$& $67.9$ &
$68.5$\\

\cline{2-5}

 & bottle & $91$ & $90.6$ &
$89.1$ \\

\cline{2-5}

 & cow & $100$ & $98.5$ & $96.9$\\

\cline{2-5}

 & cup & $81.2$ & $85$ & ${\bf 85}$ \\

\cline{2-5}

 & car\_front & $90$ & $70.6$ &
$76.5$\\

\cline{2-5}

  & car\_rear & $97.7$ & $98.2$ & $97.5$ \\

\cline{2-5}

 & horse\_side & $91.8$ & $93.7$ & $\bf{93.7}$\\

\cline{2-5}

 & motorbike & $95.6$ & $99.7$ & $93.0$\\

\cline{2-5}

 & mug & $93.3$ & $90$ & $90$\\

\cline{2-5}

 & person & $52.6$ & $52.4$ & $\bf{60.4}$ \\

 \cline{2-5}

\whline
\end{tabular}
\end{minipage}
\end{table*}

\def\IH{4.06cm}
\begin{figure*}[htb!]
\centering
\addtolength{\tabcolsep}{-5pt}
\begin{tabular}{lclclc}
(a) &
\includegraphics[height=\IH]{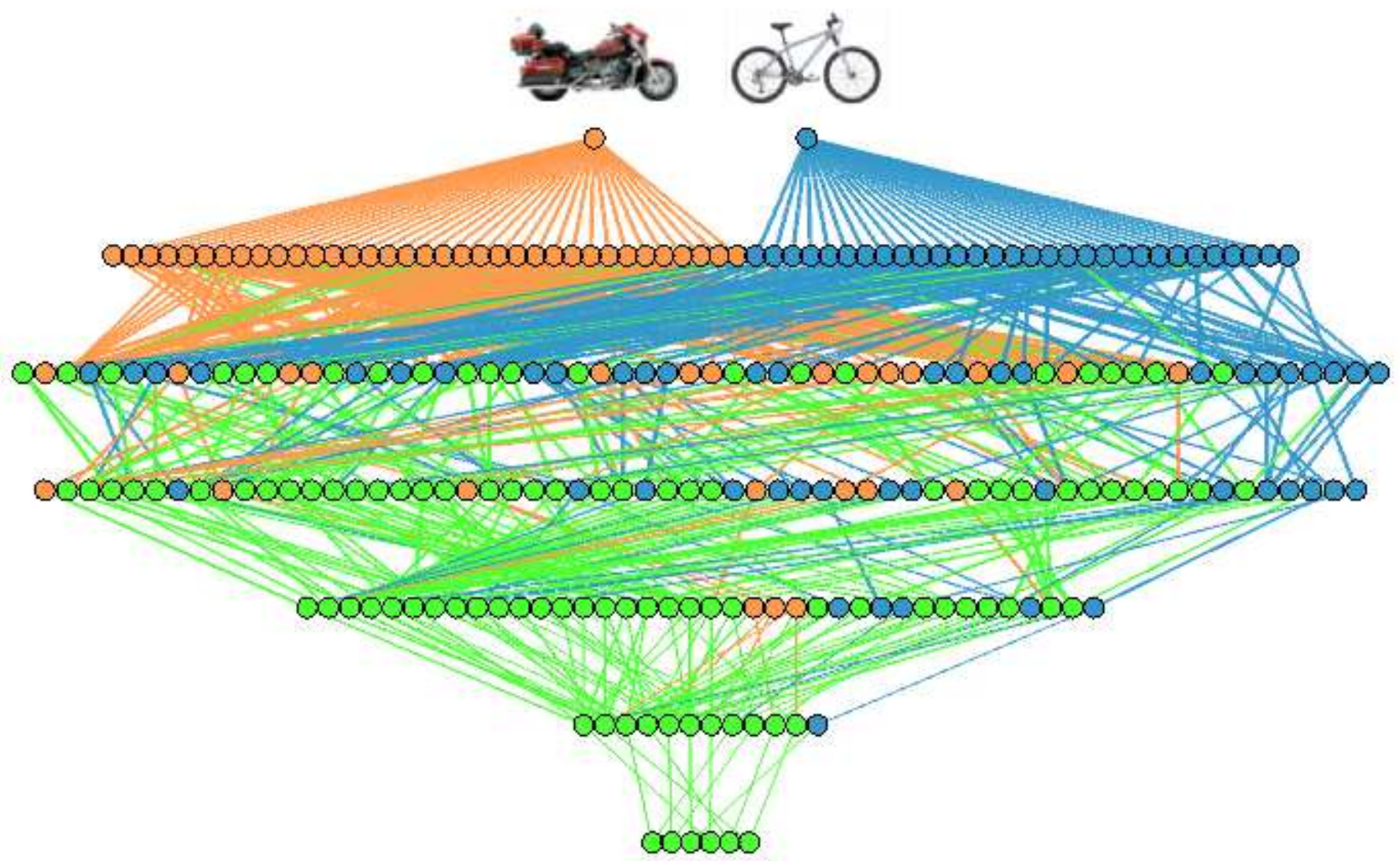} & (b) &
\includegraphics[height=\IH]{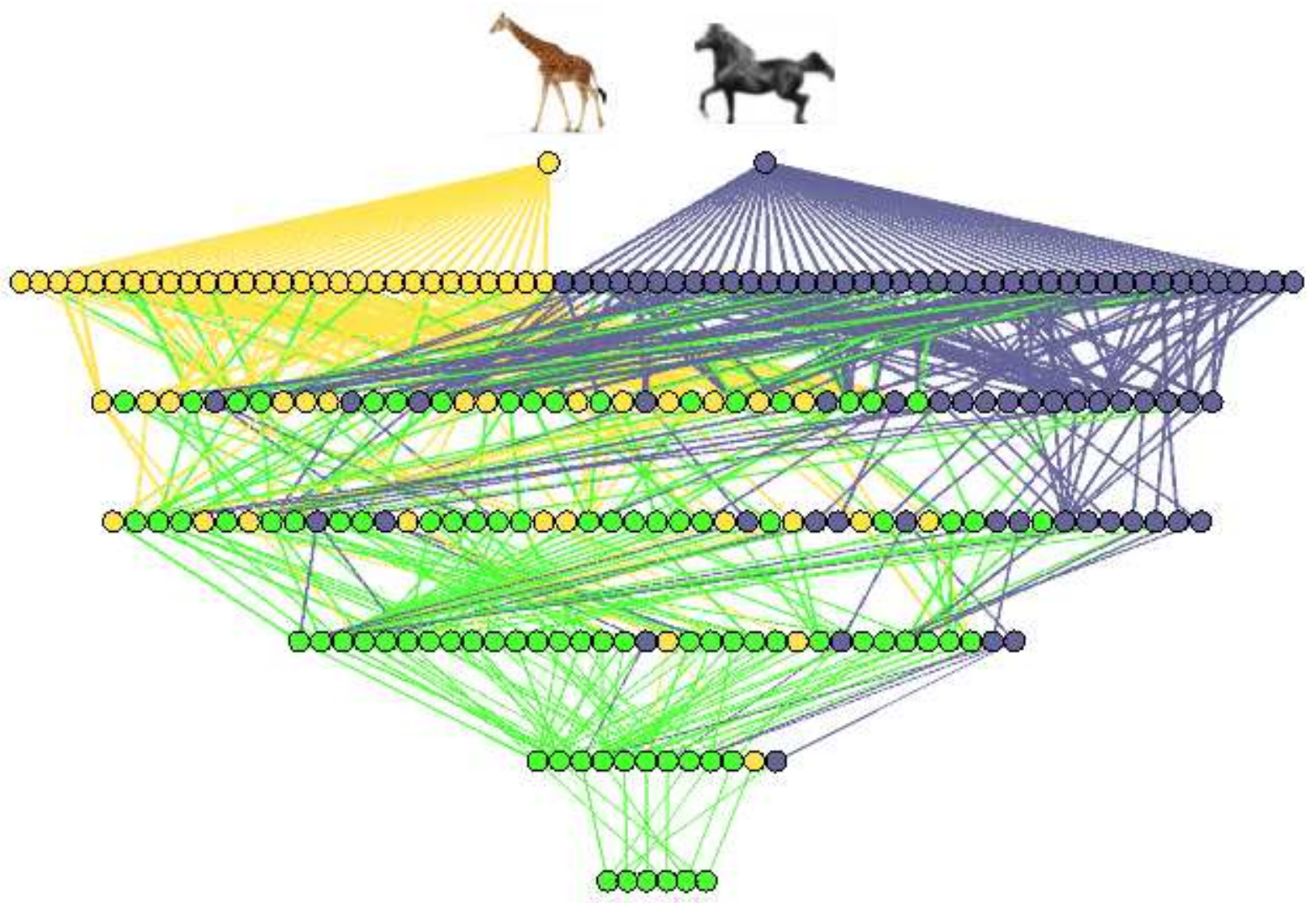} & (c) &
\includegraphics[height=\IH]{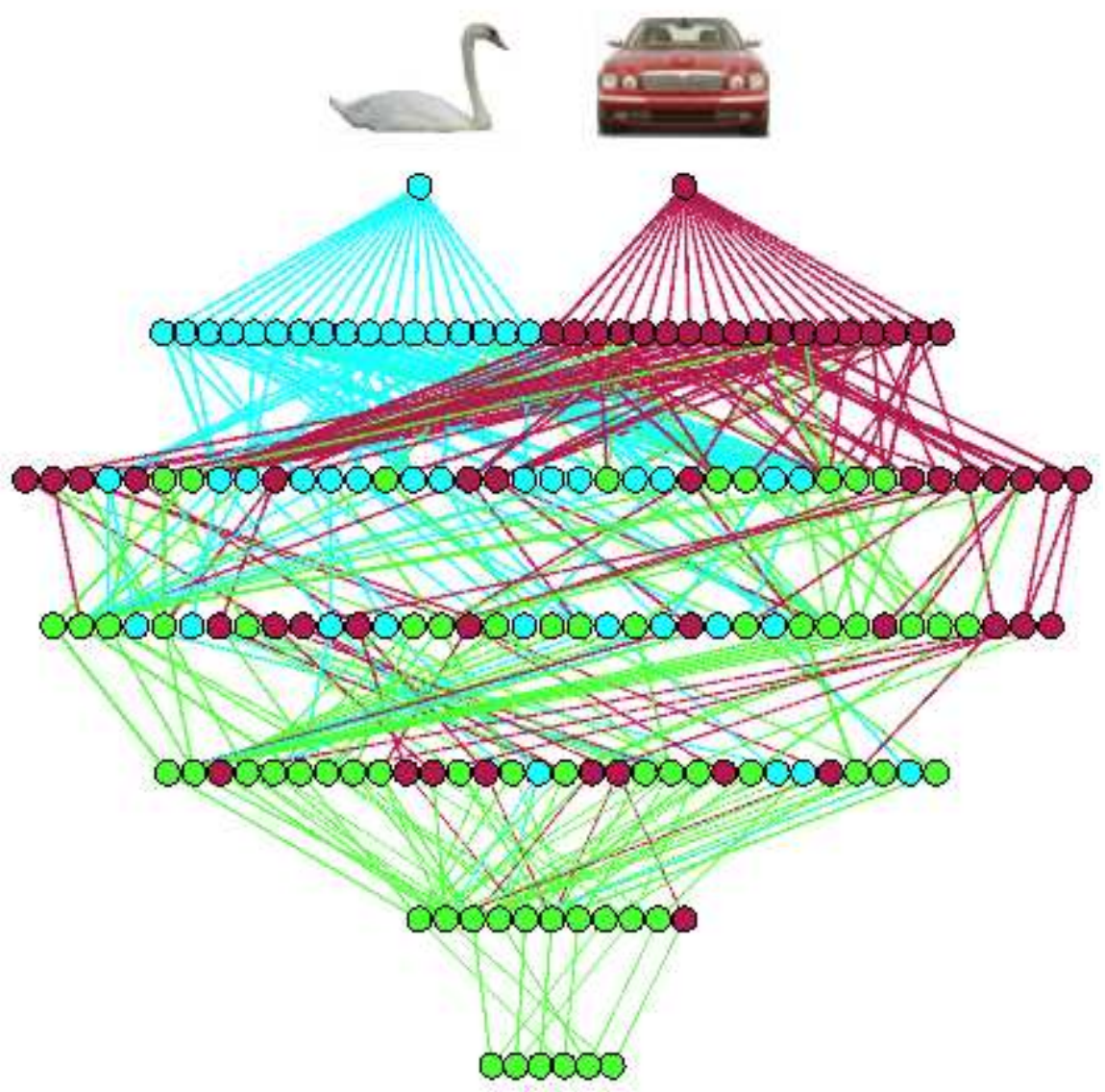}\\
\end{tabular}
\vspace{-1.9mm} \caption{{\small Sharing of compositions between
classes. Each plot shows a \emph{vocabulary} learned for two object
classes. The bottom nodes in each plot represent the $6$ edge
orientations, one row up denotes the learned second layer
compositions, etc. The sixth row (from bottom to top) are the
whole-shape, object-layer compositions (i.e., different aspects of
objects in the classes), while the top layer is the class layer
which pools the corresponding object compositions together.  The
green nodes denote the \emph{shared} compositions, while specific
colors show class-specific compositions. \emph{From left to right:}
Vocabularies for: {\bf (a)} two visually similar classes (motorbike
and bicycle), {\bf (b)} two semi-similar object classes (giraffe and
horse), {\bf (c)} two visually dissimilar classes (swan and
car\_front). Notice that even for dissimilar object classes, the
compositions from the lower layers are shared between them.} 
}
 \label{fig:sharing}
\end{figure*}

\def\IH{3.668cm}
\begin{figure*}[htb!]
\centering
\includegraphics[height=\IH,trim=6 3 35 0,clip=true]{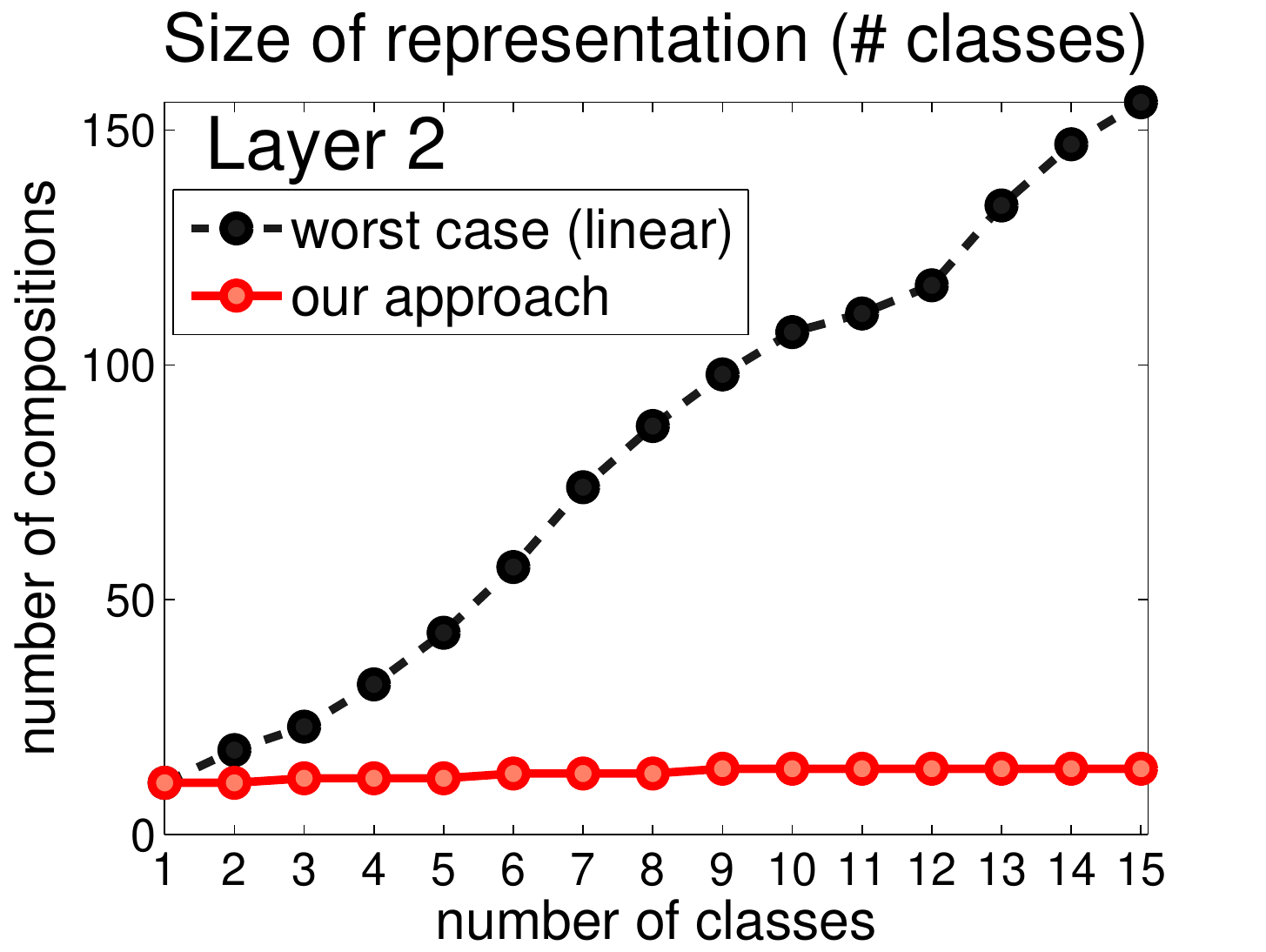}
\includegraphics[height=\IH,trim=5 3 35 0,clip=true]{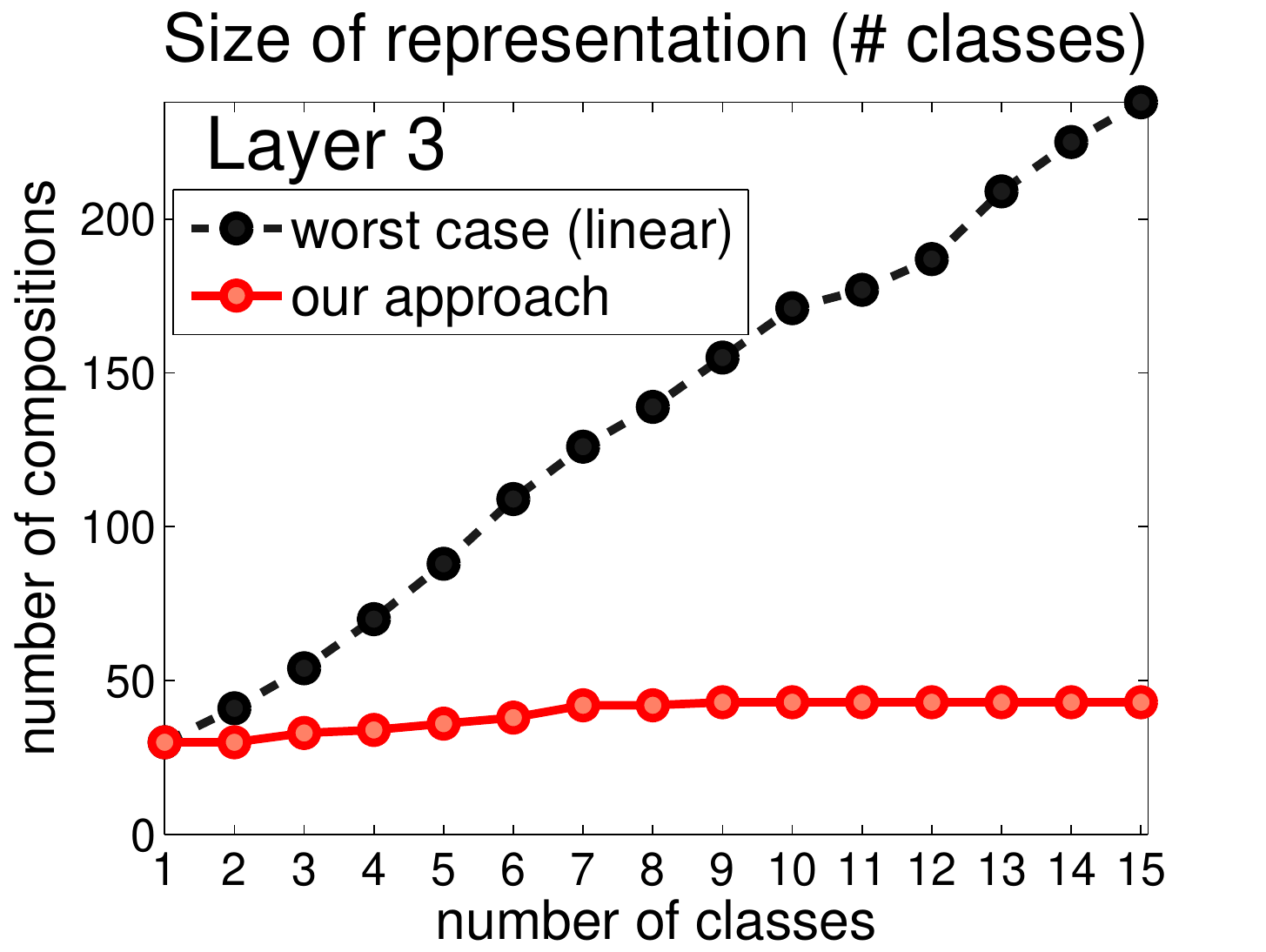}
\includegraphics[height=\IH,trim=5 3 35 0,clip=true]{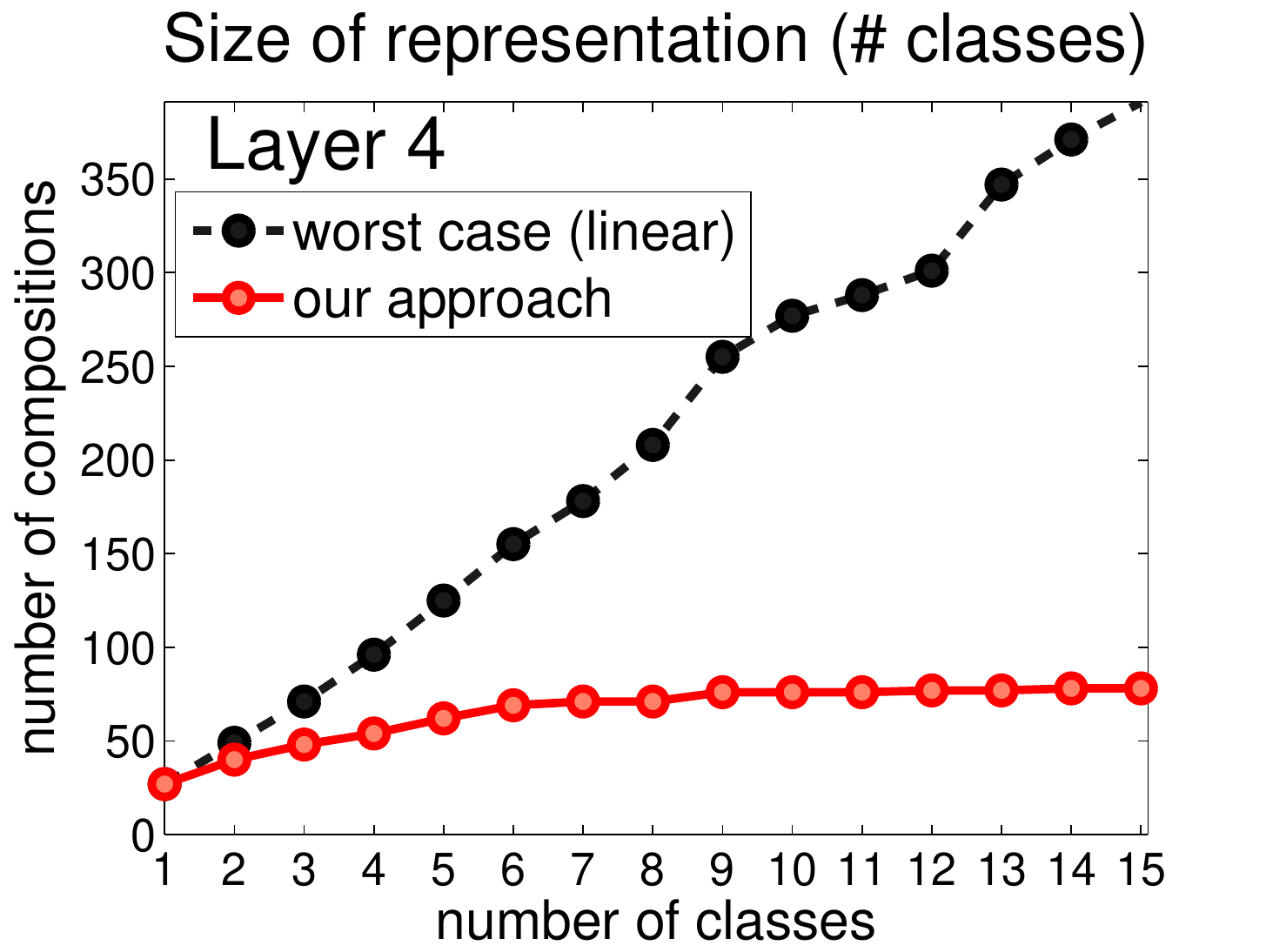}
\includegraphics[height=\IH,trim=5 3 35 0,clip=true]{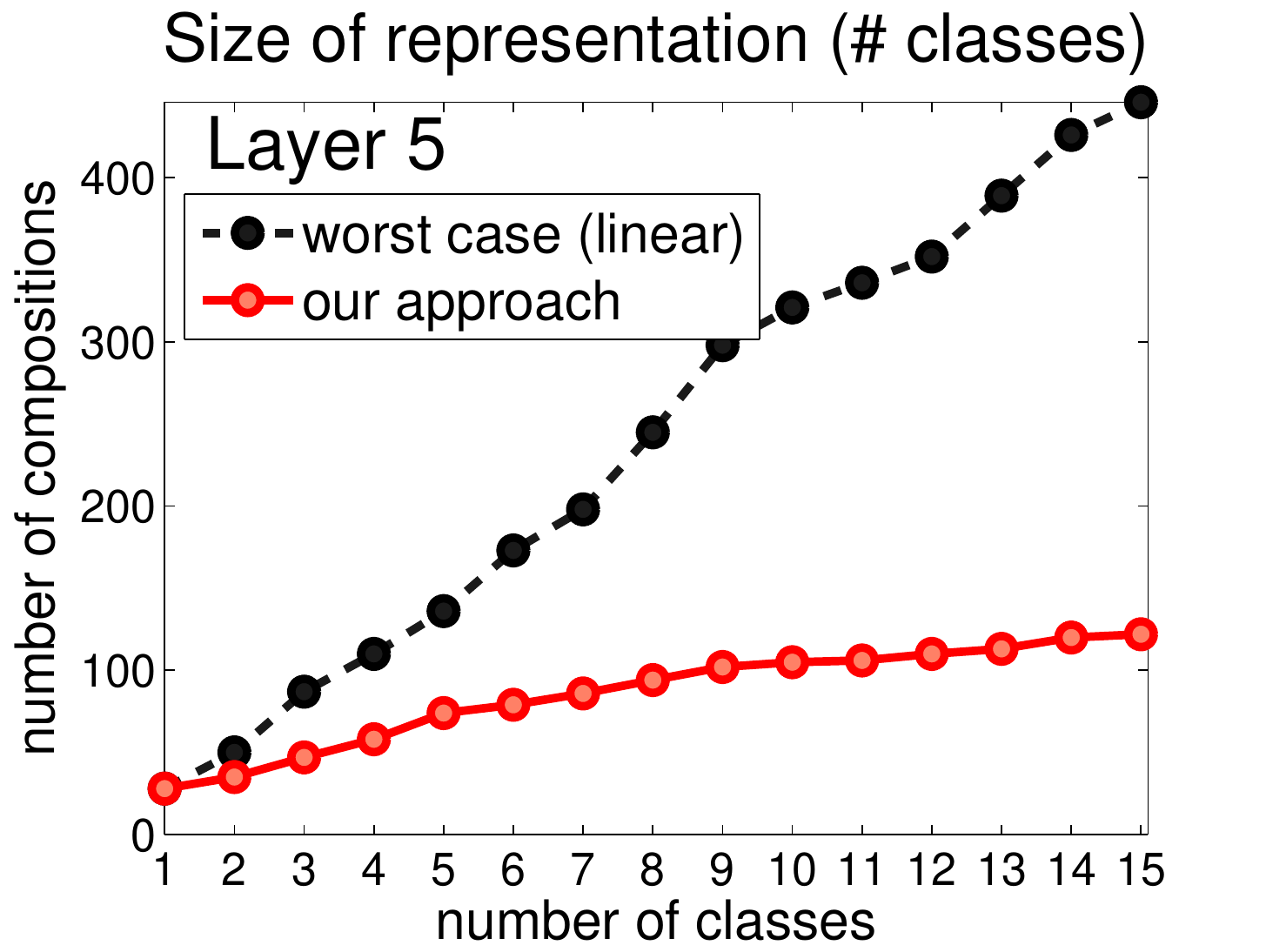}

\vspace{-1.3mm} \caption{{\small Size of representation (the number
of compositions per layer) as a function of the number of learned
classes. ``Worst case'' denotes the sum over the sizes of
single-class vocabularies.}}
 \label{fig:size_representation}
\end{figure*}

\def\IH{3.646cm}
\begin{figure*}[htb!]
\centering
\addtolength{\tabcolsep}{-5pt}
\begin{tabular}{cccc}
\includegraphics[height=\IH,trim=5 0 35
3,clip=true]{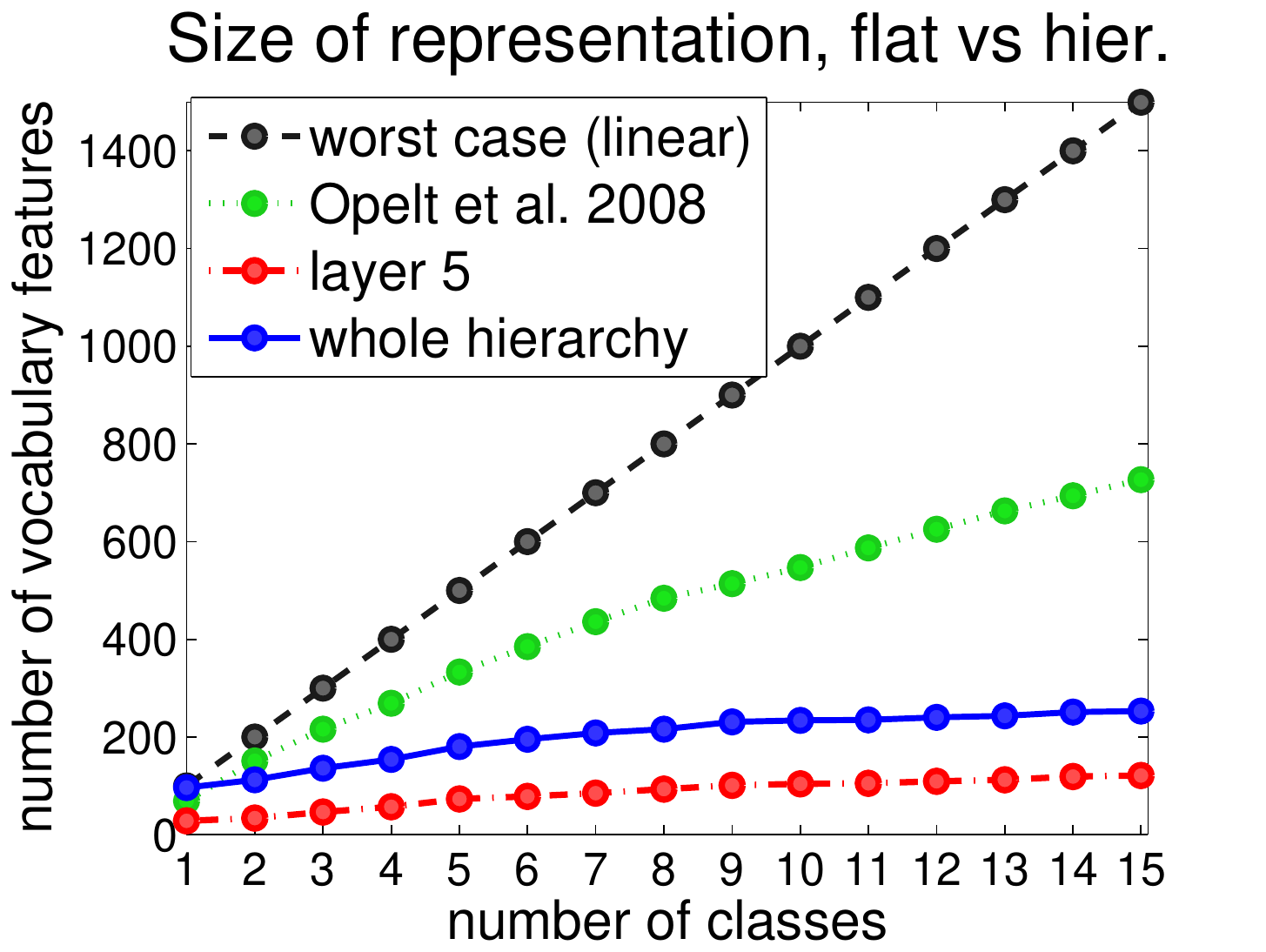} &
\includegraphics[height=\IH,trim=5 0 29
3,clip=true]{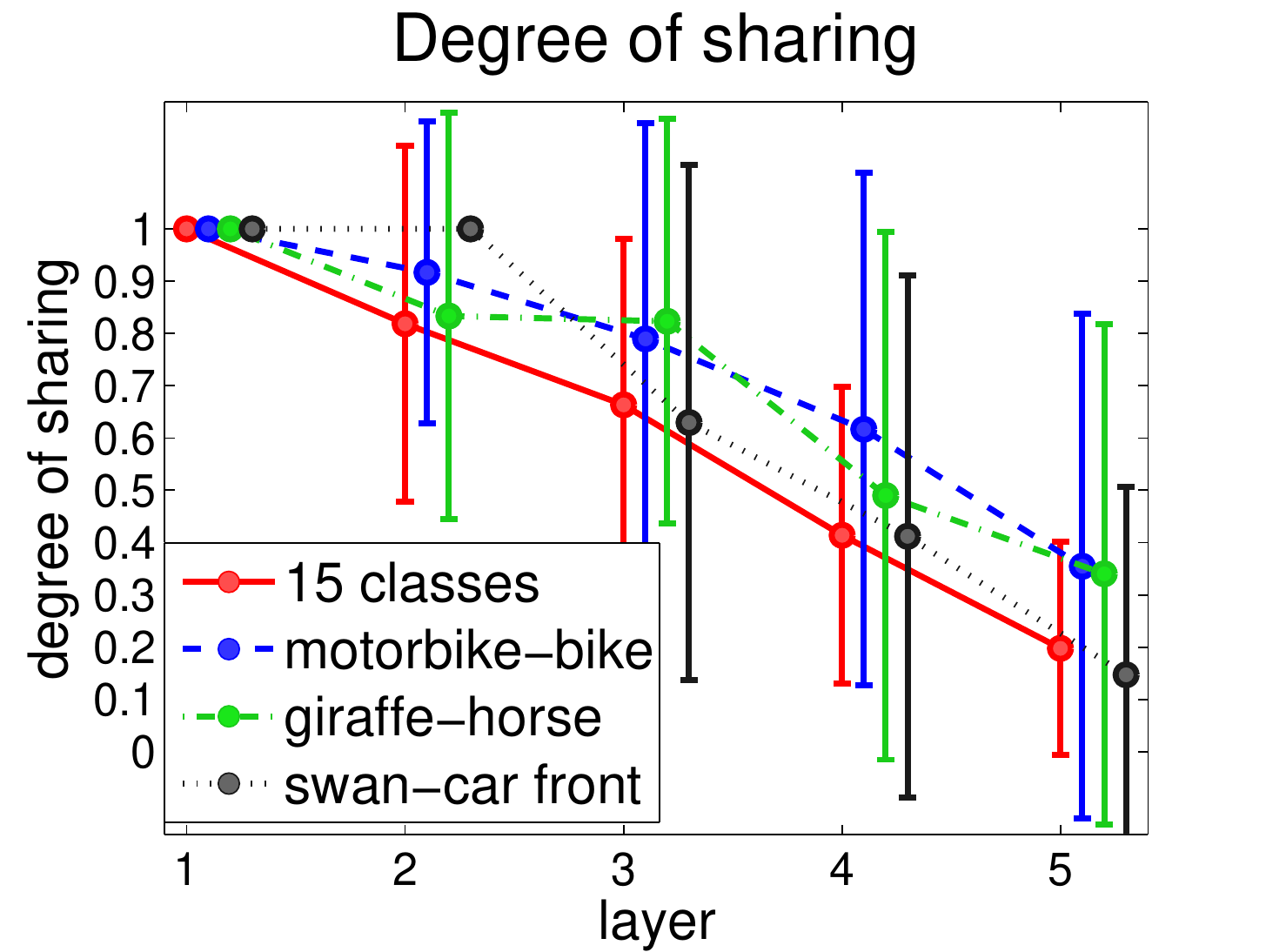} &
\includegraphics[height=\IH,trim=15 0 26
3,clip=true]{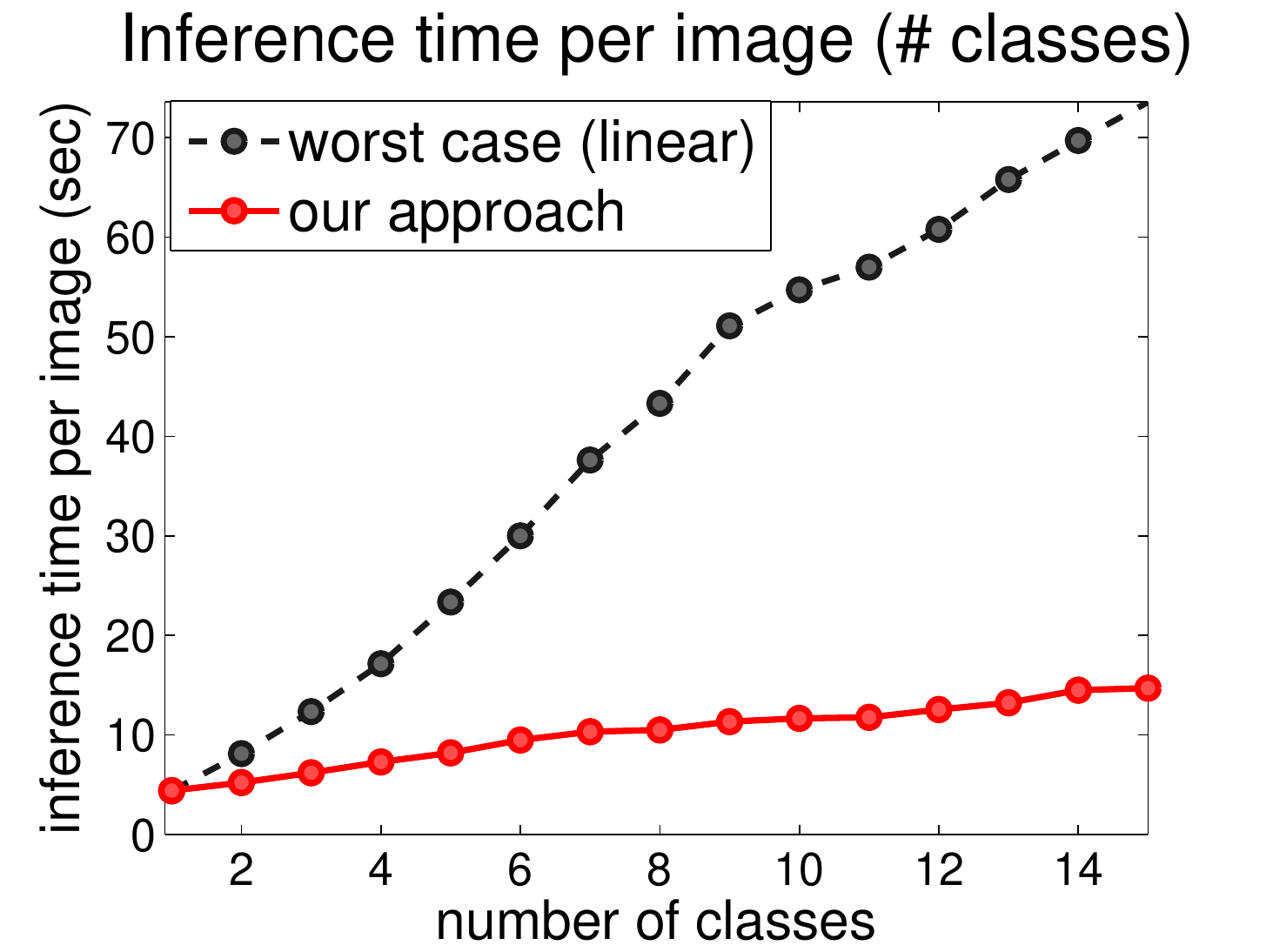} &
\includegraphics[height=\IH,trim=5 0 32
3,clip=true]{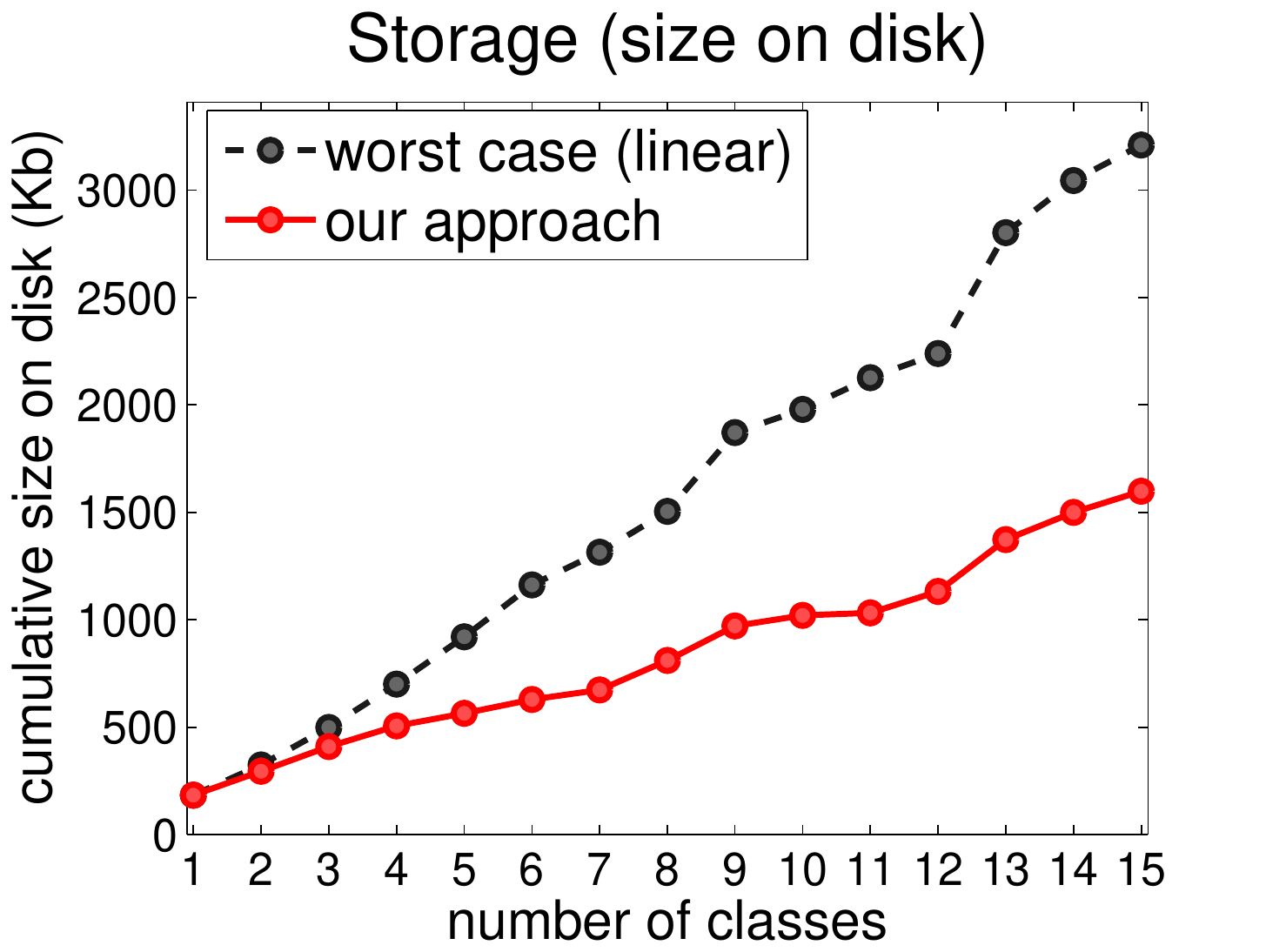}\\
(a) & (b) & (c) & (d)\\
\end{tabular}
\vspace{-1.7mm} \caption{{\small From left to right: {\bf (a)} A
comparison in scaling to multiple classes of the approach by Opelt
et al.~\cite{s:opelt08} (flat) and our hierarchical approach; {\bf
(b)} Degree of sharing for the multi-class vocabulary; {\bf (c)}
Average inference time per image as a function of the number of
learned classes; {\bf (d)} Storage (size of the hierarchy stored on
disk) as a function of the number of learned classes.}}
 \label{fig:inference_time}
 \vspace{-1.9mm}
\end{figure*}

\begin{table*}[htb!]
\addtolength{\tabcolsep}{-0.35pt}
\begin{small}
\centering
 \caption{Comparison in detection accuracy for single- and multi-class object representations and detection procedures.}
 \vspace{-3mm}
  \label{table:comparison_multi}
\addtolength{\tabcolsep}{-2.05pt}
\begin{tabular}{|c|r|r|r|r|r|r|r|r|r|r|r|r|r|r|r|} \whline
 \hline
class & Apple & bottle & giraffe & mug & swan & horse & cow & mbike & bicycle & car\_front & car\_side & car\_rear & face & person & cup\\
\hline
measure &  \multicolumn{5}{c|}{detection rate (in \%) at 0.4 FPPI} & \multicolumn{10}{c|}{recall (in \%) at EER}\\
 \whline
single & $88.6$ & $85.5$  & $83.5$  & $84.9$  &  $75.8$ & $84.5$ &
$96.9$ & $83.2$ & $68.5$ & $76.5$ & $97.5$ & $93.5$ & $94.0$ & $60.4$ & $85.0$\\
 \hline
 multi & $84.1$ & $80.0$ & $80.2$ & $80.3$ & $72.7$ & $82.3$ & $95.4$ & $83.2$ & $64.8$ & $82.4$ &
 $93.8$ & $92.0$ & $93.0$ &  $62.5$ & $85.0$\\
 \hline
\end{tabular}
\end{small}
\end{table*}

\def\IH{4.58cm}
\begin{figure*}[htb!]
\centering
\addtolength{\tabcolsep}{-5pt}
\begin{tabular}{lclclc}
(a) &
\includegraphics[height=\IH,trim=3 3 3 0,clip=true]{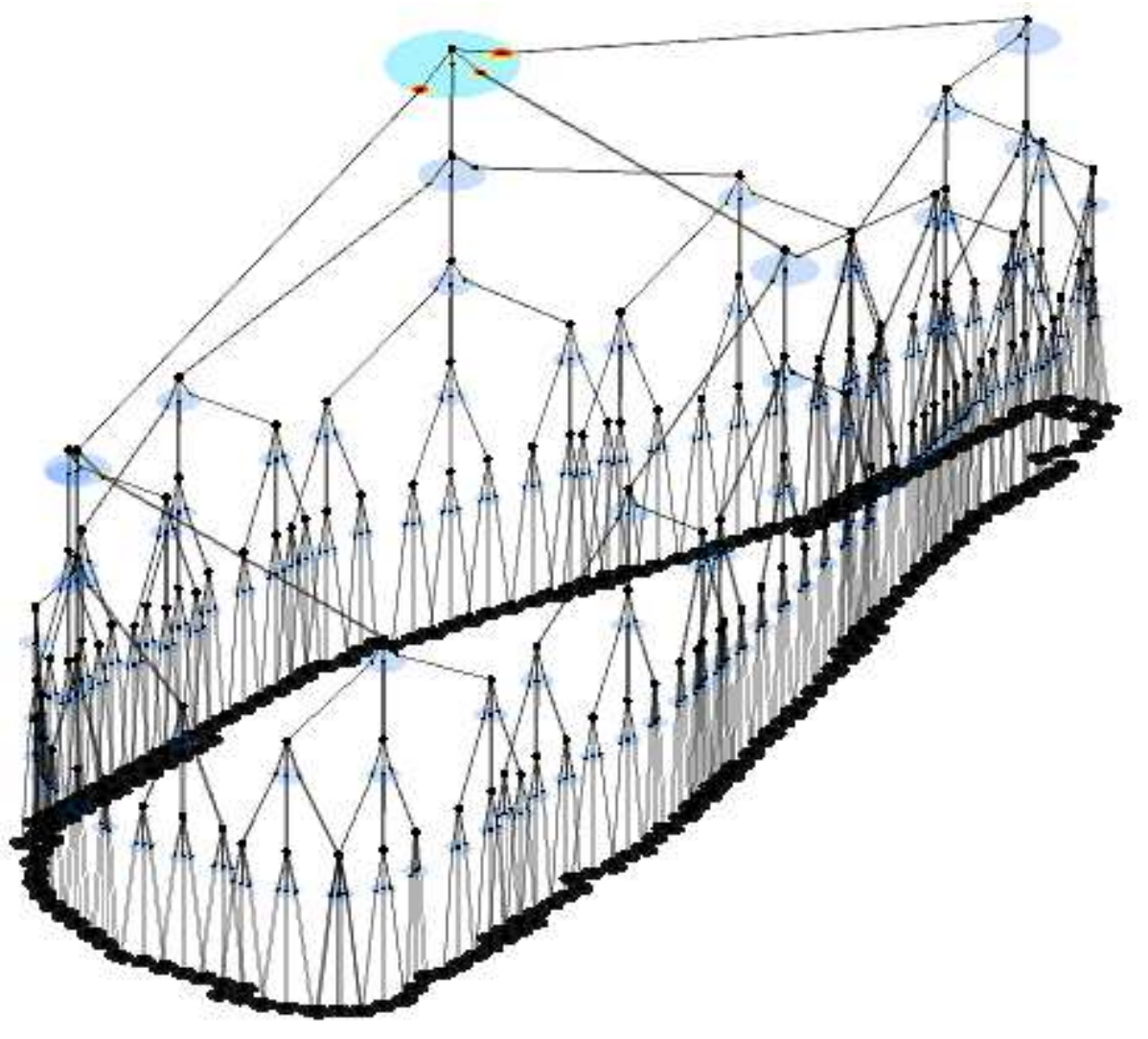} & (b) &
\includegraphics[height=\IH,trim=30 3 50 0,clip=true]{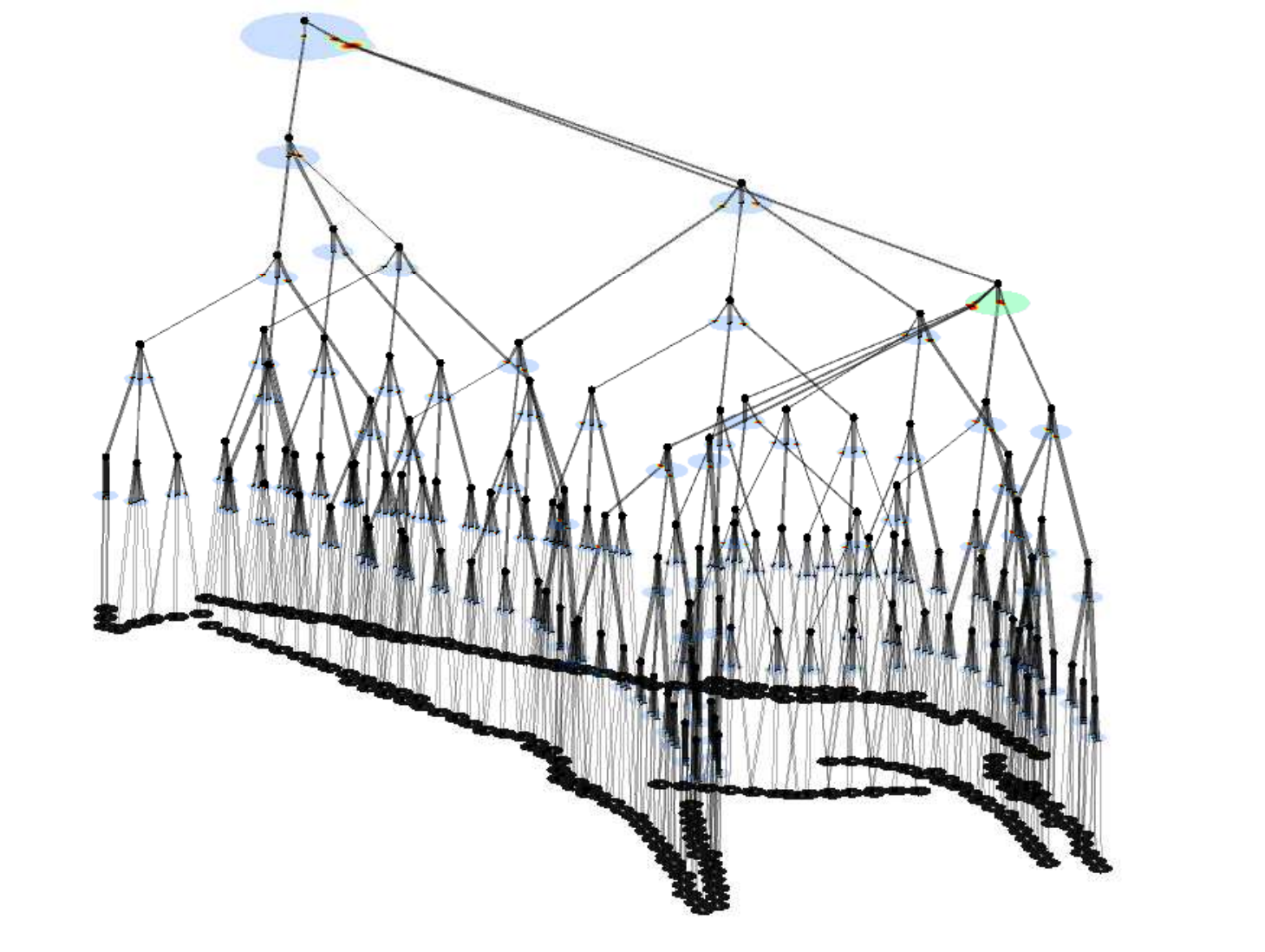}\hspace{2mm} & (c)
\hspace{-1mm}\includegraphics[height=\IH,trim=5 3 3 10,clip=true]{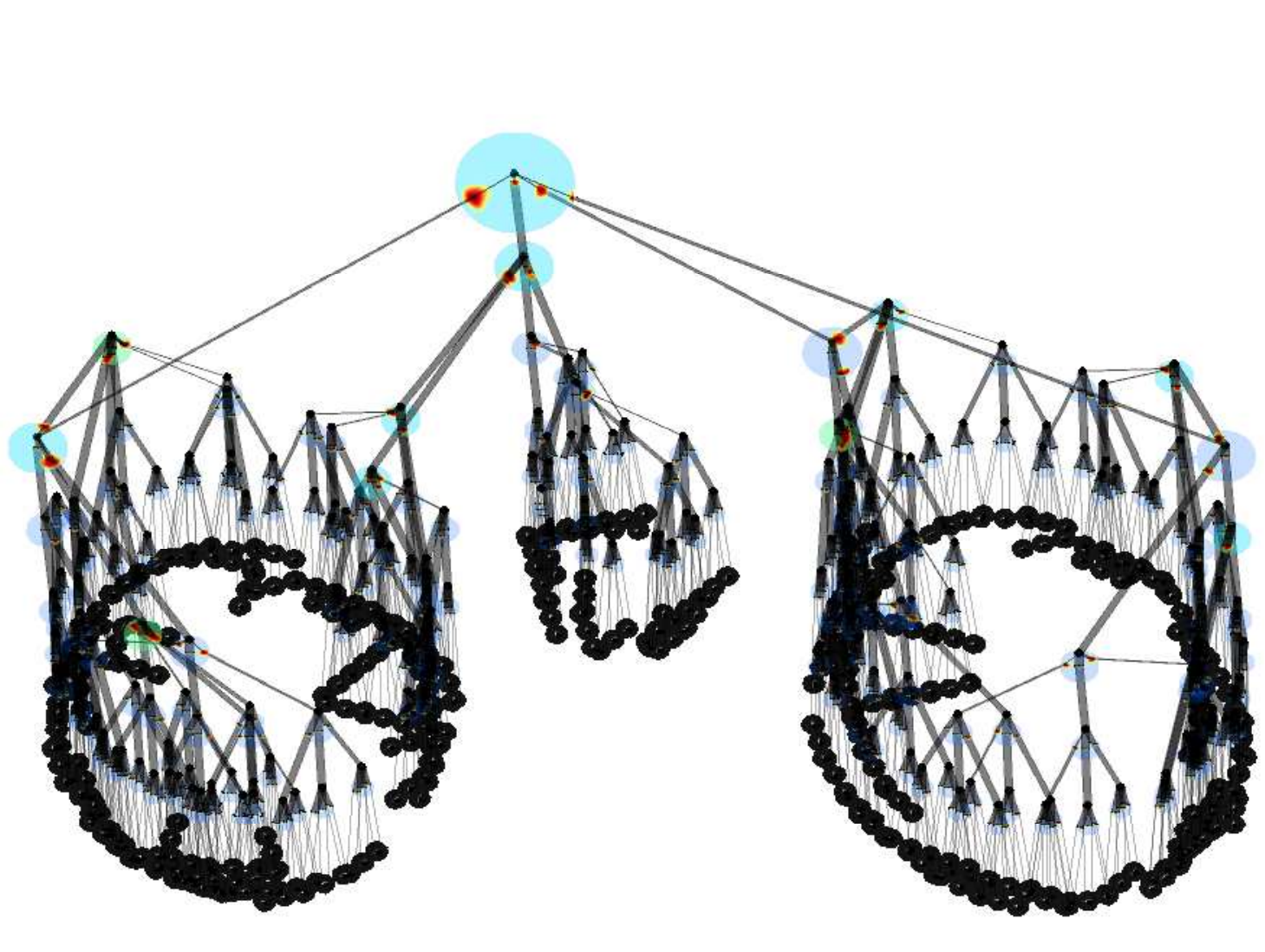}\\
\end{tabular}
\vspace{-2.mm} \caption{{\small One learned object-layer composition
$\model^\cO$ for {\bf (a)} \emph{bottle}, {\bf (b)} \emph{giraffe},
and {\bf (c)} \emph{bicycle}, with the complete model structure
shown. The blue patches denote the limiting circular regions, the
color regions inside them depict the spatial relations (Gaussians).
The nodes are the compositions in the vocabulary. Note that each
composition models the distribution over the compositions from the
previous layer for the appearance of its parts. For clarity, only
the most likely composition for the part is decomposed further.}}
 \label{fig:models}
\end{figure*}

\def\IH{2.75cm}
\def\IW{3.58cm}
\begin{figure*}[htb!]
\centering
\includegraphics[height=\IH,width=\IW]{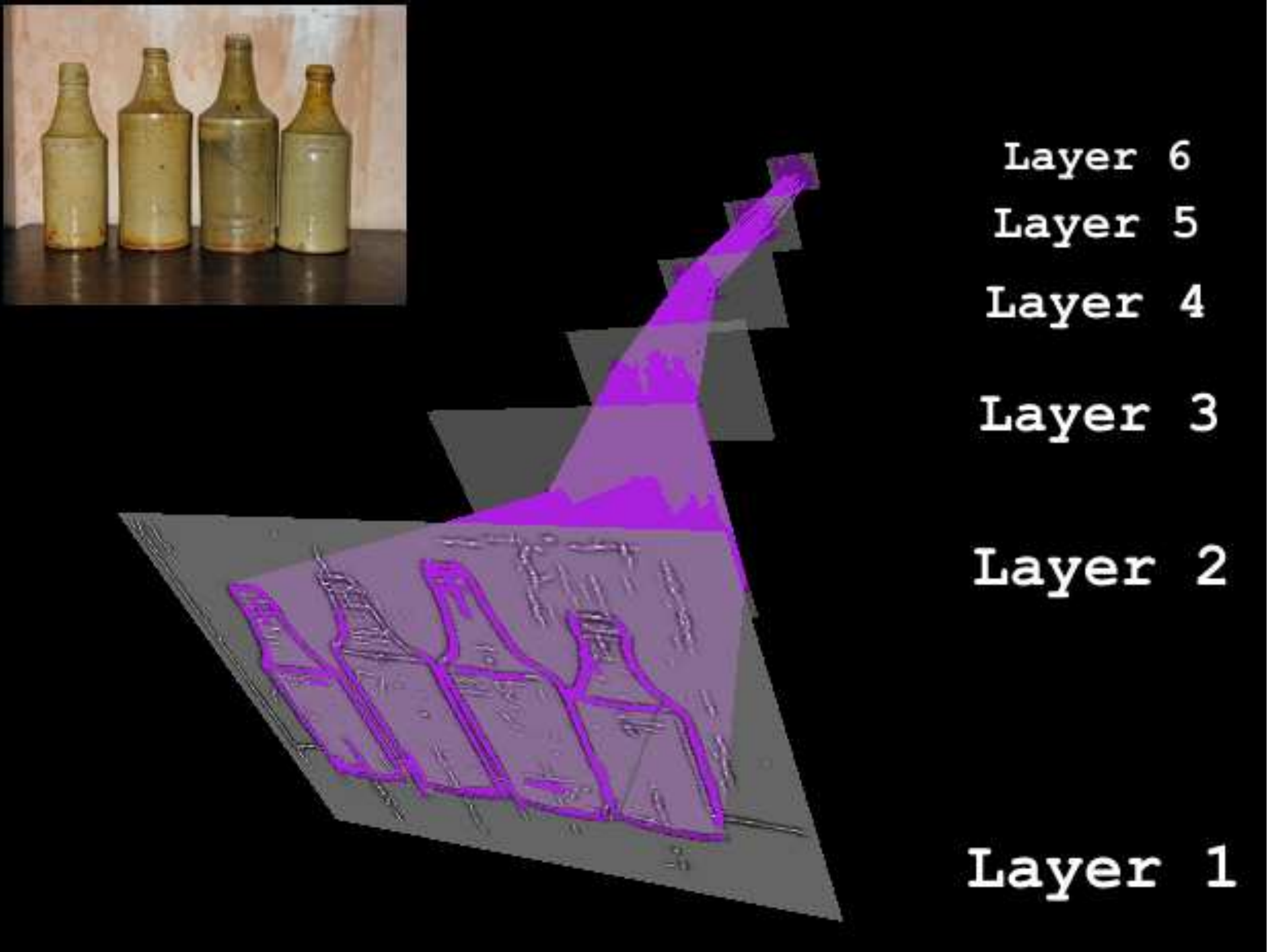}\hspace{-0.5mm}
\includegraphics[height=\IH,width=\IW]{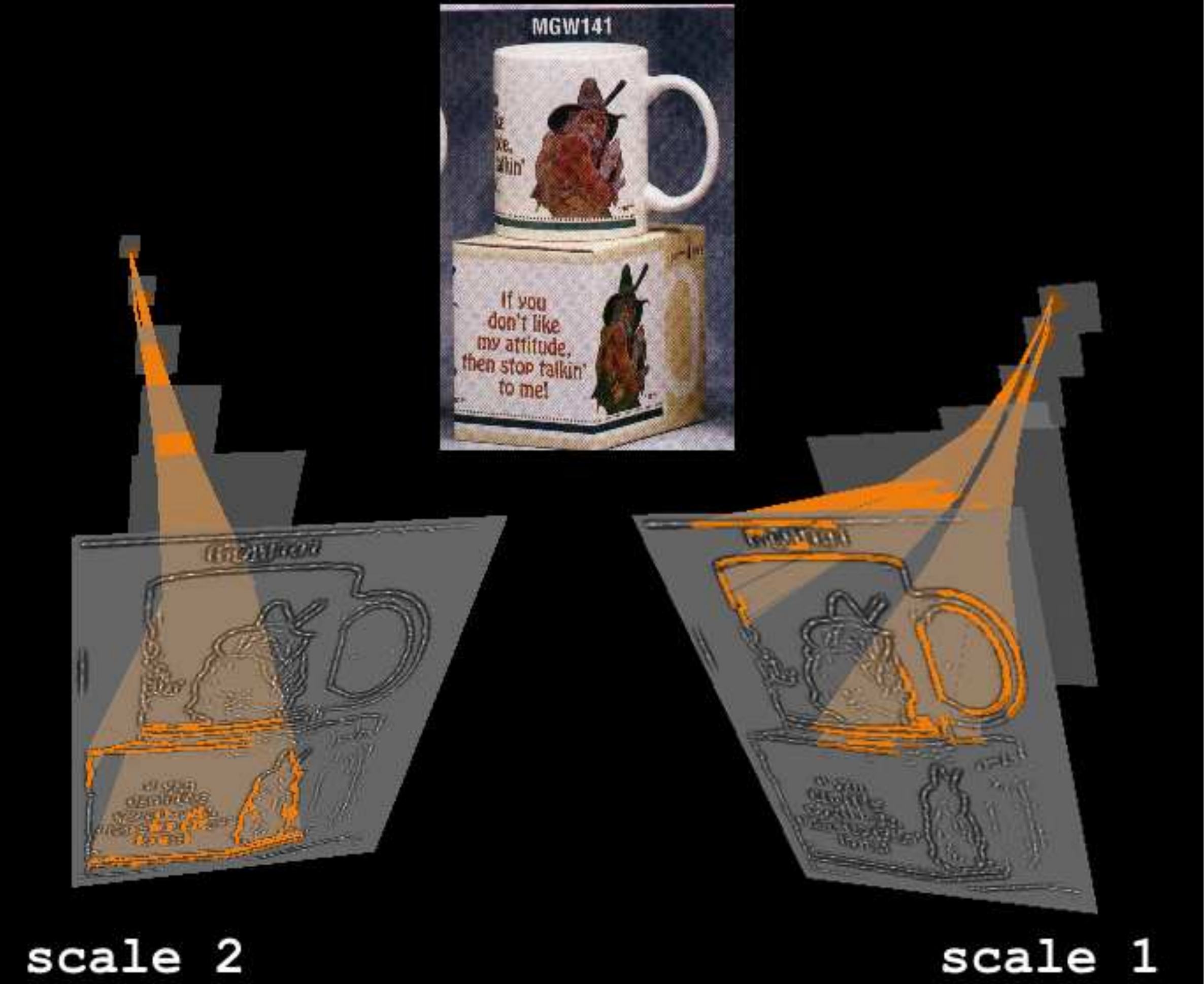}\hspace{-0.5mm}
\includegraphics[height=\IH,width=\IW]{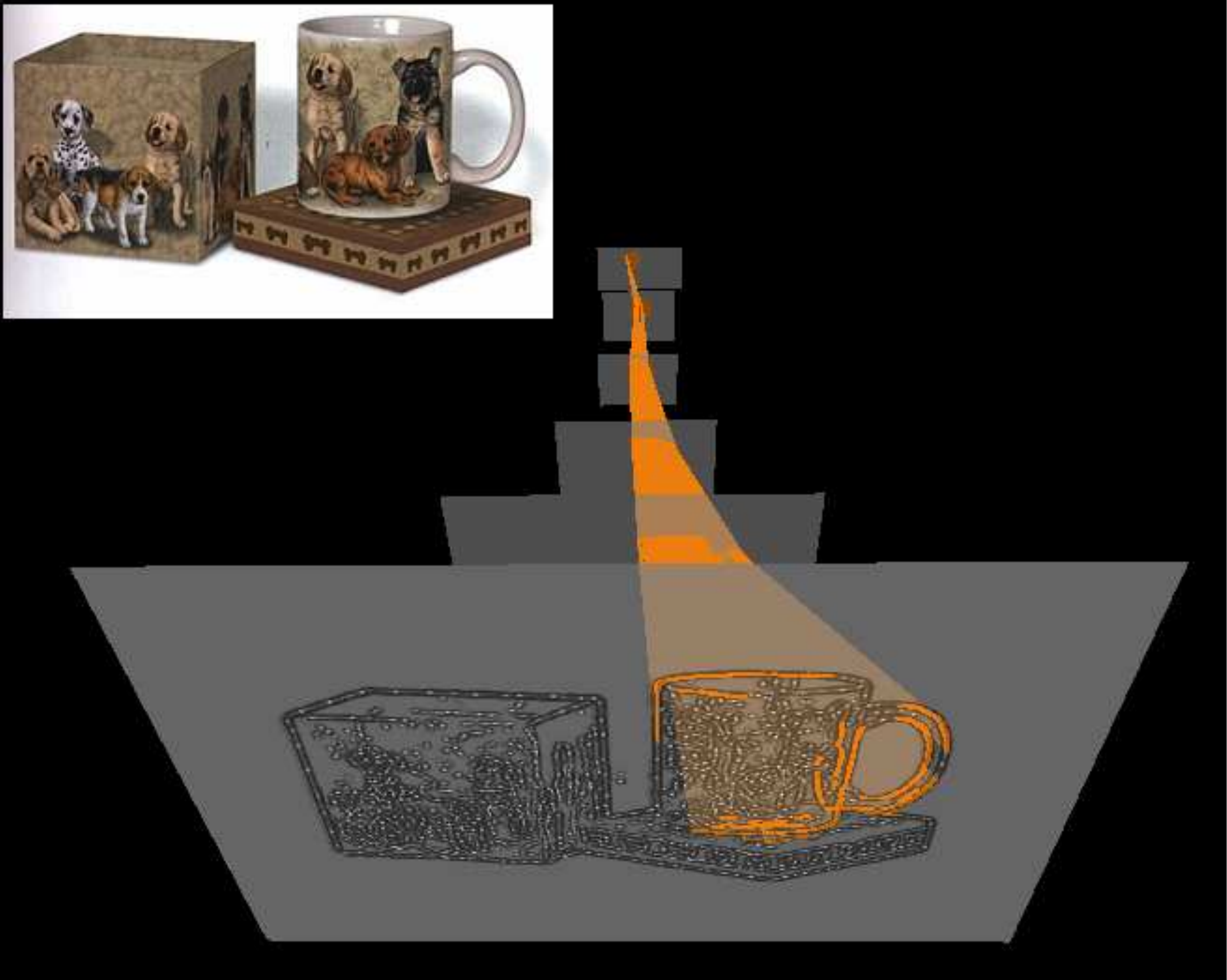}\hspace{-0.5mm}
\includegraphics[height=\IH,width=\IW]{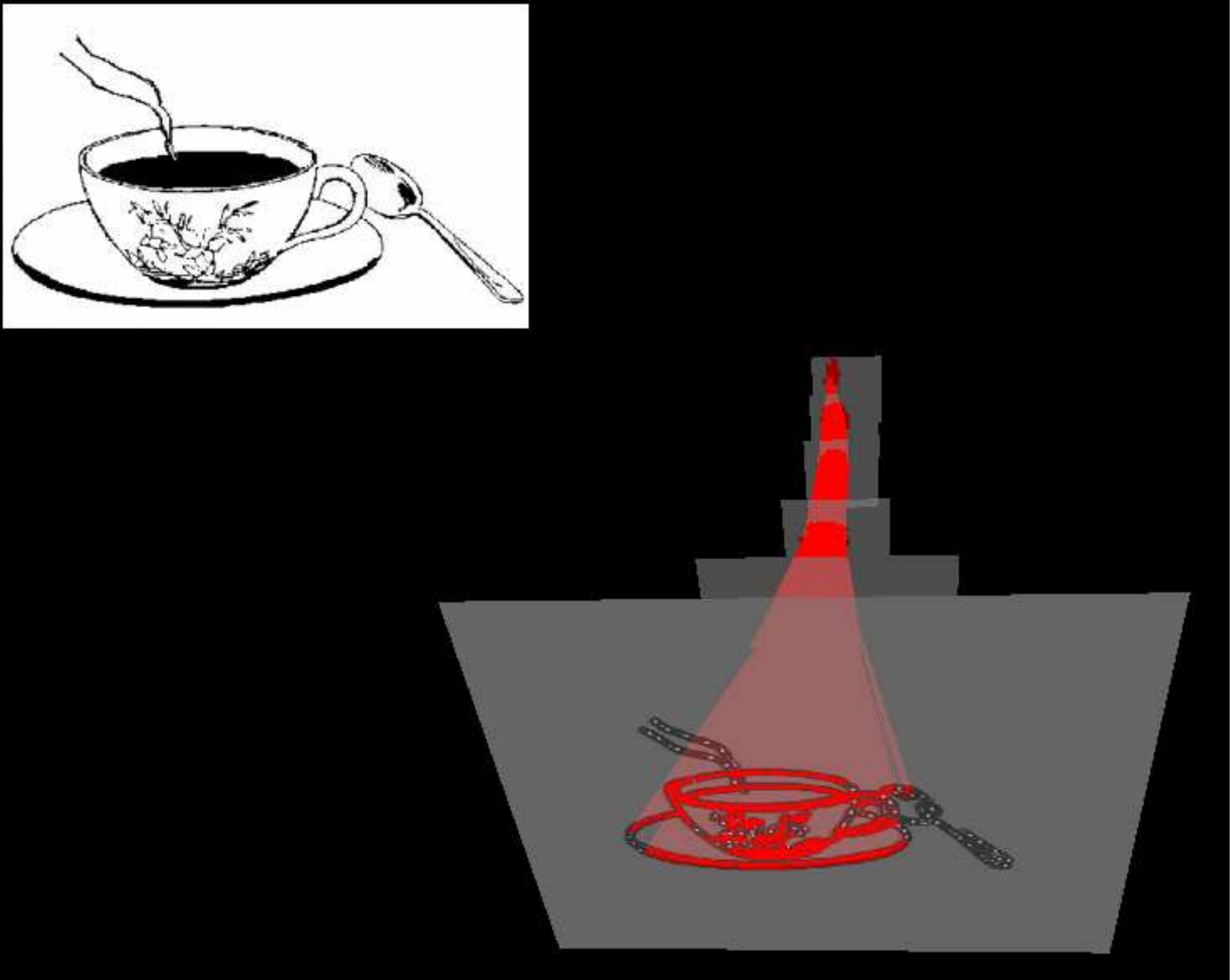}\hspace{-0.5mm}
\includegraphics[height=\IH,width=\IW]{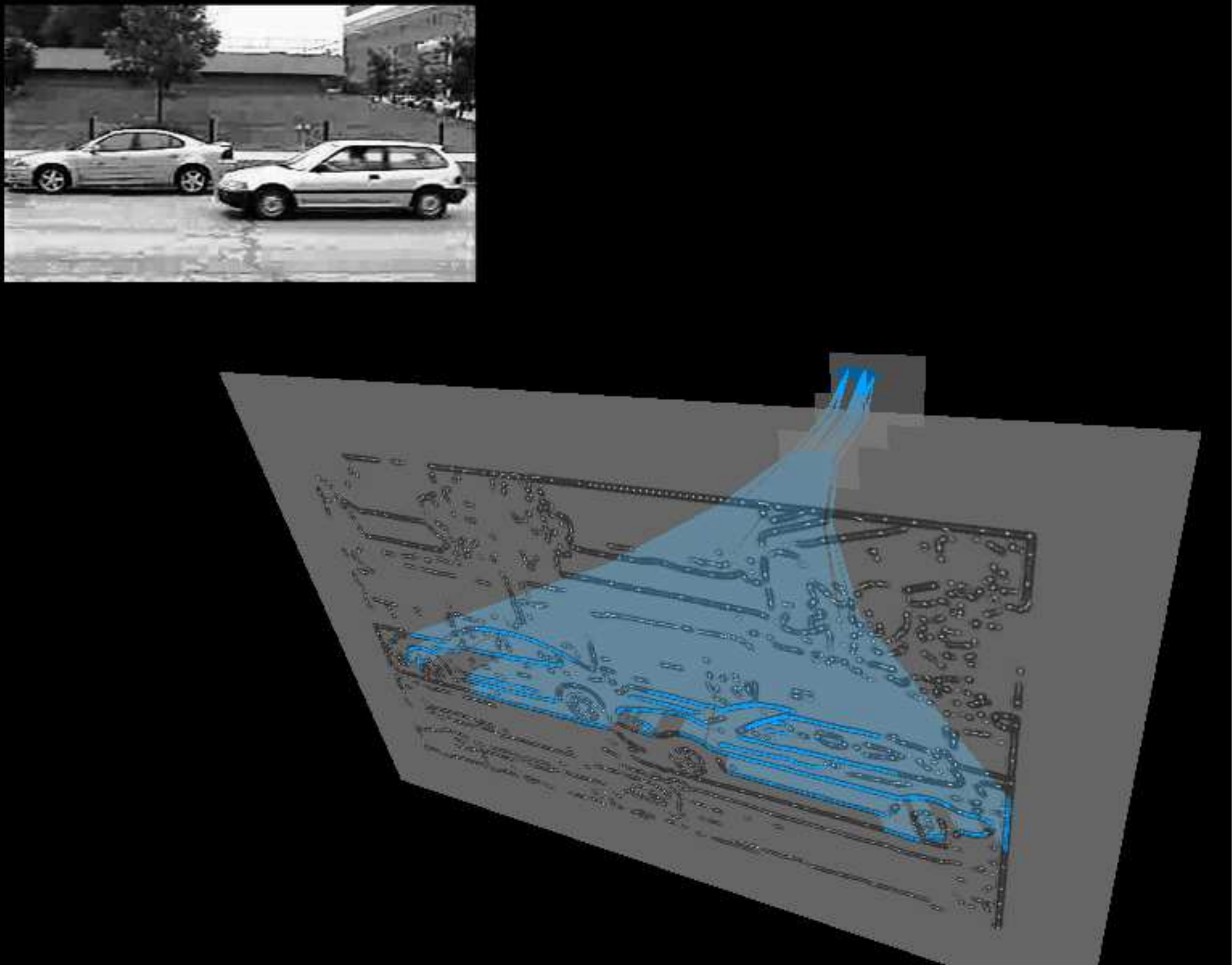}\hspace{-0.5mm}\\[0.5mm]
\includegraphics[height=\IH,width=\IW]{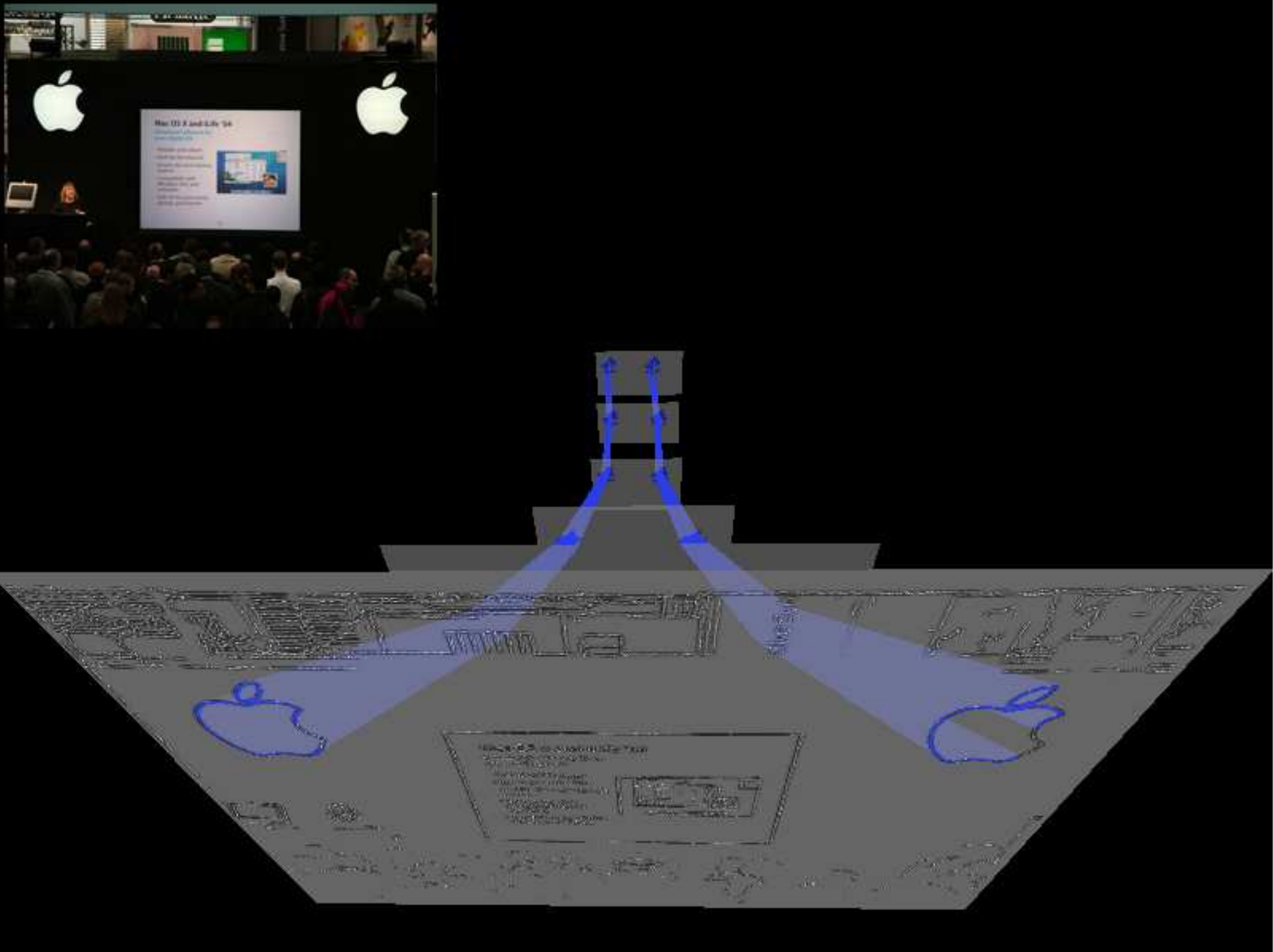}\hspace{-0.5mm}
\includegraphics[height=\IH,width=\IW]{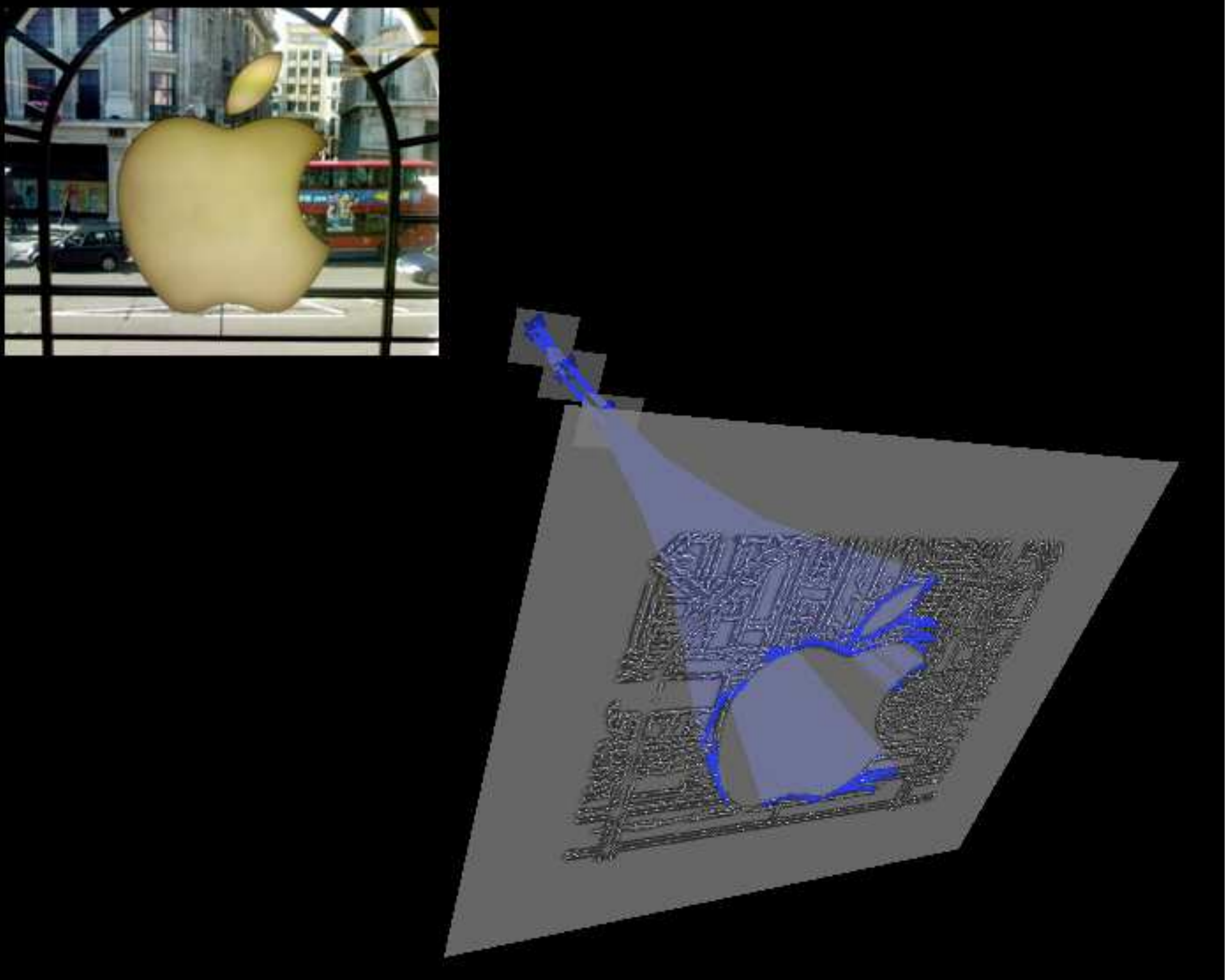}\hspace{-0.5mm}
\includegraphics[height=\IH,width=\IW]{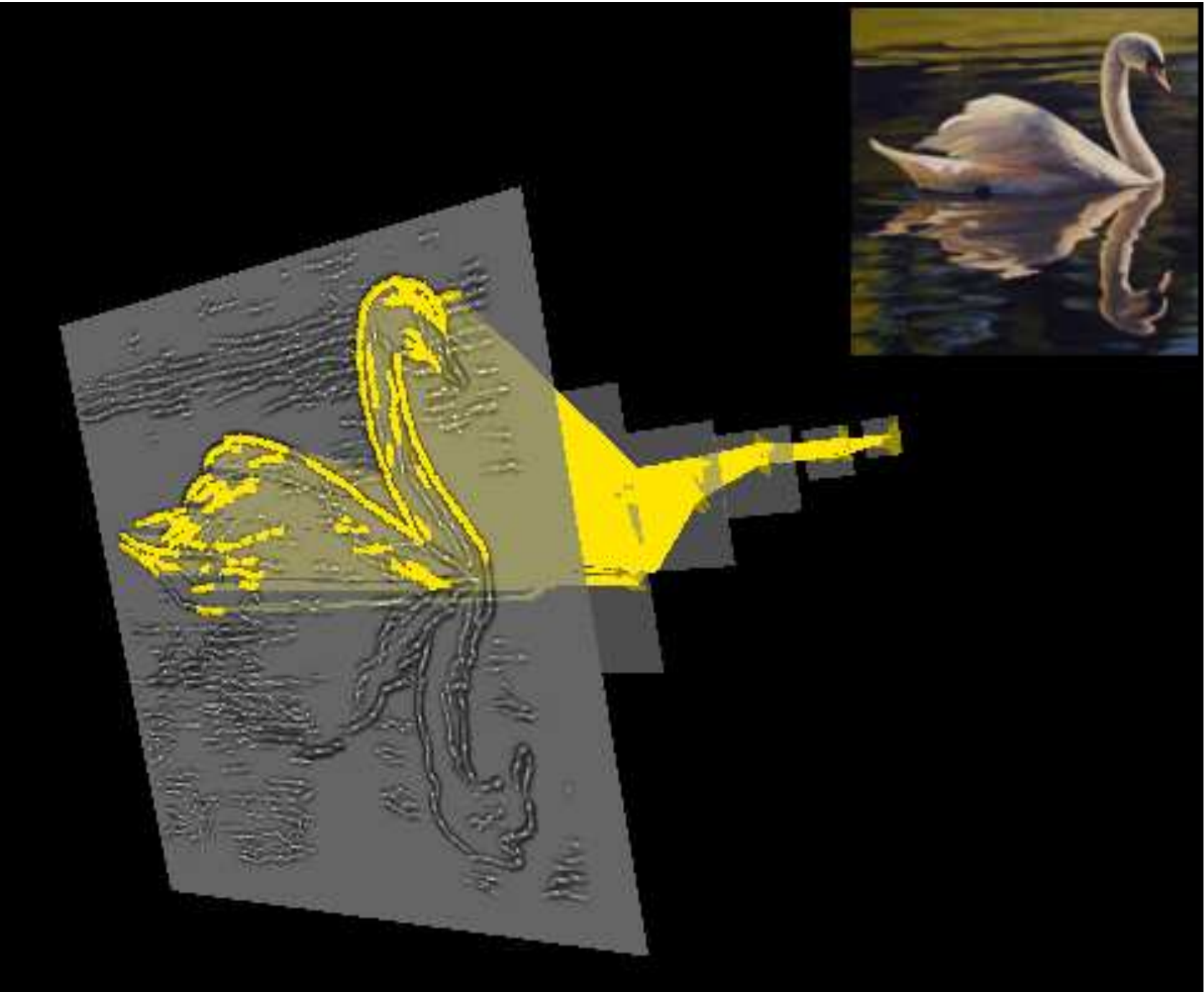}\hspace{-0.5mm}
\includegraphics[height=\IH,width=\IW]{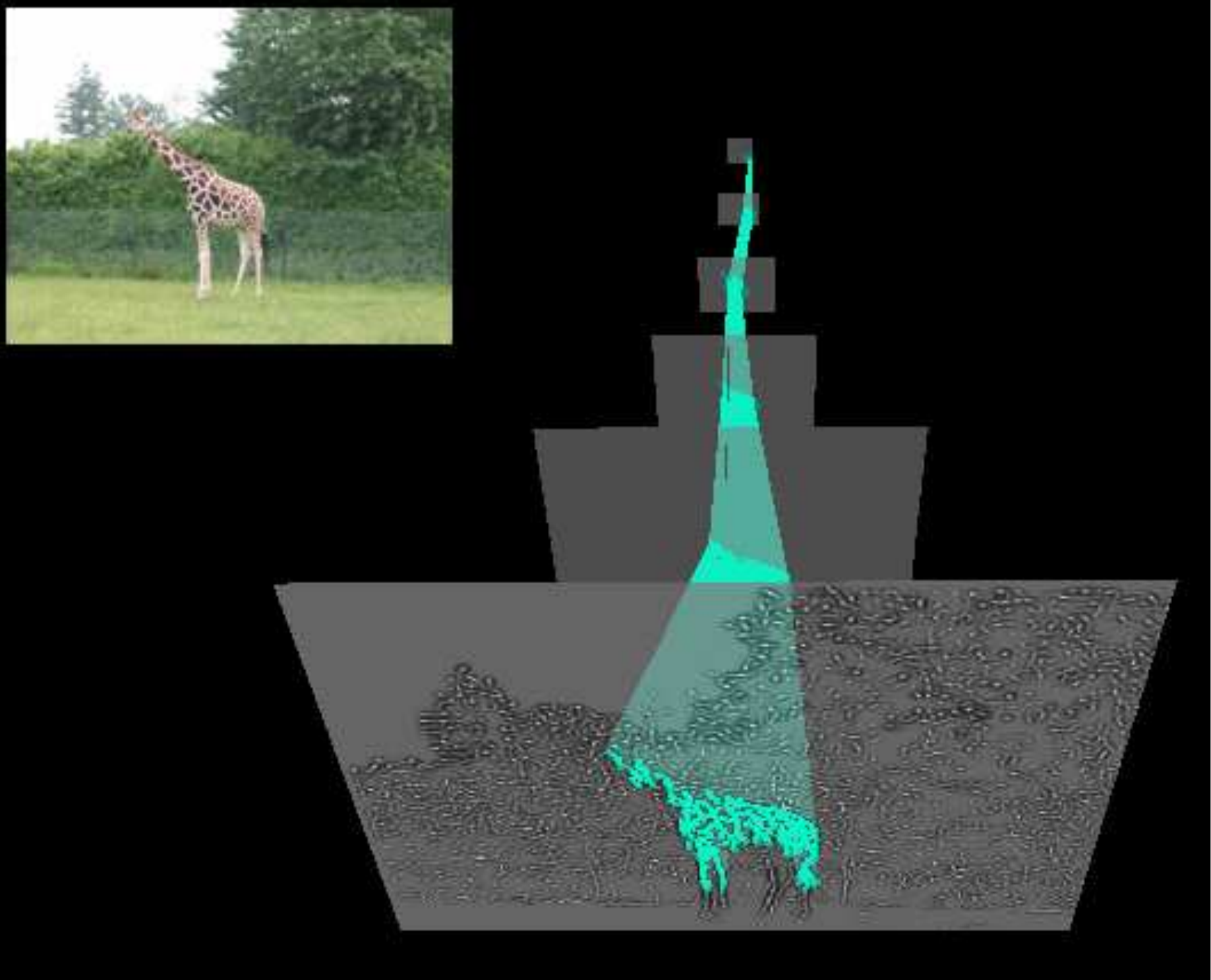}\hspace{-0.5mm}
\includegraphics[height=\IH,width=\IW]{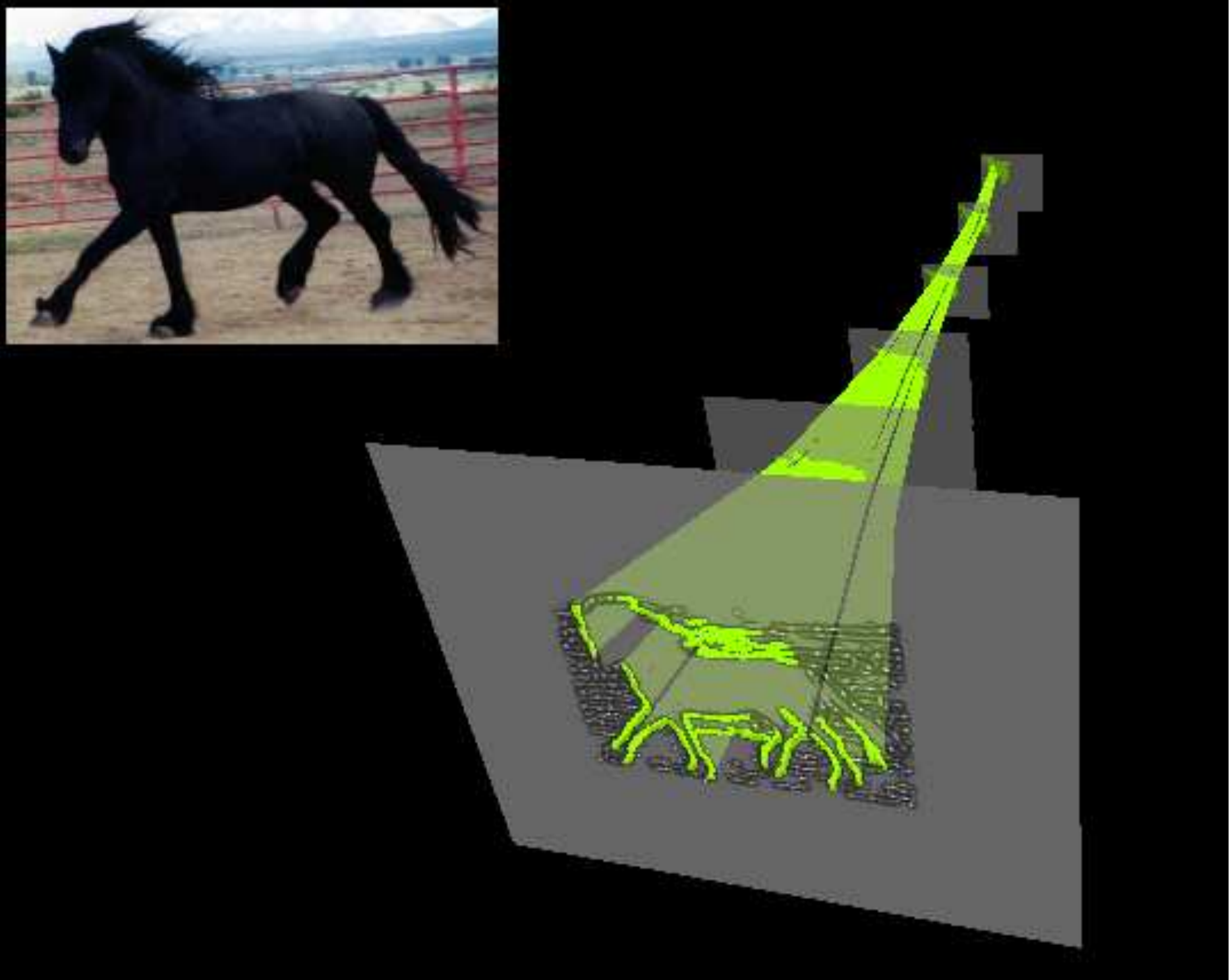}\hspace{-0.5mm}\\[0.5mm]
\includegraphics[height=\IH,width=\IW]{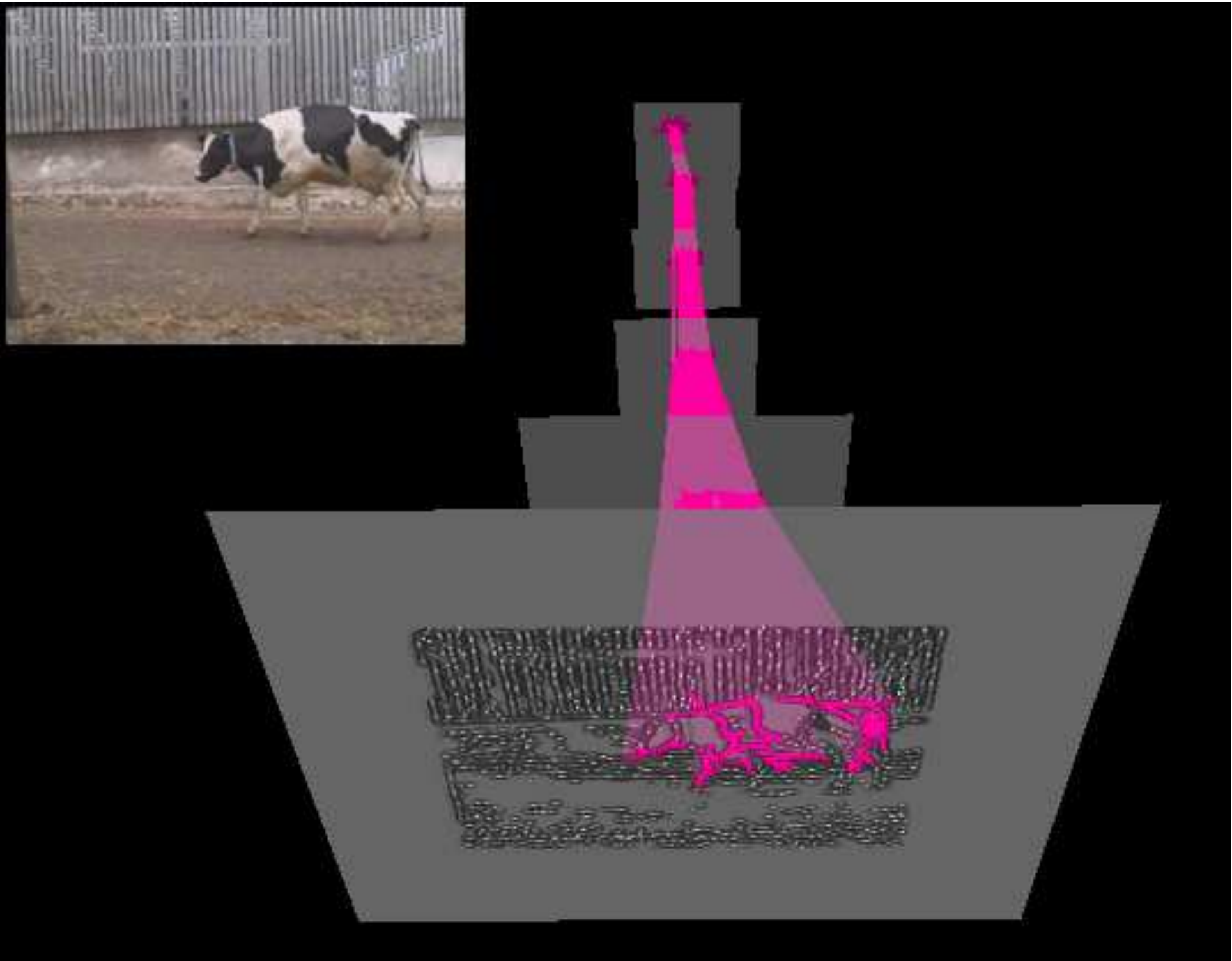}\hspace{-0.5mm}
\includegraphics[height=\IH,width=\IW]{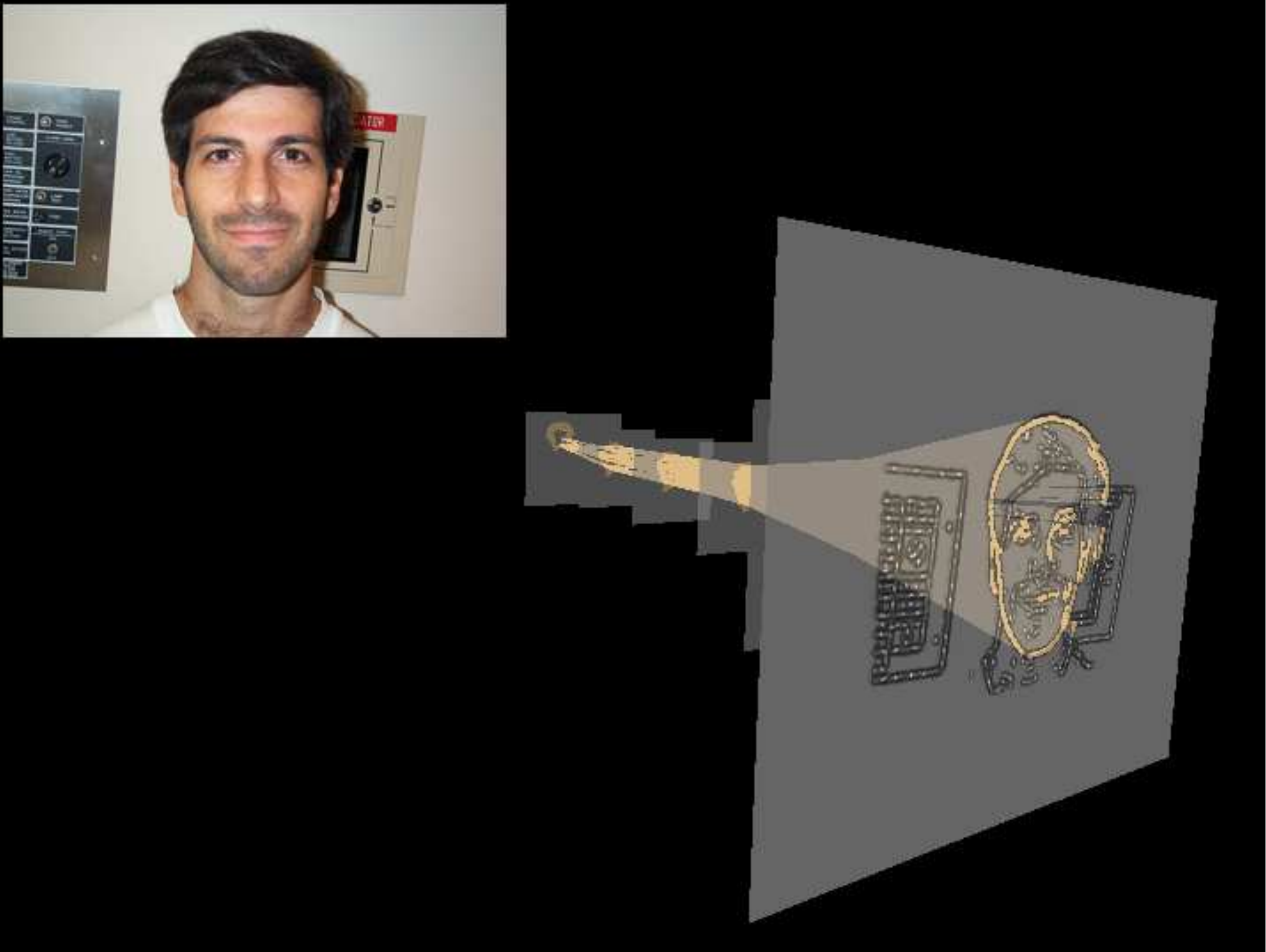}\hspace{-0.5mm}
\includegraphics[height=\IH,width=\IW]{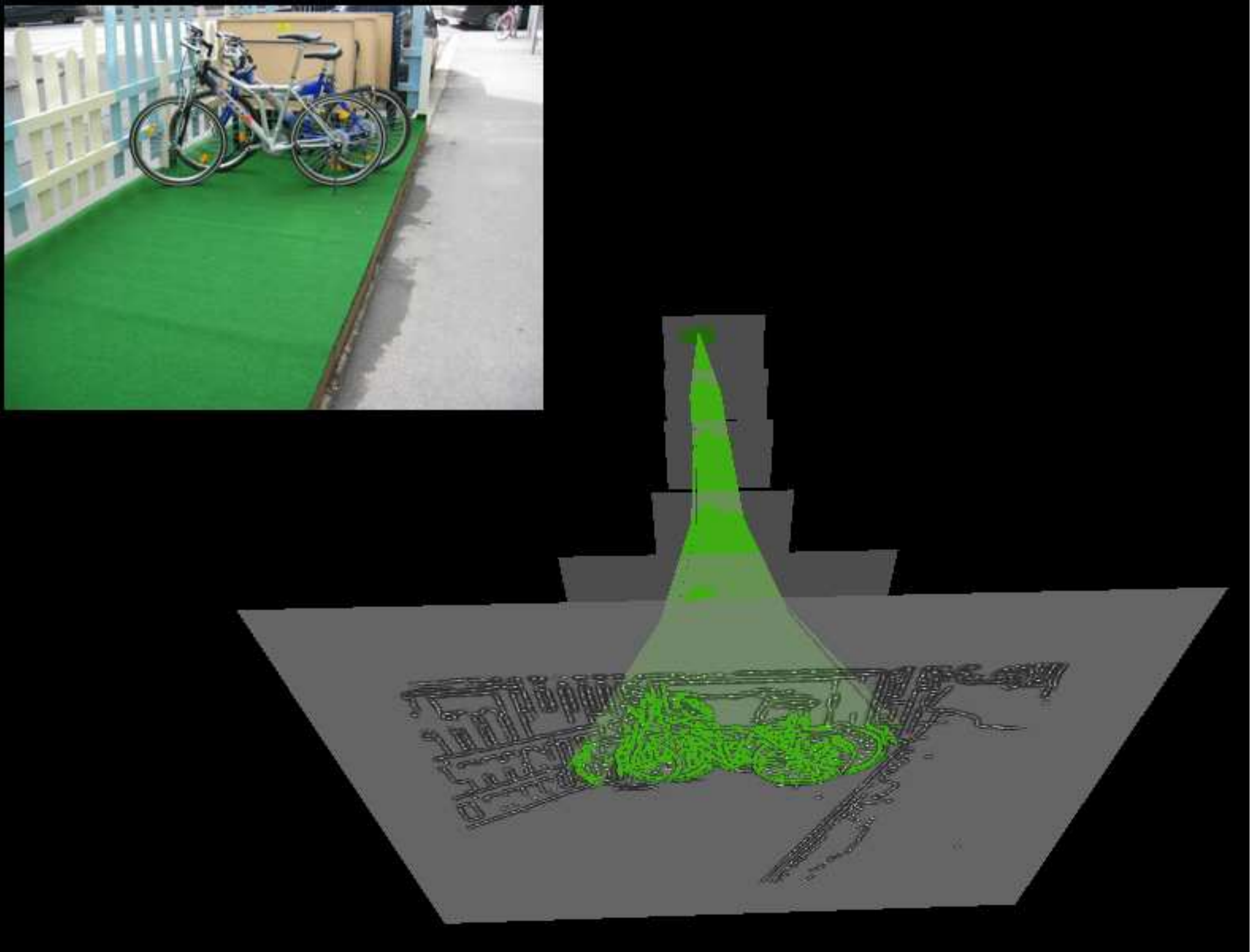}\hspace{-0.5mm}
\includegraphics[height=\IH,width=\IW]{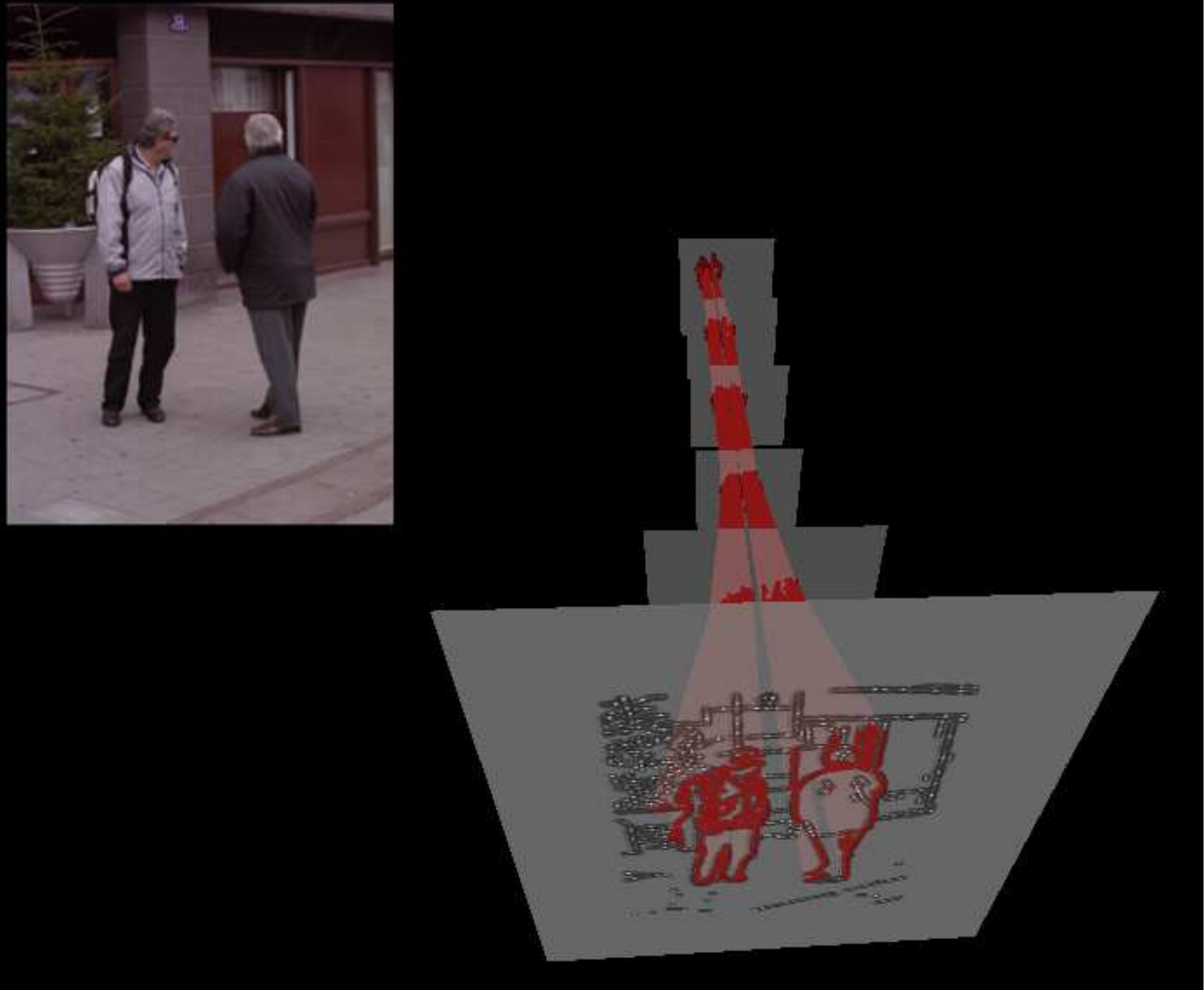}\hspace{-0.5mm}
\includegraphics[height=\IH,width=\IW]{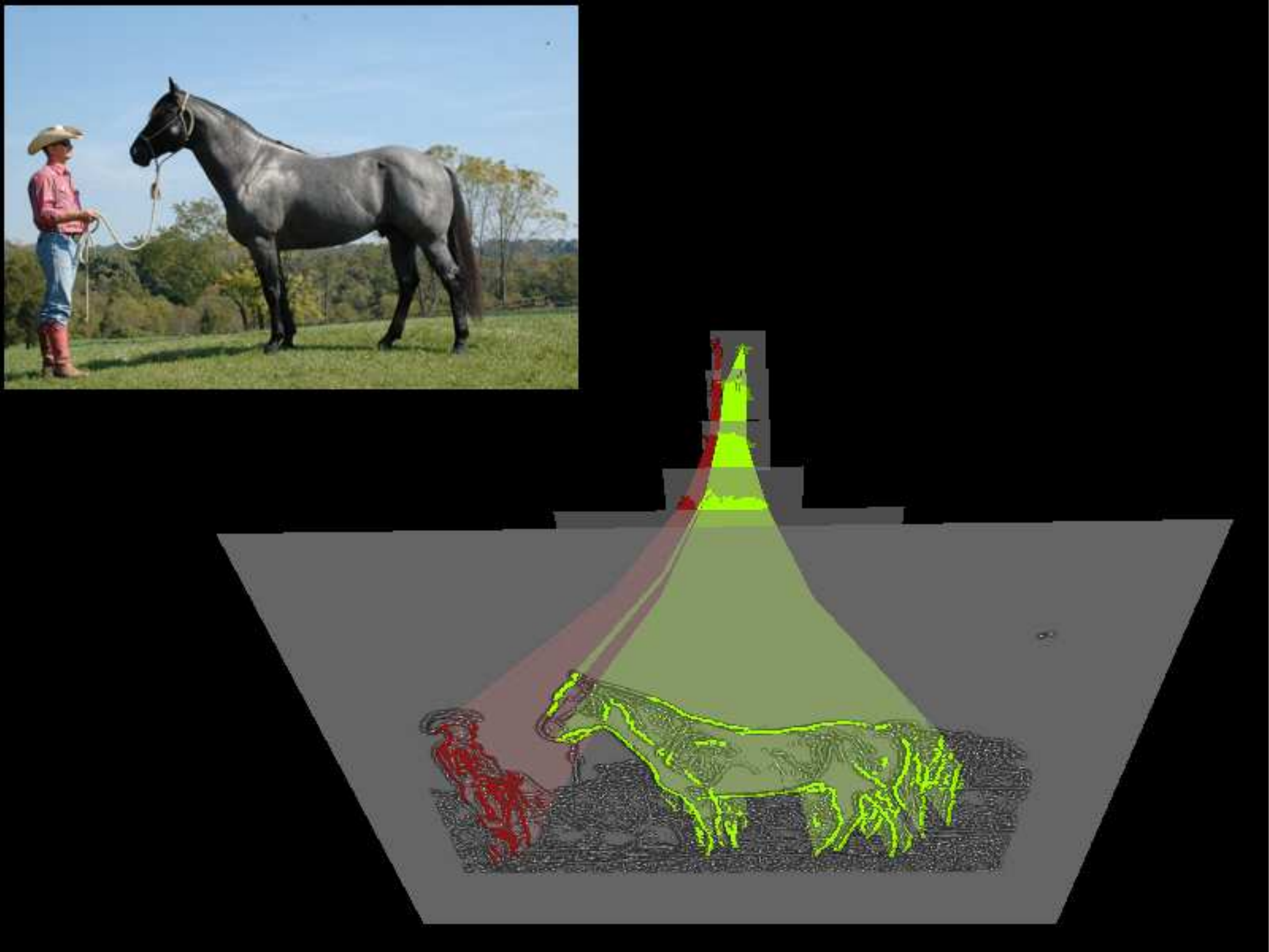}\hspace{-0.5mm}

 \vspace{-1.6mm}
\caption{{\small Examples of detections. The links show the most
probable hidden state activation for a particular (top-layer) class
detection. The links are color-coded to denote different classes.}}
 \label{fig:example_detections}
 \vspace{-1.5mm}
\end{figure*}


\section{Summary and conclusions}
\label{sec:discussion}

We proposed a novel approach which learns a hierarchical
compositional shape vocabulary to represent multiple object classes
in an unsupervised manner. Learning is performed bottom-up, from
small oriented contour fragments to whole-object class shapes. The
vocabulary is learned recursively, where the compositions at each
layer are combined via spatial relations to form larger and more
complex shape compositions.

Experimental evaluation was two-fold: one that shows the capability
of the method to learn generic shape structures from natural images
and uses them for object classification, and another one that
utilizes the approach in multi-class object detection. We have
demonstrated a competitive classification and detection performance,
fast inference even for a single class, and most importantly, a
logarithmic growth in the size of the vocabulary (at least in the
lower layers) and, consequently, scalability of inference complexity
as the number of modeled classes grows. The observed scaling
tendency of our hierarchical framework goes well beyond that of a
flat approach~\cite{s:opelt08}. This provides an important showcase
that highlights learned hierarchical compositional vocabularies as a
suitable form of representing a higher number of object classes.

\section{Future work}
\label{sec:future_work}

There are numerous directions for future work. One important aspect
would be to include multiple modalities in the representation. 
Since many object classes have distinctive textures and
color, adding this information to our model would increase the range
of classes that the method could be applied to. Additionally,
modeling texture could also boost the performance of our current
model: since textured regions in an image usually have a lot of
noisy feature detections, they are currently more susceptible to
false positive object detections. Having a model of texture could be
used to remove such regions from the current inference algorithm.
Another interesting possible way of dealing with this issue would be to use high quality bottom-up region proposals
to either re-score our detections or be used in our inference. Using segmentation in detection has recently been shown as a very promising approach~\cite{selective13,fidler13,girshick2014rcnn}.

Part of our ongoing work is to make the approach scalable to a large
number of object classes. To achieve this, several improvements and
extensions are still needed. We will need to incorporate the
discriminative information into the model (to distinguish better
between similar classes), make use of contextual information to
improve performance in the case of ambiguous information (small
objects, large occlusion, etc) and use attention mechanisms to
speed-up detection in large complex images. Furthermore, a taxonomic
organization of object classes could further improve the speed of
detection~\cite{fidler10} and, possibly, also the recognition rates of the approach.




\ifCLASSOPTIONcompsoc
  \section*{Acknowledgments}
\else
  \section*{Acknowledgment}
\fi

This research has been supported in part by the following funds:
Research program Computer Vision P2-0214 (Slovenian Ministry of
Higher Education, Science and Technology) 
and EU FP7-215843 project POETICON.

\ifCLASSOPTIONcaptionsoff
  \newpage
\fi

\bibliographystyle{IEEEtran}
\bibliography{sbiblio}




\end{document}